\documentclass[runningheads]{llncs}

\usepackage{accv}

\usepackage{accvabbrv}

\usepackage{graphicx}
\usepackage{booktabs}

\usepackage[accsupp]{axessibility}  

\usepackage[pagebackref,breaklinks,colorlinks,citecolor=accvblue]{hyperref}

\usepackage{orcidlink}

\usepackage{graphicx}
\usepackage{amsmath,amssymb} 
\usepackage{color}
\usepackage[export]{adjustbox}
\usepackage{hyperref}

\usepackage{xcolor}

\usepackage[thinc]{esdiff} 

\usepackage{multirow}
\usepackage{graphicx}
\usepackage{booktabs}

\usepackage{amssymb}
\usepackage{amsmath}

\usepackage{nccmath}

\usepackage{pifont}

\usepackage{mathtools}

\usepackage{tcolorbox}

\usepackage{colortbl}

\usepackage{array}

\usepackage{caption}
\usepackage{subcaption}

\usepackage{xspace}
\newcolumntype{P}[1]{>{\centering\arraybackslash}p{#1}}

\usepackage{dsfont} 

\definecolor{royalblue}{HTML}{4169E1}

\definecolor{susielamber}{HTML}{FFC000}
\definecolor{susielgold}{HTML}{FFD700}
\definecolor{susielpeach}{HTML}{FFB14E}
\definecolor{susielsalmon}{HTML}{FA8775}
\definecolor{susielraspberry}{HTML}{EA5F94}
\definecolor{susielhibiscus}{HTML}{D92C70}
\definecolor{susielorchid}{HTML}{CD34B5}
\definecolor{susieldarkorchid}{HTML}{9D02D7}
\definecolor{susielblue}{HTML}{0000FF}

\definecolor{turquoise}{HTML}{2DB3C3}

\newcommand{\checked}{\hspace{0.1em}\ding{51}}

\colorlet{semigray}{black!40}
\newcommand{\refConf}[1]{\textcolor{black!50!white}{(#1)}}

\newcommand{\sbul}[1]{{\color[rgb]{0.5,0.5,1.0}#1}}

\newcommand{\rred}[1]{Lazy-XMem}
\usepackage{xspace}
\usepackage{siunitx}

\makeatletter
\DeclareRobustCommand\onedot{\futurelet\@let@token\@onedot}
\def\@onedot{\ifx\@let@token.\else.\null\fi\xspace}  

\def\eg{\emph{e.g}\onedot}             
            
\def\ie{\emph{i.e}\onedot}             
            
\def\wrt{w.r.t\onedot}                  
                   
\def\etal{\emph{et al}\@onedot}          
\makeatother

\newcommand{\tabref}[1]{Table~\ref{#1}}

\usepackage[printonlyused]{acronym}
\acrodef{ml}[ML]{Machine Learning}
\acrodef{convnet}[ConvNet]{Convolutional Neural Network}

\acrodef{vos}[VOS]{Video Object Segmentation}
\acrodef{sota}[SOTA]{State-of-the-Art}
\acrodef{avos}[aVOS]{automatic Video Object Segmentation}
\acrodef{svos}[sVOS]{semi-automatic Video Object Segmentation}
\acrodef{iios}[iIOS]{interactive Image Object Segmentation}
\acrodef{ivos}[iVOS]{interactive Video Object Segmentation}
\acrodef{vots}[VOTS]{Video Object Traking and Segmentation}
\acrodef{aivos}[ziVOS]{la\underline{z}y \underline{i}nteractive \underline{V}ideo \underline{O}bject \underline{S}egmentation}
\acrodef{dnns}[DNNs]{Deep Neural Networks}
\acrodef{sam}[SAM]{Segment Anything Model}

\acrodef{idi}[IDI]{Interaction Density Index}
\acrodef{aci}[ACI]{Average Correction Interval}
\acrodef{noc}[NoC]{Number of Correction}

\acrodef{iou}[IoU]{Intersection over Union}
\acrodef{auc}[AUC]{Are Under Curve}

\def\JandF{$\mathcal{J}\mathcal{\&}\mathcal{F}$\xspace}

\usepackage{tikz}			
\usepackage{pgfplots}		

\usepackage{pgfplotstable}
\usepgfplotslibrary{statistics}
\pgfplotsset{compat=newest}
\usepackage{filecontents}  	
\usepackage{subcaption}		
\usetikzlibrary{patterns}  

\usepackage{booktabs}  
\usepackage{tabularx}  
\usepackage{lipsum} 

\usetikzlibrary{shapes.geometric} 
\newcommand\drawstar[1]{%
	\tikzset{scorestars/.style={star, star points=5, star point ratio=2.25, draw, inner sep=1.3pt, anchor=outer point 3, fill=#1, line width=0.8}}%
	\begin{tikzpicture}[baseline]
		\node[scorestars] {};
	\end{tikzpicture}%
}

\def\DeepLabv3+{DeepLabv3+~\cite{Chen2018}}
\def\ResNet50{ResNet50~\cite{He2016}}

\def\DAVIS17{DAVIS 2017~\cite{Pont-Tuset_arXiv_2017}\xspace}
\def\D17{D17~\cite{Pont-Tuset_arXiv_2017}\xspace}
\def\LV1{LV1~\cite{AFB-URR}\xspace}

\def\BL30K{{\href{https://github.com/hkchengrex/MiVOS}{BL30K}}~\cite{Cheng_CVPR_2021}}

\def\Pascal_VOC{PASCAL VOC~\cite{Everingham2011}}

\usepackage{xspace}
\usepackage{siunitx}

\usepackage{xcolor}

\usepackage{multirow}
\usepackage{graphicx}
\usepackage{booktabs}

\usepackage{amssymb}
\usepackage{amsmath}

\usepackage{nccmath}

\usepackage{pifont}

\usepackage{mathtools}

\usepackage{tcolorbox}

\usepackage{colortbl}

\usepackage{array}

\newcommand*{\tran}{^{\mkern-1.5mu\mathsf{T}}}

\DeclareMathOperator*{\argmax}{argmax}   

\graphicspath{{./figures/}, {./images/}}

\usepackage{multibib}
\newcites{supp}{References for Supplementary Material}

\usetikzlibrary{positioning}

\begin{document}
	
\title{Strike the Balance: On-the-Fly Uncertainty based User Interactions for Long-Term Video Object Segmentation} 

\titlerunning{Lazy-XMem}

\author{Stéphane Vujasinovi\'{c}\inst{1}\orcidlink{0000-0001-6916-786X} \and
	Stefan Becker\inst{1}\orcidlink{0000-0001-7367-2519} \and
	Sebastian Bullinger\inst{1}\orcidlink{0000-0002-1584-5319} \and
	Norbert Scherer-Negenborn\inst{1}\orcidlink{0000-0001-8085-2147} \and
	Michael Arens\inst{1}\orcidlink{0000-0002-7857-0332} \and
	Rainer Stiefelhagen\inst{2}\orcidlink{0000-0001-8046-4945}}

\authorrunning{S. Vujasinovi\'{c} et al.}

\institute{Fraunhofer IOSB, Ettlingen, Germany 		\\
	\email{\{name\}.\{surname\}@iosb.fraunhofer.de} \\ 
	\and Karlsruhe Institut of Technology, Germany 	\\
	\email{\{name\}.\{surname\}@kit.edu}}
\maketitle


\begin{abstract}
In this paper, we introduce a variant of video object segmentation (VOS) that bridges interactive and semi-automatic approaches, termed Lazy Video Object Segmentation (ziVOS).
In contrast, to both tasks, which handle video object segmentation in an off-line manner (\ie, pre-recorded sequences), we propose through ziVOS to target online recorded sequences.
Here, we strive to strike a balance between performance and robustness for long-term scenarios by soliciting user feedback's on-the-fly during the segmentation process.
Hence, we aim to maximize the tracking duration of an object of interest, while requiring minimal user corrections to maintain tracking over an extended period.
We propose \rred{} as a competitive baseline, that estimates the uncertainty of the tracking state to determine whether a user interaction is necessary to refine the model's prediction.
We introduce complementary metrics alongside those already established in the field, to quantitatively assess the performance of our method and the user's workload.
We evaluate our approach using the recently introduced LVOS dataset, which offers numerous long-term videos.
Our code is available at \href{https://github.com/Vujas-Eteph/LazyXMem}{https://github.com/Vujas-Eteph/LazyXMem}.
\keywords{Video Object Segmentation \and Interactive}
\end{abstract}

\section{Introduction}
\label{sec:introduction}
\ac{vos} is a fundamental challenge involving various tasks, including \ac{svos} and \ac{ivos}~\cite{zhou2023survey}.
In~\ac{svos}, given an initial segmentation mask for the first video frame, methods classify each pixel in the subsequent video frames as a part of the object of interest (\ie, foreground) or the background.
Here, a user only interacts at the start of the sequence by providing the corresponding annotation mask to indicate which object to segment in the video.
In contrast, \ac{ivos} methods incorporate user interactions in a multi-round scheme, where the user interacts with the method before each round, to improve the segmentation quality on the subsequent rounds.
Both applications are suited for pre-recorded sequences, \ie, offline segmentation, as \ac{svos} methods assumes that the user has unlimited time to annotate the initial frame with utmost accuracy, whereas \ac{ivos} approaches expect the user to inspect the segmentation quality of the previous round, and interact for multiple rounds until the desired segmentation quality is achieved.
However, while \ac{svos} methods demonstrate impressive performances on short-term datasets~\cite{Pont-Tuset_arXiv_2017,xu2018youtube}, their applicability to long-term sequences remains under-explored~\cite{AFB-URR,XMEM,QDMN,DEAOT,cheng2023putting,Hong_2023_ICCV,bekuzarov2023xmem,Vujasinovic_2023_BMVC} and yet to be addressed by \ac{ivos} methods.

\begin{figure}[t]
	\centering
	\includegraphics[width=1\textwidth, trim={0 0 0 0}, clip]
	{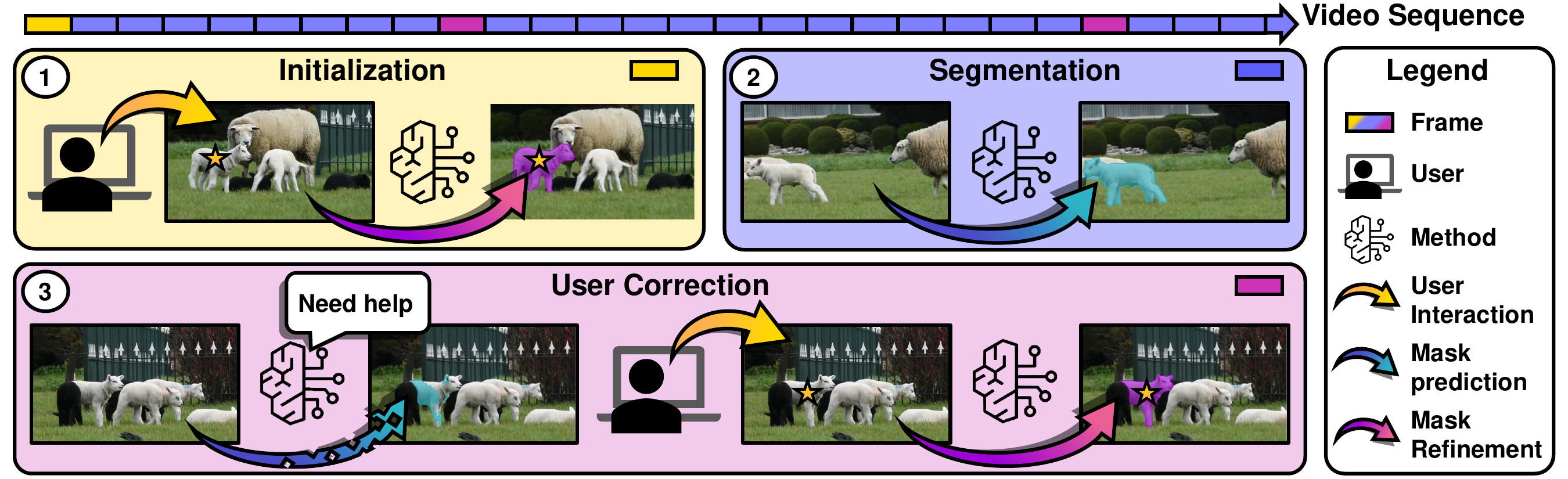}
	\caption{
		Visual representation of the \ac{aivos} framework.
		(1) The user initiates the segmentation by clicking to identify the object of interest in the video, (2) thus indicating which object to segment.
		Only when requested by the method (3) does the user provides corrective clicks on-the-fly.
	}
	\label{Intro Fig}
\end{figure}

This underscores a gap in methodologies suited for prolonged sequences, where maintaining error-free segmentation under challenging conditions becomes increasingly difficult.
In this paper, we explore a hybrid framework, named~\ac{aivos} (depicted in \cref{Intro Fig}), that bridges the methodologies of \ac{svos} and \ac{ivos}, focusing on maintaining robust object tracking with minimal user interactions.
Unlike \ac{ivos}, we discard the round based scheme and integrate user corrections on-the-fly, refining the model's prediction as needed, while the method segments the video.
Moreover, distinct from \ac{svos}, in \ac{aivos} the object of interest is indicated with a user interaction (\ie, click).
To achieve this, we only allow one interaction per frame and per object, and solely rely on click-based interactions, as pointing an object is the quickest, most intuitive and predictable interaction type for humans~\cite{clark2005coordinating,firestone2014please}.
Hence, we propose~\ac{aivos} to emulate a human-in-the-loop process when segmenting a video in an online fashion, that is better suited for dynamic applications, when user engagement is feasible and where maintaining consistent object tracking in challenging conditions is more critical than achieving segmentation accuracy.
Concretely, our objective shifts from segmenting an object with high accuracy to maximizing the number of frames in which the object is segmented above a minimal alignment ratio (\ie, \ac{iou}), denoted as $\tau_{\text{iou}}$, by integrating user corrections on-the-fly (only at critical events), while simultaneously reducing the user's workload.

We propose \rred{}, as a baseline for future works addressing~\ac{aivos}.
\rred{} assess the uncertainty of a predicted object mask on-the-fly and refines it accordingly (through SAM-HQ~\cite{sam_hq}), if the uncertainty is too high, through either pseudo-corrections or user-corrections.
In our approach, the Shannon entropy~\cite{shannon1948mathematical} serves as a proxy to estimate the performance of the tracking state \textendash\xspace~alignment (\ie, \ac{iou}) ratio between the predicted mask and a hypothetical ground-truth.
Similarly, recent studies in~\ac{ivos} \cite{yin2021learning,delatolas2024learning} evaluate which frame to interact with at the end of a round.
They compare the embeddings of each frame in the video sequence against all other frame embeddings to determine which frame to suggest to the user for new interactions, limiting this strategy to only pre-recorded videos.
In contrast, our criterion is solely defined \wrt the tracker's state, how accurate the prediction is for the current observed frame, allowing us to work with non-prerecorded sequences.
To our knowledge, only QDMN by Liu~\etal~\cite{QDMN} also estimates the tracking state in an online fashion, by predicting a quality score through a second head (following a similar design to~\cite{huang2019msrcnn}). 
However, unlike prior works~\cite{QDMN,yin2021learning,delatolas2024learning}, we estimate the tracker's state on pixel level in a \textit{post-hoc} fashion, removing the need to train an auxiliary network.
An additional benefit, to computing the uncertainty on pixel-level is that we can visually indicate ambiguous regions to the user where an interaction might be the most helpful.
Moreover, depending on the confidence of the predicted mask, \rred{} decides whether the current predicted mask will be stored in the memory.
Hence, by selectively refining masks based on entropy-driven uncertainty estimation, we aim to maintain a balance between robustness and user-workload in \ac{aivos}, specifically in long-term scenarios.
Thus, while on-the-fly interactions may suggest constant user supervision, our approach minimizes this need by prompting the user (ideally) only at critical events \textendash\xspace~when Lazy-XMem is uncertain about its prediction.
Our goal is to reduce the cognitive load by allowing the user to intervene only when necessary, enabling them to focus on other tasks simultaneously.

In this context the paper presents the following contributions:
(1) Online (on-the-fly) assessment of the tracking state quality, by leverage entropy, to minimize the user's monitoring by providing interactions only at critical events (\eg, occlusions, distractors).
(2) A scheme to integrate pseudo-interactions, into our interactive feed-back loop to reduce the user's workload. We generate pseudo-interactions based on the original mask and the corresponding uncertainty.
(3) Suitable metrics to evaluate the robustness of our method, and the user’s workload \wrt the standard \JandF metric proposed by Perazzi~\etal~\cite{Perazzi_CVPR_2016}.
(4) Evaluation on long-term~\cite{Hong_2023_ICCV} sequences to highlight the suitability of our method to maintain robust tracks.

\section{Related Work}
\label{sec:related_work}
\subsection{Semi-Automatic Video Object Segmentation}
Early deep learning methods in \ac{svos} follow an \textit{online fine-tuning approache}~\cite{voigtlaenderonline,caelles2017one,maninis2018video,xiao2018monet,xiao2019online,meinhardt2020make}, which adapts the network's parameters on-the-fly while segmenting the objects of interest in the video sequence.
This results, in slow inference times and poor generalization capabilities~\cite{zhou2023survey}.
Concurrently, \textit{propagation-based methods}~\cite{perazzi2017learning,jang2017online,hu2018motion,zhang2019fast,johnander2019generative} propagate the masks from the previous adjacent frame to the current one for segmentation, but they are prone to error accumulation and often fail during occlusions~\cite{zhou2023survey}.
\textit{Matching-based methods}~\cite{DEAOT,voigtlaender2019feelvos,CFBI,CFBIP,AOT,TBD,JOINT} leverage features from the initial and previous adjacent frame to segment the current frame.
The leading methods in the field further integrate features from in-between frames (previously processed) into an external memory~\cite{DEAOT,AOT,STM,RMNet,KMN,HMMN,GSFM,MiVOS,STCN,PCVOS}, using cross-attention to link features from previous frames to the current frame to segment.
However, these methods are limited in real-world applications due to their expanding memory requirements, making long-term segmentation on consumer-grade GPUs challenging.

Recent works address this bottleneck by selectively integrating frame representations into the external memory~\cite{QDMN,RDE-VOS,Vujasinovic_2023_BMVC} or by generating compact representations to summarize similar features together~\cite{AFB-URR,XMEM,cheng2023putting,bekuzarov2023xmem,FVOSGCM}.
These methods effectively manage the memory footprint, enabling more efficient \ac{svos} on long-term videos. Newer methods~\cite{cheng2023putting} also explore improved ways to differentiate similar objects (distractors) from each other.
Additionally, new datasets have recently been introduced~\cite{Hong_2023_ICCV,athar2023burst,MOSE} to provide an alternative to the classical DAVIS~\cite{Pont-Tuset_arXiv_2017} and YouTube-VOS~\cite{xu2018youtube} datasets, with some targeted specifically for long-term video segmentation~\cite{AFB-URR,Hong_2023_ICCV} and tracking~\cite{VOTS2023}.

Contemporary works~\cite{yang2023track,cheng2023segment,hqtrack} leverage \ac{sam}~\cite{kirillov2023segany} or a variant~\cite{sam_hq,FastSAM,mobile_sam} to refine the original mask predicted by an \ac{svos} baseline~\cite{XMEM,DEAOT}.
However, in contrast to our framework, they refine every $n$-th mask predicted by the \ac{svos} backbone with a SAM based approach~\cite{sam_hq,kirillov2023segany,FastSAM,mobile_sam}, and require continuous user monitoring to identify when interventions are needed.
Furthermore, they diminish the influence to enhance the predictive accuracy for subsequent frame as they do not update the memory with the refined mask.

\subsection{Interactive Image Segmentation}
In \ac{iios}, methods predict a mask for an object of interest based on user interactions for a single image.
These approaches aim to reduce the user's workload by replacing densely annotated mask for sparse annotations (\eg, clicks~\cite{xu2016deep,li2018interactive,Jang2019,forte2020getting,Sofiiuk2020,kontogianni2020continuous,Sofiiuk2021}, extreme points~\cite{maninis2018deep,dupont2021ucp,zhang2020interactive}, or bounding boxes~\cite{rother2004grabcut,wu2014milcut,xu2017deep}).
Most notable approach is f-BRS~\cite{Sofiiuk2020} which optimizes internal auxiliary features of the segmentation network to align its prediction's at the clicked position with the user annotated label.
A follow up work by Sofiiuk \etal~\cite{Sofiiuk2021}, replaces the previous f-BRS backbone with an HRNet~\cite{HRNet} + OCR~\cite{OCR} network, to maintain high quality features through out the network to obtain a preciser segmentation mask.
Since the introduction of SAM~\cite{kirillov2023segany}, a plethora of SAM-based methods have been proposed to solve the task in medical imaging~\cite{marinov2023deep} and natural images~\cite{chunhui2023samsurvey}.
For instance, SAM-HQ~\cite{sam_hq} improves upon SAM by better handling complex shapes, such as thinner structures and objects with holes.
Additionally, faster approaches like FastSAM~\cite{FastSAM} and MobileFast~\cite{mobile_sam} have been developed to enhance performance and efficiency.

\subsection{Interactive Video Object Segmentation}
Originally intended to reduce the user's workload during video annotations~\cite{Caelles_arXiv_2018}, \ac{ivos} methods integrate user interactions in a round-based process.
Most approaches follow the design introduced by Benard \etal~\cite{benard2017interactive}, which combines \ac{svos} and \ac{iios} pipelines.
The blueprint process for the \ac{ivos} task is as follow: (1) Firstly, the method predicts a segmentation mask for each frame in the video (through a \ac{svos} baseline), based on an initial mask provided for a frame.
(2) Next, a user scrolls through the resulting masks and selects a frame to interact with (\eg, through clicks~\cite{Wang2014TouchCutFI,jain2016click,Vujasinovic_2022_ICIP}, scribbles~\cite{MiVOS,Oh_CVPR_2019,Miao_CVPR_2020,Heo_ECCV_2020,heo2021guided}).
Based on the provided interaction, a new mask is predicted for the annotated frame, serving as a new starting point when repeating step (1).
Steps (1) and (2) are repeated one after the other, until the user is satisfied with the final results.

A persistent bottleneck is determining which frame to annotate for the next round.
Recent approaches address this by identifying a quartet of candidate frames for the user to annotate~\cite{heo2021guided}, estimating which frame would yield the most improvement~\cite{yin2021learning}, or using a weakly supervised method to indicate the frame and type of interaction~\cite{delatolas2024learning} to the user.
To determine which frame or set of frames to annotate, these methods map each frame in the sequence into an embedding space, restricting them to short videos, as it requires storing the embedding of every frame.
Here, each embedding encode the frame's representation and the quality of the corresponding predicted mask.
The best candidate frame is selected by comparing each embedding \wrt others and against those of annotated frames, either through an agent~\cite{yin2021learning} or by choosing the embedding that is furthest from any annotated embedding~\cite{delatolas2024learning}.
In contrast, our approach introduces corrections on-the-fly by directly assessing the tracking state during segmentation, thereby proposing an online methodology that is also not restricted to short sequences.

\subsection{Uncertainty Estimation in Video Object Segmentation}
Uncertainty estimations is essential to improve the reliability and explainability of a model, however estimating the uncertainty of \ac{dnns}, remains a challenging topic.
To our knowledge, only the work by Liu~\etal~\cite{QDMN} incorporates a confidence score to asses the tracking state on-the-fly for the~\ac{svos} task by leveraging an auxiliary head (\ie, QAM module), predicting a confidence score on how likely the predicted mask would align with a ground-truth annotation.
Similar to our approach, QDMN manages its memory updates based on a threshold value that determines whether a predicted mask, given its confidence level, is reliable enough to be stored in the external memory.
However, as the QAM module only predicts a single score per object, it is unable to guide the user during an interaction, as to where a correction might be the most valuable.
In contrast, we explore uncertainty estimation through information theory~\cite{shannon1948mathematical} (\ie, Shannon entropy) and update the memory with the refined mask.

\section{Method}
\label{sec:method}
\begin{figure}[t]
	\centering
	\includegraphics[width=\textwidth]
	{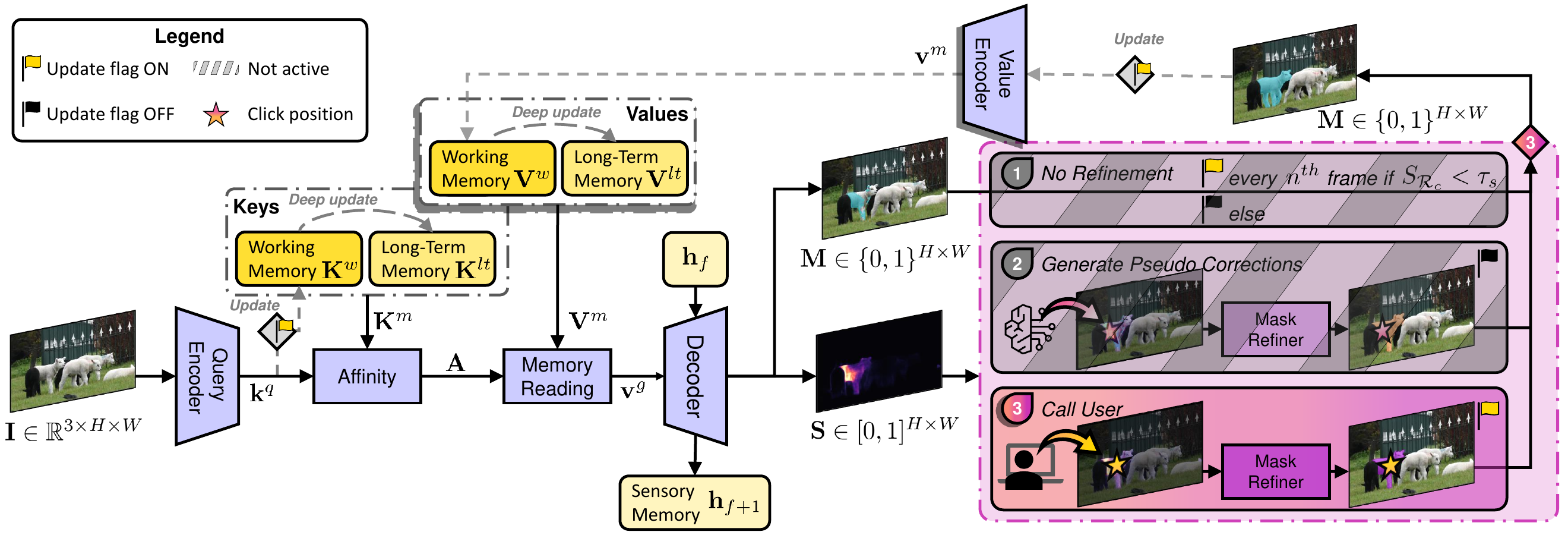}
	\caption{Overview of~\rred{NAME} for Lazy Video Object Segmentation.
	Our method is relies on an \ac{svos} baseline (\ie, XMem~\cite{XMEM}).
	We leverage the entropy to estimate on-the-fly the tracking state.
	Based on the tracking state's, the method either uses the original mask of the \ac{svos} baseline, or refine the original mask by generating pseudo-interactions, or requesting user interaction.}	
	\label{Method}
\end{figure}

We present \rred{NAME}, depicted in \cref{Method}, as a baseline for future works targeting \ac{aivos}.
\rred{NAME} comprises the following key components: (1) An \ac{svos} baseline, to predict object masks; (2) An uncertainty assessment component; (3) A mask refiner, to refine the original prediction from the \ac{svos} baseline; (4) An interaction-issuer, to issue either pseudo- or user-corrections and (5) a memory update mechanism.

\subsection{XMem as Baseline}
\label{XMem}
We rely on XMem~\cite{XMEM} as our~\ac{svos} baseline.
Initialized with an object mask at the beginning, the network predicts masks for subsequent frames.
For simplicity, we assume the network segments a single object.
The key components are:

\textbf{Convolutional Blocks}:
(1) A \textit{query encoder}, that extracts query key features~$\mathbf{k}^{q} \in \mathbb{R}^{C_k \times \frac{H}{16}\frac{W}{16}}$ from the current image to segment.
(2) A \textit{decoder}, which predicts an object mask~$\mathbf{M}\in\{0,1\}^{H\times W}$ for a query frame~$\mathbf{I} \in \mathbb{R}^{3 \times H \times W}$.
(3) Lastly, a \textit{value encoder}, that extracts value features $\mathbf{v}^{m} \in \mathbb{R}^{C_v \times \frac{H}{16}\frac{W}{16}}$ based on the current image~$\mathbf{I}$ and the predicted mask~$\mathbf{M}$.

\textbf{Memories}:
Unlike previous works~\cite{STM,MiVOS,STCN}, XMem~\cite{XMEM} employs three distinct memories: a \textit{working memory}, a \textit{long-term memory}, and a \textit{sensory memory}.

(1) The \textit{working memory}, is updated every $n$-th frame with query and value representations $\mathbf{K}^{w} \in \mathbb{R}^{C_k \times t\frac{H}{16}\frac{W}{16}}$ and $\mathbf{V}^{w} \in \mathbb{R}^{C_v \times t\frac{H}{16}\frac{W}{16}}$, until it reaches a capacity of $t = T_{max}$.

(2) When the working memory is full, it is distilled into $l$ prototype features $\mathbf{k}^{p} \in \mathbb{R}^{C_k \times l}$ and $\mathbf{v}^{p} \in \mathbb{R}^{C_v \times l}$, based on usage frequency during memory reads.
These prototypes are added to the long-term memory $\mathbf{K}^{lt} \in \mathbb{R}^{C_k \times L}$ and $\mathbf{V}^{lt} \in \mathbb{R}^{C_v \times L}$, with least-frequent-usage (LFU) filtering to remove obsolete features.

(3) The \textit{sensory memory}, uses GRU cells~\cite{cho2014learning} to update a hidden representation $\mathbf{h}_f \in \mathbb{R}^{C_h \times \frac{H}{16} \times \frac{W}{16}}$ every frame, encoding prior information like position~\cite{XMEM}.

\textbf{Memory Reading}: During the memory read operation, feature representations from both working and long-term memories are used, totaling $N = T\frac{H}{16}\frac{W}{16} + L$ elements. The model computes the similarity between memory keys and query keys using an anisotropic $\ell_2$-similarity function~\cite{XMEM}, resulting in a similarity matrix $\mathbf{W}(\mathbf{K}^m,\mathbf{k}^q) \in \mathbb{R}^{N \times \frac{H}{16}\frac{W}{16}}$. Applying a softmax along the rows yields the affinity matrix $\mathbf{A}$.
A new value $\mathbf{v}^{g} \in \mathbb{R}^{C_v \times \frac{H}{16} \frac{W}{16}}$ is then generated for the decoder through
\begin{equation}
	\mathbf{v}^{g} = (\mathbf{V}^{m})\tran\mathbf{A}(\mathbf{K}^m,\mathbf{k}^{q}),
\end{equation}
where $\mathbf{K}^{m} = \mathbf{K}^{w} \oplus \mathbf{K}^{lt}$ and $\mathbf{V}^{m} = \mathbf{V}^{w} \oplus \mathbf{V}^{lt}$ are the concatenated working and long-term memories. The keys provide robust semantic information for matching, while the values encode boundary and texture cues needed for decoding~\cite{STM}.
For more detailed information, we refer readers to the original paper~\cite{XMEM}.

\subsection{Uncertainty Estimation via Entropy}
\label{Entropy}

To estimate the uncertainty of the tracking state (\ie, the predicted segmentation mask), we leverage the Shannon entropy~\cite{shannon1948mathematical}, denoted as $S$.
We consider pixels as discrete random variables whose classes $c$ belong to a set $\mathcal{C}$, contains every object observed in a given video, including the background (\textit{i.e.}, $c=0$).
We use the output values of the softmax layer as a approximation of a probability mass functions $p_{\mathcal{C}}(c \mid x_{h,w})$ for each pixel located~$x$ at~$(h,w)$.
Here, \(c \in \mathcal{C}\) denotes the class of the pixel, and $(h, w)$ specifies the pixel's location in terms of height $h$ and width $w$ within a mask $\mathbf{M} \in \{0,\dots,|\mathcal{C}| \}^{H \times W}$.

However, as the number of classes $|\mathcal{C}|$ can vary over time (\ie, from one video to another, or even from one frame to another in the same video), we normalize the entropy for consistency and comparability.
Hence, we express the entropy of a pixel $x_{h,w}$ by
\begin{equation}
	\mathbf{S}_{h,w} = - \frac{\sum_{c \in \mathcal{C}} p_{\mathcal{C}}(c \mid x_{h,w}) \log\left(p_{\mathcal{C}}(c \mid x_{h,w})\right)}{\log(|\mathcal{C}|)},
	\label{Normalized_Entropy}
\end{equation}
where $\mathbf{S} \in [0,1]^{H \times W}$ denotes the corresponding entropy map of the current frame to segment.
To compute the entropy (\ie, uncertainty) for a specific object class~$c$, we use a dilated mask~$\mathbf{M}_c^d$ based on the original object mask~$\mathbf{M}_c$, such that the dilated mask allows us to exclude the background noise, while still considering the uncertainty around the predicted object's edges.
We compute the dilated mask~$\mathbf{M}_c^d$ through
	\begin{equation}
		\mathbf{M}_c^d(h, w) = \max_{(i, j) \in \mathbf{K}} \mathbf{M}_c(h + i, w + j),
	\end{equation}
where~$\mathbf{K}$, a circular dilation kernel, determines the increase of $\mathbf{M}_c^d$ relative to $\mathbf{M}_c$ by a specified ratio. The variables $i$ and $j$ represent the coordinates within $\mathbf{K}$.
The supplementary material contains an empirical evaluation of suitable values for this hyper-parameter.
Otherwise, using directly the predicted mask~$\mathbf{M}_c$ might truncate the aleatoric uncertainty (especially near the edges of the object).
Hence, we compute the total uncertainty via the joint entropy $S_{\mathcal{R}_c}$ of a considered object region $\mathcal{R}_c = \{(h, w) \mid \mathbf{M}_c^d(h, w) = 1\}$, through
\begin{equation}
	S_{\mathcal{R}_c} = \sum_{r \in \mathcal{R}_c} \mathbf{S}_{r}\left(x_{r} \mid x_{r-1}, \dots, x_{1}\right) \approx \sum_{r \in \mathcal{R}_c} \mathbf{S}_{r},
	\label{region_entropy}
\end{equation}
by essentially summing the conditional entropies of each random variable within the considered region.
This approach captures the inter-dependencies among all variables, reflecting their collective impact on $\mathcal{R}_c$.
However, computing the joint entropy is impractical as the network does not provide any joint or conditional distributions for a formal evaluation.
Additionally, the computational cost would grow exponentially with respect to the number of classes~$\mathcal{C}$ and the size of the region~$\mathcal{R}_c$ (\ie, $O(|\mathcal{C}|^{|\mathcal{R}_c|})$ time complexity). To reduce the computational complexity, we assume zero mutual information between the predicted probability distributions of pixels in the region~$\mathcal{R}_c$.
This allows us to sum the entropy of each pixel~$\mathbf{S}_{h,w}$ belonging to the region of interest~$\mathbf{M}_{c}^{d}$ (refer to~\cref{region_entropy}), allowing us to significantly reduce the computational cost (\ie, to~$O(|\mathcal{C}| \times |\mathcal{R}_c|)$ complexity).

Additionally, considering that the object size may vary from one image to another, we divide~$S_{\mathcal{R}_c}$ by the size of the corresponding region~$|\mathcal{R}_c|$.
This dampens the fluctuation of~$S_{\mathcal{R}_c}$ due to object size variations.

\subsection{Mask Refinement}
\label{mask_refiner}
For the mask-refinement component, we rely on SAM-HQ~\cite{sam_hq}, which extends SAM~\cite{kirillov2023segany} to segment intricate object structures in more details, while preserving its zero-shot capabilities and flexibility.
SAM-HQ~\cite{sam_hq} introduces two additional components on top of SAM~\cite{kirillov2023segany}: (1) An \textit{HQ-output token} to correct the original SAM's mask. (2) A \textit{global-local features fusion}, which fuses early features with later ones (\ie, after the first and last global attention block respectively) to enrich the features used by the mask decoder.
For more details about SAM~\cite{kirillov2023segany} and SAM-HQ~\cite{sam_hq} we refer the reader to the original sources.

\subsection{Issuing Corrections}
\label{interAKT}
For a given object $c$, we record the corresponding masked entropy~$S_{\mathcal{R}_c}$ at each frame, such that~$\mathbf{s}_{\mathcal{R}_c} = \left[ S_{\mathcal{R}_c}(f=0), \dots, S_{\mathcal{R}_c}(f=F) \right]$, where~$f$ denotes the frame index and~$F$ the latest frame to segment.
Let~$\mathbf{s}_{\mathcal{R}_c}'$ denote the derivative of $\mathbf{s}_{\mathcal{R}_c}$, such that~$\mathbf{s}_{\mathcal{R}_c}' = \left[ \Delta S_{\mathcal{R}_c}(f=1), \dots, \Delta S_{\mathcal{R}_c}(f=F) \right]$, where~$\Delta S_{\mathcal{R}_c}(f) = S_{\mathcal{R}_c}(f) - S_{\mathcal{R}_c}(f-1)$. 
Depending on~$\Delta S_{\mathcal{R}_c}(F)$ we either generate a pseudo- or request a user-correction.

Following the definition in \ac{iios}, we denote a positive or negative click as a point-wise interaction to indicate a falsely classified region as either foreground or background.
\textbf{User-Correction (U-C)}:
We prompt a user correction whenever $\Delta S_{\mathcal{R}_c}(f) \geq \tau_{u}$, where $\tau_{u}$ denotes the threshold above which a user correction is requested.
The user indicates a foreground or background region via a positive or negative click, which is then processed by the mask refiner to generate a new mask.
The original mask is not used during refinement due to its high uncertainty.
Note, that during a user-interactions, Lazy-XMem does not pre-select candidates regions.
Instead, our method overlays the entropy on the image to visually guide the user's interaction towards the most uncertain regions in the model's predictions.
We illustrate this in the qualitative experiments of the supplementary material.
\textbf{Pseudo-Correction (P-C)}:
In addition to requesting on-the-fly user corrections, the model generates pseudo-corrections when $\tau_u > \Delta S_{\mathcal{R}_c}(f) \geq \tau_p$, with $\tau_p$ representing the lower bound for a pseudo-correction to be generated.
A pseudo-correction $p_f^c$ for object $c$, given frame~$f$, is by
\begin{align}
	\mathbf{E}_c(h, w) &= \min_{(h_r,w_r) \in \Omega} \sqrt{(h - h_r)^2 + (w - w_r)^2}, \label{eq:distance_field} \\
	p_f^c &= \argmax_{(h,w)} \left( \mathbf{M}^d_c  \odot \mathbf{E}_c \odot (\mathbf{1}_{H \times W} - \mathbf{S}) \right), \label{eq:pseudo_correction}
\end{align}
where~$\Omega$ denotes the set of pixel that belong to the boundaries of the object mask~$\mathbf{M}_c$, $\mathbf{E}_c$ a distance field, and $\odot$ represents the Hadamard product.

\subsection{Interaction and Uncertainty Driven Memory Updates}
\label{mem_upd}
At each user-correction, we update the working memory of our \ac{svos} baseline with the newly refined mask.
This update strategy, termed \textit{Interaction-Driven Update} (IDU), improves the method's robustness as the refined mask can influence the segmentation of the subsequent frames.
An additional update mechanism, named \textit{Uncertainty-Driven Update} (UDU), prevents updating the working memory with the original representation when the corresponding uncertainty~$S_{\mathcal{R}_c}$, exceeds~$\tau_m$ (similarly to QDMN~\cite{QDMN}).

\section{Metrics}
\label{metrics}
Since we introduce \ac{aivos}, we propose complementary metrics to the standard \JandF presented by Perazzi \etal~\cite{Perazzi_CVPR_2016} to quantify the user's workload in providing on-the-fly corrections and to evaluate the robustness of a given method.

\subsection{Robustness Metric}
We take inspiration from Kristan \etal~\cite{VOTS2023} and propose $\text{R@}\tau_{\text{IoU}}$ (higher is better) to measure the robustness of a method.
Given a threshold value~$\tau_{\text{IoU}}$, we compute the ratio of frames, in which the predicted object mask attains an \ac{iou} above or equal to~$\tau_{IoU}$ for all objects in a given dataset.
More formally, let $\mathcal{O}$ be the set of objects in the dataset, where $\mathcal{F}_o$ is the set of frames in which object $o$ is present, we define $R@\tau_{\text{IoU}}$, such that
\begin{equation}
	R@\tau_{\text{IoU}} = \frac{1}{|\mathcal{O}|} \sum_{o \in \mathcal{O}} \frac{1}{|\mathcal{F}_{o}|} \sum_{f \in \mathcal{F}_{o}} \mathds{1}_{[\text{IoU}(\mathbf{M}_f^o, \mathbf{GT}_f^o) \geq \tau_{IoU}]},
\end{equation}
where $\mathbf{M}_f^o$ and $\mathbf{GT}_f^o$ denote respectively the predicted mask and ground-truth annotation for object $o$ at frame $f$.
Like in~\cite{VOTS2023}, whenever the method correctly predicts the absence of an object we set $\mathds{1}_{\text{IoU}(\mathbf{M}_f^o, \mathbf{GT}_f^o) \geq \tau_{IoU}}$ to $1$, otherwise to $0$.

\subsection{User-Workload Metrics}
To quantitatively evaluate the workload for the user we introduce the following metrics:
(1) \ac{noc} to denote the total number of user-corrections issued by the model to refine its current prediction.
(2) \ac{idi} (higher is better), is introduced as an intuitive metric that reports the average time between two user-corrections reported in seconds.
Note that some sequences might have no user interactions; however, to include every sequence in the evaluation, we consider the initialization and the end of a sequence as user-interactions.
(3) As \ac{idi} does not reflect the underlying distribution of the interactions, we provide through \ac{aci} (which encapsulates both \ac{noc} and \ac{idi}) a score to indicate this distribution.
In essence, we compute the cumulative count over user interactions and their respective distance to each other.
Consequently, a low \ac{aci} score indicates more spread out user interactions, while a higher score indicates consecutive interactions more closer to each other within a short period.
More formally, let $\mathcal{N}_o = \{f_{p=0},\cdots, f_{p=P_o} \mid f_p \in \mathcal{F}_o \}$ denote the set containing the frame indexes~$f_p$ where a user prompt~$p$ is issued for object $o$.
Hence, we define \ac{aci}, such that
\begin{equation}
	\text{\ac{aci}} = \sum_{o \in \mathcal{O}}\frac{1}{|\mathcal{F}_o|}\sum_{i=1}^{|\mathcal{F}_o|}\sum_{j=1}^{i}n_j,
\end{equation}
where $n_j = \sum_{f_{p}=1}^{\mathcal{N}_o} \mathds{1}_{\left[ j = f_{p} - f_{p-1}\right]}$ denotes the number of occurrences a user provided corrections at a distance of $j$ frames from one prompt to the next.

\section{Experiments}
\label{sec:evaluation}
In \cref{entropy_proxy}, we assess the effectiveness of the proposed masked entropy $S_{\mathcal{R}_c}$ to estimate the tracking's state on-the-fly. We present the evaluation protocol for the \ac{aivos} benchmark in~\cref{metrics_eval}, and present our results on the LVOS dataset~\cite{Hong_2023_ICCV} in~\cref{quant}.
Lastly, an ablation study in \cref{ablation} examines the impact of each design choice. We provide qualitative results in the supplementary material.

\subsection{Entropy as a Proxy}
\label{entropy_proxy}
To evaluate the effectiveness of the masked entropy $\mathcal{S}_{\mathcal{R}_c}$ to estimate the tracking's state, we compare against the following approaches:
(1) Using the Quality-Aware Module~(QAM) from QDMN~\cite{QDMN}, which predicts a confidence score through an auxiliary network.
(2) Computing the entropy~$S$ and its masked version~$S_{\mathcal{R}}$ for various models: single models denoted as Q and X respectively for the QDMN\cite{QDMN} and XMem~\cite{XMEM} networks, an ensemble model denoted as E, and a Monte-Carlo dropout model denoted as M.
We provide in the supplementary material details to the ensemble and Monte Carlo dropout approaches.
(3) We also consider for the ensemble and Monte-Carlo dropout variants the epistemic uncertainty, denoted as $V$ (and the masked version~$V_{\mathcal{R}}$).
For each method, we compute the Spearman coefficient~\cite{Spearman1904} to measure the correlation between each variant's output for the tracking state \wrt the actual \ac{iou}.
We conduct our evaluations on the DAVIS 2017~\cite{Pont-Tuset_arXiv_2017} and LVOS~\cite{Hong_2023_ICCV} validation sets, featuring short and long videos respectively.

\begin{figure}[t]
	\centering
	\begin{subfigure}{0.49\textwidth}
		\begin{tikzpicture}
			\begin{axis}
				[
				xtick={1, 2, 3, 4, 5, 6, 7, 8, 9, 10, 11, 12, 13},
				xticklabels={
					QAM,
					Q-$S$, Q-$S_{\mathcal{R}}$,
					X-$S$, X-$S_{\mathcal{R}}$, 
					E-$S$, E-$S_{\mathcal{R}}$,
					E-$V$, E-$V_{\mathcal{R}}$, 
					M-$S$, M-$S_{\mathcal{R}}$,
					M-$V$, M-$V_{\mathcal{R}}$},
				x tick label style={rotate=90},
				boxplot/draw direction=y,
				width=6.5cm,height=4.5cm,
				ymin=-1.05, ymax=1.05]
				\addplot[
				fill,
				fill opacity=0.2,
				pattern=north east lines,
				pattern color=black
				] [boxplot prepared={
					upper whisker=0.873,
					upper quartile=0.712,
					median=0.504,
					lower quartile=0.120,
					lower whisker=-0.473
				}]
				coordinates {};
				\addplot[
				fill,
				fill opacity=0.2,
				boxplot prepared={
					upper whisker=0.80,
					upper quartile=0.52,
					median=0.22,
					lower quartile=-0.53,
					lower whisker=-0.90
				},
				color=black
				] coordinates {};
				\addplot[
				fill,
				fill opacity=0.2,
				boxplot prepared={
					upper whisker=0.93,
					upper quartile=0.858,
					median=0.721,
					lower quartile=0.544,
					lower whisker=-0.62
				},
				color=black
				] coordinates {};
				\addplot[
				fill,
				fill opacity=0.2,
				boxplot prepared={
					lower whisker=0.949,
					lower quartile=0.484,
					median=0.116,
					upper quartile=-0.283,
					upper whisker=-0.924
				},
				color=susielblue
				] coordinates {};
				\addplot[
				fill,
				fill opacity=0.2,
				boxplot prepared={
					lower whisker=0.934,
					lower quartile=0.826,
					median=0.718,
					upper quartile=0.419,
					upper whisker=-0.823
				},
				color=susielblue
				] coordinates {};
				\addplot[
				fill,
				fill opacity=0.2,
				boxplot prepared={
					lower whisker=0.916,
					lower quartile=0.454,
					median=0.08,
					upper quartile=-0.49,
					upper whisker=-0.91
				},
				color=susieldarkorchid
				] coordinates {};
				\addplot[
				fill,
				fill opacity=0.2,
				boxplot prepared={
					lower whisker=0.93,
					lower quartile=0.80,
					median=0.69,
					upper quartile=0.39,
					upper whisker=-0.73
				},
				color=susieldarkorchid
				] coordinates {};
				\addplot[
				fill,
				fill opacity=0.2,
				boxplot prepared={
					lower whisker=0.78,
					lower quartile=0.38,
					median=0.16,
					upper quartile=-0.2,
					upper whisker=-0.9
				},
				color=susielhibiscus
				] coordinates {};
				\addplot[
				fill,
				fill opacity=0.2,
				boxplot prepared={
					lower whisker=0.95,
					lower quartile=0.79,
					median=0.46,
					upper quartile=0.01,
					upper whisker=-0.86
				},
				color=susielhibiscus
				] coordinates {};
				\addplot[
				fill,
				fill opacity=0.2,
				boxplot prepared={
					lower whisker=0.95,
					lower quartile=0.52,
					median=0.25,
					upper quartile=-0.27,
					upper whisker=-0.88
				},
				color=susielsalmon
				] coordinates {};
				\addplot[
				fill,
				fill opacity=0.2,
				boxplot prepared={
					lower whisker=0.92,
					lower quartile=0.82,
					median=0.51,
					upper quartile=0.27,
					upper whisker=-0.75
				},
				color=susielsalmon
				] coordinates {};
				\addplot[
				fill,
				fill opacity=0.2,
				boxplot prepared={
					lower whisker=0.94,
					lower quartile=0.67,
					median=0.46,
					upper quartile=0.03,
					upper whisker=-0.88
				},
				color=susielamber,
				] coordinates {};
				\addplot[
				fill,
				fill opacity=0.2,
				color=susielamber,
				boxplot prepared={
					lower whisker=0.9,
					lower quartile=0.64,
					median=0.35,
					upper quartile=0.14,
					upper whisker=-0.71
				}
				] coordinates {};
			\end{axis}
		\end{tikzpicture}
		\caption{On DAVIS 2017~\cite{Pont-Tuset_arXiv_2017}}
	\end{subfigure}
	\begin{subfigure}{0.49\textwidth}	
		\begin{tikzpicture}
			\begin{axis}
				[
				xtick={1, 2, 3, 4, 5, 6, 7, 8, 9, 10, 11, 12, 13},
				xticklabels={
					QAM,
					Q-$S$, Q-$S_{\mathcal{R}}$,
					X-$S$, X-$S_{\mathcal{R}}$, 
					E-$S$, E-$S_{\mathcal{R}}$,
					E-$V$, E-$V_{\mathcal{R}}$, 
					M-$S$, M-$S_{\mathcal{R}}$,
					M-$V$, M-$V_{\mathcal{R}}$},
				x tick label style={rotate=60, anchor=north east, yshift=5pt},
				boxplot/draw direction=y,
				width=6.5cm,height=4.5cm,
				ymin=-1.05, ymax=1.05,
				]
				\addplot[
				fill,
				fill opacity=0.2,
				pattern=north east lines,
				pattern color=black
				] [boxplot prepared={
					upper whisker=0.990,
					upper quartile=0.674,
					median=0.383,
					lower quartile=0.123,
					lower whisker=-0.384
				}]
				coordinates {};
				\addplot[
				fill,
				fill opacity=0.2,
				boxplot prepared={
					upper whisker=0.63,
					upper quartile=0.12,
					median=-0.16,
					lower quartile=-0.33,
					lower whisker=-0.75
				},
				color=black
				] coordinates {};
				\addplot[
				fill,
				fill opacity=0.2,
				color=black,
				boxplot prepared={
					upper whisker=0.96,
					upper quartile=0.70,
					median=0.55,
					lower quartile=0.23,
					lower whisker=-0.36,
				},
				] coordinates {};
				\addplot[
				fill,
				fill opacity=0.2,
				color=susielblue,
				boxplot prepared={
					lower whisker=0.809,
					lower quartile=0.30,
					median=0.023,
					upper quartile=-0.33,
					upper whisker=-0.8
				},
				] coordinates {};
				\addplot[
				fill,
				fill opacity=0.2,
				color=susielblue,
				boxplot prepared={
					lower whisker=0.922,
					lower quartile=0.682,
					median=0.568,
					upper quartile=0.33,
					upper whisker=-0.0504
				},
				] coordinates {};
				\addplot[
				fill,
				fill opacity=0.2,
				color=susieldarkorchid,
				boxplot prepared={
					lower whisker=0.779,
					lower quartile=0.358,
					median=0.101,
					upper quartile=-0.11,
					upper whisker=-0.80
				},
				] coordinates {};
				\addplot[
				fill,
				fill opacity=0.2,
				color=susieldarkorchid,
				boxplot prepared={
				lower whisker=0.83,
				lower quartile=0.72,
				median=0.47,
				upper quartile=0.23,
				upper whisker=-0.31
				},
				] coordinates {};
				\addplot[
				fill,
				fill opacity=0.2,
				color=susielhibiscus,
				boxplot prepared={
					lower whisker=0.85,
					lower quartile=0.51,
					median=0.19,
					upper quartile=-0.06,
					upper whisker=-0.70
				},
				] coordinates {};
				\addplot[
				fill,
				fill opacity=0.2,
				color=susielhibiscus,
				boxplot prepared={
					lower whisker=0.88,
					lower quartile=0.65,
					median=0.39,
					upper quartile=0.15,
					upper whisker=-0.28
				},
				] coordinates {};
				\addplot[
				fill,
				fill opacity=0.2,
				color=susielsalmon,
				boxplot prepared={
					lower whisker=0.71,
					lower quartile=0.36,
					median=-0.01,
					upper quartile=-0.36,
					upper whisker=-0.9
				},
				] coordinates {};
				\addplot[
				fill,
				fill opacity=0.2,
				color=susielsalmon,
				boxplot prepared={
					lower whisker=0.93,
					lower quartile=0.54,
					median=0.2,
					upper quartile=-0.2,
					upper whisker=-0.7
				},
				] coordinates {};
				\addplot[
				fill,
				fill opacity=0.2,
				color=susielamber,
				boxplot prepared={
					lower whisker=0.74,
					lower quartile=0.30,
					median=0.07,
					upper quartile=-0.26,
					upper whisker=-0.98
				},
				] coordinates {};
				\addplot[
				fill,
				fill opacity=0.2,
				color=susielamber,
				boxplot prepared={
					lower whisker=0.89,
					lower quartile=0.49,
					median=0.2,
					upper quartile=-0.1,
					upper whisker=-0.64
				},
				] coordinates {};
			\end{axis}
		\end{tikzpicture}
		\caption{On LVOS~\cite{Hong_2023_ICCV}}
	\end{subfigure}
	\caption{Comparison of correlation coefficients across the DAVIS 2017~\cite{Pont-Tuset_arXiv_2017} and LVOS~\cite{Hong_2023_ICCV} datasets: i) the QAM module~\cite{QDMN} and entropy based QDMN~\cite{QDMN} (as Q-$S$ and Q-$S_{\mathcal{R}}$), ii) entropy results for a single baseline (X-$S$, X-$S_{\mathcal{R}}$), an ensemble (E-$S$, E-$S_{\mathcal{R}}$), and Monte-Carlo methods (M-$S$, M-$S_{\mathcal{R}}$), and iii) epistemic uncertainty variants for ensemble and Monte-Carlo (E-$V$, E-$V_{\mathcal{R}}$, M-$V$, M-$V_{\mathcal{R}}$).}
	\label{Coerr. Entropy Perf}
\end{figure}
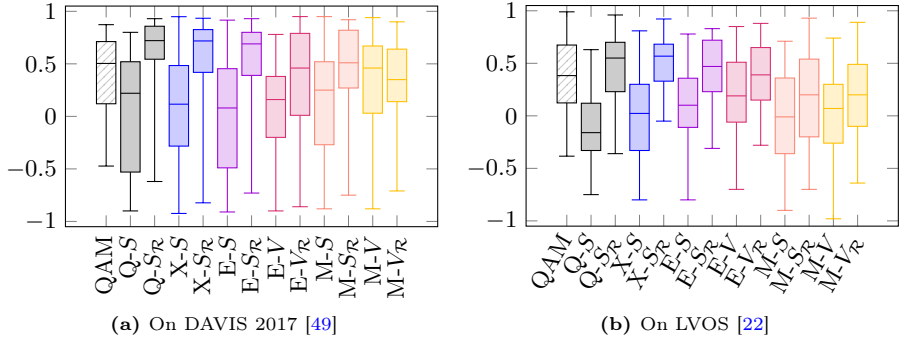

\cref{Coerr. Entropy Perf} presents the distribution (\ie, box-plots) of the correlation coefficients when computing the coefficient for every object present in a dataset.
Aside from the QAM based methods, we expect an inverse correlation, however, to facilitate the comparison, we invert the correlation results for all methods except for the QAM version.
Consequently, values closer to 1 indicate a higher correlation, suggesting a more accurate estimate of the tracking state.
Across both the DAVIS 2017~\cite{Pont-Tuset_arXiv_2017} and LVOS~\cite{Hong_2023_ICCV} datasets, variants employing masked entropy (\ie, \(S_{\mathcal{R}}\)) demonstrate notably stronger correlations.
This highlights the effectiveness of isolating uncertainty at the object level using a mask.
Among the different model variants \textendash\xspace single (Q and X), ensemble (E), and Dropout (M) \textendash\xspace the single models (Q and X) leveraging \(S_{\mathcal{R}}\) outperform even the advanced learning-based QAM module~\cite{QDMN}.

Hence, by examining \cref{Coerr. Entropy Perf}, the most effective method for estimating the tracking state on-the-fly appears to be the masked entropy approach, particularly the X-\(S_{\mathcal{R}}\) variant, as its median value is closer to 1 and the distribution is notably narrower.
This underscores the efficacy of masked entropy as a straightforward yet robust approach to estimate the tracking's state on-the-fly.

\subsection{Evaluation Framework for ziVOS}
\label{metrics_eval}
As our goal is to improve the robustness of video object segmentation methods by incorporating user corrections on-the-fly, while mimizing the user's workload, we only allow one interaction per object per frame.
Following standard practices only two form of interactions are possible, \ie, positive and negative, to indicate foreground and background regions respectively.
Moreover, we limit the type of interactions to only clicks, as pointing an object is the quickest and most intuitive interaction type for humans~\cite{clark2005coordinating,firestone2014please}.

To automatically evaluate \ac{aivos} methods, we simulate a user interaction~$u_f^o$ at frame~$f$ for object $o$ in the sequence using a \textit{simulated agent}, whenever the \ac{aivos} method requests a user correction.
We simulate a user interaction $u_f^o$ at the center of gravity of the largest misclassified region or from the ground-truth object mask (both approaches shown similar results, but for the remaining we rely on the latter approach for simplicity).

\subsection{Quantitative Results (ziVOS)}
\label{quant}
In \cref{tab:quant_interAkt_only_click}, we present quantitative results of \rred{NAME} compared to \ac{sota} methods on the LVOS validation set~\cite{Hong_2023_ICCV} by following the \ac{aivos} evaluation process outlined in~\cref{metrics_eval}.
Hence, unlike \ac{svos}, which relies on curated masks, we rely on click to indicate which object to track in the video.
As a results, we employ imperfect masks generated by the mask refiner~(\ie, SAM-HQ~\cite{sam_hq}), which more closely resembles real-world scenarios.
We provide additional results on LVOS~\cite{Hong_2023_ICCV} when following the protocol in \ac{svos} in our supplementary material.

To allow for a better comparability, we also evaluate a modified version of QDMN~\cite{QDMN}, that adopts the same design as LazyXMem for integrating user and pseudo-corrections, with the notable exception that the tracking state estimation is based on the QAM module~\cite{QDMN}.
Moreover, we evaluate an alternative approach that simply requests user corrections at random intervals throughout the sequence (denoted as Rand-\rred{}).
In addition, we introduce a variant of \rred{NAME} (denoted by \rred{NAME}$^\dagger$), which operates without user corrections to facilitate the comparison \wrt~\ac{sota}~\ac{svos} methods. 
We report the popular \JandF metric, alongside our complementary metrics (see~\cref{metrics}).

As shown in \cref{tab:quant_interAkt_only_click}, our proposed \rred{NAME}$^\dagger$ achieves competitive results \wrt to the \ac{sota}~\ac{svos} methods.
However the robustness is still close to the original XMem~\cite{XMEM} version, despite of the increase in accuracy.
By incorporating user corrections (\ie, \rred{NAME}), we manage to improve the robustness by $13$ points on average over all robustness metrics, while requesting in total $325$ interactions from the user for the entire datasets, averaging one interaction every $18.4$ seconds.
This corresponds to approximately $1.05$\% of the total number of frames in the LVOS validation set, which contains $30,876$ frames~\cite{Hong_2023_ICCV}.

While \rred{NAME} incorporates user corrections on-the-fly to enhance its robustness, it requires a continuous participation of the user throughout the segmentation process.
Therefore, we present \rred{NAME} as an alternative to~\ac{svos} and~\ac{ivos} methods to segment offline and online videos, in scenarios where user engagement is feasible and where segmenting over an extended period with high reliability is the priority.

\begin{table}[t]
	\centering
	\caption{Quantitative evaluation of \ac{aivos} and \ac{svos} methods on the LVOS validation set~\cite{Hong_2023_ICCV} following the \ac{aivos} framework. Here, we initialize each methods with an imperfect mask, in contrast to \ac{svos}, to indicate which object to segment in the sequence.}
	\label{tab:quant_interAkt_only_click}
	\begin{tabularx}{\textwidth}{l *{8}{>{\centering\arraybackslash}X}}
		\toprule
		& & \multicolumn{3}{c}{\textbf{Robustness}} & \multicolumn{3}{c}{\textbf{User-Workload}} \\
		\cmidrule(lr){3-5} \cmidrule(lr){6-8}
		\textbf{Method} & \JandF & $R@0.1$ & $R@0.25$ & $R@0.5$ & \ac{aci} & \ac{noc} & \ac{idi} \\
		\hline
		\rowcolor{gray!40!white}
		\multicolumn{8}{c}{\textit{\ac{svos} Methods}} \\
		\hline
		QDMN~\cite{QDMN}~\refConf{ECCV 2022} 					& $44.2$ & $47.8$ & $45.5$ & $36.2$ & - & - & - \\
		XMem~\cite{XMEM}~\refConf{ECCV 2022}        			& $52.8$ & $57.0$ & $55.0$ & $49.0$ & - & - & - \\
		DEVA~\cite{cheng2023tracking}~\refConf{ICCV 2023}  		& $55.1$ & $\mathbf{63.6}$ & $\mathbf{59.3}$ & $52.4$ & - & - & - \\
		Cutie-base~\cite{cheng2023putting}~\refConf{CVPR 2024}  & $57.0$ & $59.2$ & $57.8$ & $52.4$ & - & - & - \\
		Cutie-small~\cite{cheng2023putting}~\refConf{CVPR 2024} & $\mathbf{57.6}$ & $58.6$ & $57.0$ & $\mathbf{52.5}$ & - & - & - \\
		\sbul{\rred{NAME}$^\dagger$~\refConf{ours}}				& $56.4$ & $58.8$ & $56.8$ & $50.6$ & - & - & - \\
		\hline
		\rowcolor{gray!40!white}
		\multicolumn{8}{c}{\textit{\ac{aivos} Methods}} \\
		\hline
		\sbul{Rand-\rred{}}	& $61.3$ & $67.9$ & $65.8$ & $59.3$ & $5.17$ & $335$ & $17.9$ \\
		\sbul{Lazy-QDMN}	& $52.7$ & $58.2$ & $52.0$ & $42.9$ & $5.64$ & $360$ & $16.7$ \\
		\sbul{\rred{}~\refConf{ours}}	& $\mathbf{64.3}$ & $\mathbf{70.2}$ & $\mathbf{67.8}$ & $\mathbf{62.3}$ & $5.02$ & $\mathbf{325}$ & $18.4$ \\
		\bottomrule
	\end{tabularx}
\end{table}

\subsection{Ablations}
\label{ablation}

To provide more insights into our pipeline, we detail the influence of each design choice in \cref{tab:ablations}.
Using the Uncertainty Driven Update (UDU), we achieve improvements over the baseline by selectively integrating memory predictions that present sufficiently low uncertainty.
By soliciting user interactions to refine the initial mask predicted by the \ac{svos} baseline (i.e., XMem~\cite{XMEM}), we achieve slight improvements at the cost of $507$ interactions across the dataset.
While, storing the refined masks as references for future segmentation after a user correction through the Interaction Driven Update (IDU), we attain substantial improvements in both robustness and user workload.
However, using the original mask from XMem~\cite{XMEM}, associated with a high uncertainty, as an additional prompt to the user's interaction for the mask refiner leads to a decrease in performance.

By generating pseudo-interactions following the strategy outlined in~\cref{interAKT} to refine XMem's initial mask, we enhance the robustness even further while slightly reducing the user's workload.
However, saving the resulting refined mask from a pseudo-interaction (pseudo-IDU), affect only marginally the robustness, but increases the user workload considerably.
When discarding the user interactions and only relying on pseudo-corrections, \ie, \rred{NAME}$^\dagger$, we obtain a similar setup to \ac{svos} methods and manage to improve the results of the XMem~\cite{XMEM} baseline, even attain competitive results against the current~\ac{sota}~\ac{svos} methods as shown in~\cref{tab:quant_interAkt_only_click}.
Thus, our extension improves the baseline by: (1) discarding non-confident predictions from being added to the memory; (2) issuing pseudo-corrections to prompt SAM-HQ~\cite{sam_hq}, thereby refining the baseline's initial prediction when the method's uncertainty increases sharply (\cref{interAKT}); (3) and requests user-corrections on-the-fly as needed to improve the robustness.

\begin{table}[t]
	\centering
	\caption{Ablation study for \rred{} on the \ac{aivos} framework.
		We initialize each method with an imperfect mask, to indicate which object to segment in the sequence.}
	\begin{tabularx}{\textwidth}{c *{12}{>{\centering\arraybackslash}X}}
		\toprule
		\multicolumn{5}{c}{\textbf{Configuration}} & & & & & & & \\ 
		\cmidrule(lr){0-4}
		&\multicolumn{2}{c}{\textit{Pseudo}} & \multicolumn{2}{c}{\textit{User}} & &\multicolumn{3}{c}{\textbf{Robustness}} & \multicolumn{3}{c}{\textbf{User-Workload}}  \\
		\cmidrule(lr){2-3} \cmidrule(lr){4-5} \cmidrule(lr){7-9} \cmidrule(lr){10-12}
		UDU & Corr. & IDU & Corr. & IDU & \JandF & $R@0.1$ & $R@0.25$ & $R@0.5$ & \ac{aci} & \ac{noc} & \ac{idi} \\
		\midrule 
		- & - & - & - & - & $52.8$ & $57.0$ & $55.0$ & $49.0$ & - & - & - \\
		\checked & - & - & - & - & $54.7$ & $56.3$ & $54.5$ & $50.0$ & - & - & - \\ 
		\checked& \checked& - & - & - & $56.4$ & $58.8$ & $56.8$ & $50.6$ & - & - & - \\
		\checked& \checked& \checked & - & - &$53.1$ & $57.0$ & $55.1$ & $49.6$ & - & - & - \\
		\midrule
		\checked& - & - &\checked& - & $55.6$ & $58.2$ & $56.4$ & $51.8$ & $7.80$ & $507$ & $12.6$ \\
		\checked& - & - &\checked& \checked & $62.9$ & $67.8$ & $66.2$ & $60.9$ & $5.05$ & $327$& $18.3$\\
		\checked& \checked& - &\checked& \checked & $\mathbf{64.3}$ & $\mathbf{70.2}$ & $67.8$ & $\mathbf{62.3}$ & $5.02$ & $\mathbf{325}$ & $18.4$ \\
		\checked& \checked& \checked &\checked& \checked & $\mathbf{64.3}$ & $70.1$ & $\mathbf{68.2}$ & $62.1$ & $5.91$ & $352$ & $17.3$ \\
		\bottomrule
	\end{tabularx}
	\label{tab:ablations}
\end{table}

\section{Conclusion}
\label{sec:conclusion}
We introduce \rred{} as a reference for future work tackling \ac{aivos}, a hybrid combination of \ac{svos} and \ac{ivos}, that emulates a human-in-the-loop process for online video segmentation.
Through \rred{}, we enhance the robustness (i.e., the ratio of frames segmented above a certain \ac{iou} threshold) \wrt the \ac{svos} baseline (\ie, XMem~\cite{XMEM}) by integrating pseudo and user corrections on-the-fly.
However, as we solicit user corrections, we also aim to reduce the user's workload, striking a balance between performance and user engagement by requesting help only during critical events, where the method is likely to fail.
We estimate the tracking state (\ie, confidence) of the method by leveraging entropy (from information theory) and demonstrate that our proposed approach is an effective means of estimating the tracking state on-the-fly.
As we propose \ac{aivos}, we also introduce complementary metrics to the popular \JandF metrics~\cite{Perazzi_CVPR_2016}, to evaluate the robustness of our approach and the user's workload.
Our evaluation on the long-term LVOS dataset~\cite{Hong_2023_ICCV} shows that \rred{} improves the robustness relative to the baseline, albeit at the cost of additional user interactions.
Thus, we present \rred{} as an alternative to \ac{svos} and~\ac{ivos} methods to segment online video, particularly when on-the-fly corrections by a user are possible and when maintaining the tracking for an extended period is preferred over the accuracy of a method.

\bibliographystyle{splncs04}
\bibliography{bibliography}

\newpage     
\begin{center}
	\LARGE \textbf{\textit{SUPPLEMENTARY MATERIAL} \\ Strike the Balance: On-the-Fly Uncertainty based User Interactions for Long-Term Video Object Segmentation}
\end{center}

\setcounter{section}{0}
\renewcommand{\thesection}{\Alph{section}}
\counterwithin{figure}{section}
\counterwithin{table}{section}
\renewcommand{\thefigure}{S\arabic{figure}}
\renewcommand{\thetable}{S\arabic{table}}

In this supplementary document we provide additional quantitative and qualitative experiments alongside insights on the current limitation and future directions for improvements.

\section{Additional Evaluations}

\subsection{Quantitative Results (Perfect Mask Initialization)}
\label{svos_supp_quant_eval}
\cref{tab:quant_svos_init} reports the evaluation of \ac{svos} and \ac{aivos} methods on the LVOS validation set~\cite{Hong_2023_ICCV}, using ground-truth annotations to indicate which object to segment in the sequence (as in \ac{svos}).
We re-evaluated each method, and compute the robustness metric $R@\tau_{IoU}$, expect for DDMemory~\cite{Hong_2023_ICCV} as the code is unavailable at the time of writing.
Similarly to Table 1 (refer to the paper), \rred{}$^\dagger$ with only pseudo-interaction achieves competitive results to \ac{sota} \ac{svos} methods.
However, by including user interactions on-the-fly to aid \rred{}, we manage to improve the results robustness for the cost of $315$ interactions (about $1.02\%$ of the total number of frames in LVOS).

\begin{table}
	\centering
	\caption{Quantitative evaluation of \ac{svos} and \ac{aivos} methods on the LVOS validation set ~\cite{Hong_2023_ICCV}, when initialized with the ground-truth annotations (curated masks as in \ac{svos}).}
	\begin{tabularx}{\textwidth}{l *{8}{>{\centering\arraybackslash}X}}
		\toprule
		& & \multicolumn{3}{c}{\textbf{Robustness}} & \multicolumn{3}{c}{\textbf{User-Workload}} \\
		\cmidrule(lr){3-5} \cmidrule(lr){6-8}
		\textbf{Method} & \JandF & $R@0.1$ & $R@0.25$ & $R@0.5$ & ACI & NoC & IDI \\
		\hline
		\rowcolor{gray!40!white}
		\multicolumn{8}{c}{\textit{\ac{svos}}} \\
		\hline
		QDMN\cite{QDMN}~\refConf{ECCV 2022}          			& $48.2$ & $54.0$ & $50.1$ & $41.5$ & - & - & - \\
		XMem~\cite{XMEM}~\refConf{ECCV 2022}        			& $53.7$ & $54.6$ & $51.7$ & $41.3$ & - & - & - \\
		DDMemory~\cite{Hong_2023_ICCV}~\refConf{ICCV 2023}  	& $60.7$ & - & - & - & - & - & - \\
		DEVA~\cite{cheng2023tracking}~\refConf{ICCV 2023}  		& $58.2$ & $65.3$ & $62.7$ & $56.8$ & - & - & - \\
		Cutie-base~\cite{cheng2023putting}~\refConf{CVPR 2024}  & $60.3$ & $62.9$ & $62.0$ & $58.3$ & - & - & - \\
		Cutie-small~\cite{cheng2023putting}~\refConf{CVPR 2024} & $59.0$ & $61.3$ & $59.0$ & $56.5$ & - & - & - \\
		\sbul{\rred{}}~\refConf{ours} 							& $57.2$ & $60.3$ & $58.5$ & $49.6$ & - & - & - \\
		\hline
		\rowcolor{gray!40!white}
		\multicolumn{8}{c}{\textit{\ac{aivos}}} \\
		\hline
		\sbul{Rand-\rred{}}~\refConf{ours}  				& $60.3$ & $66.3$ & $64.3$ & $58.8$ & $5.05$ & $320$ & $18.2$ \\
		\sbul{\rred{}}~\refConf{ours} 				& $63.5$ & $70.0$ & $68.3$ & $63.1$ & $4.86$ & $315$ & $18.9$ \\
		\bottomrule
	\end{tabularx}
	\label{tab:quant_svos_init}
\end{table}

\subsection{Additional Ablations}

\tabref{tab:ablations_2} tabulates the results when relying directly on the masked entropy~$S_{\mathcal{R}_c}$ and its respective derivative~$\Delta S_{\mathcal{R}_c}$ as a condition to request user help.
To isolate the influence each strategy for calling the user's help, we discard the mask refiner and the pseudo interaction.
We only consider user interactions and rely directly on the ground-truth annotations to correct the model's predictions, instead of the mask refiner.
We can see in table \cref{tab:ablations_2}, that both strategies enhance the robustness and the accuracy, especially when updating the memory of the \ac{svos} baseline (XMem~\cite{XMEM}) with the refined masks through the Interaction Driven Update (IDU). However, by issuing an interaction based on the derivative~$S_{\mathcal{R}_c}$, we manage to significantly reduce the number of user calls from $787$ to $327$ calls

\begin{table}
	\centering
	\caption{Results for \rred{NAME} when requesting user corrections through $S_{\mathcal{R}_c}$ or $\Delta S_{\mathcal{R}_c}$ (note that for this table we discard the pseudo-interaction).
		We initialize each method with perfect masks. UDU denotes Uncertainty Driven Update.}
	\begin{tabularx}{\textwidth}{l *{8}{>{\centering\arraybackslash}X}}
		\toprule
		& & \multicolumn{3}{c}{\textbf{Robustness}} & \multicolumn{3}{c}{\textbf{User-Workload}} \\
		\cmidrule(lr){3-5} \cmidrule(lr){6-8}
		\textbf{Configuration} & \JandF & $R@0.1$ & $R@0.25$ & $R@0.5$ & ACI & NoC & IDI \\
		\hline
		XMem~\cite{XMEM}~\refConf{baseline}  & $53.7$ & $54.6$ & $51.7$ & $41.3$ & - & - & - \\
		\hline
		\rowcolor{gray!40!white}
		\multicolumn{8}{c}{\textit{Call user corrections based on} $S_{\mathcal{R}_c}$} \\
		\hline
		XMem + UDU		 & $54.7$ & $56.3$ & $54.5$ & $50.0$ & $56.1$ & $3647$ & $1.9$ \\ 
		XMem + UDU + IDU & $63.5$ & $67.6$ & $66.1$ & $61.7$ & $12.1$ & $787$ & $8.5$ \\ 
		\hline
		\rowcolor{gray!40!white}
		\multicolumn{8}{c}{\textit{Call user corrections based on} $\Delta S_{\mathcal{R}_c}$} \\
		\hline
		XMem + UDU		 & $55.6$ & $58.2$ & $56.4$ & $51.8$ & $7.80$ & $507$ & $12.6$ \\
		XMem + UDU + IDU & $62.9$ & $67.8$ & $66.2$ & $60.9$ & $5.05$ & $327$& $18.3$ \\
		\bottomrule 
	\end{tabularx}
	\label{tab:ablations_2}
\end{table}

\section{Implementation Details}
For our \ac{svos} baseline, we rely on the original weights provided by the authors of XMem~\cite{XMEM}, which is trained on the static and DAVIS 2017 training set~\cite{Pont-Tuset_arXiv_2017}

\textbf{Deep Ensemble variant}: We experiment with an ensemble approach that combines three XMem models
The first model is trained on the static~\cite{STCN} and DAVIS 2017 training set~\cite{Pont-Tuset_arXiv_2017}.
The second model (which we use as a baseline in \rred{}) is trained similarly to the first model but also includes the synthetic dataset BL30K~\cite{MiVOS}.
The third model is trained like the first model but with the addition of the MOSE~\cite{MOSE} dataset.

\textbf{Monte Carlo variant}: We rely on spatial pooling~\cite{tompson2015efficient} applied to the key-projection of XMem~\cite{XMEM}, with a dropout ratio of $0.2$ for our Monte Carlo Dropout variant during training, which is maintained during inference.
For more details, we refer the reader to the original paper~\cite{XMEM}.

\textbf{Thresholds}: We using the training set of the LVOS dataset~\cite{Hong_2023_ICCV} to identify the values for $\tau_u = 0.5$, $\tau_p = 0.2$ and $\tau_m = 0.8$.

\textbf{Hardware}: All experiments are performed on an Nvidia GeForce GTX 1080~Ti.

\section{Qualitative Results}
\label{qual_res}
In this section we provide qualitative results that highlight both success and failure cases whenever Lazy-XMem issues either pseudo- or user-corrections to generate a refined mask.
\cref{tab:images_S} and \cref{tab:images_F} displays success and failure cases, respectively, for generating a refined mask through pseudo-corrections.
\cref{tab:images_U_good} and \cref{tab:images_U_bad} show results when a refined mask is generated via a simulated user-correction, as described in \cref{metrics_eval}.

We indicate a ground-truth mask in \textcolor{susielgold!70!black}{yellow}, the original prediction in \textcolor{turquoise}{turquoise}, the refined mask in \textcolor{red!30!susielpeach}{orange} or \textcolor{red!30!susieldarkorchid}{purple} after a pseudo-and user-correction respectively.
We mark the location of a pseudo- or user-corrections through a yellow star \drawstar{yellow}.

For small objects, we provide a cropped version to better visualize the different predictions. In these cases, a small image of the original image is shown on the first column, surrounded by a red border.
Note that in \cref{tab:images_U_bad}, we do not display refined masks for the third, fourth and fifth rows, as Lazy-XMem missed for those instances the generation of either a user- or pseudo-corrections.

\subsection{Pseudo-Corrections}
\begin{figure}
	\centering
	\begin{tabular}{cccc}
		Ground-truth & Original Mask & Entropy & Refined Mask \\
		\includegraphics[width=0.24\textwidth]{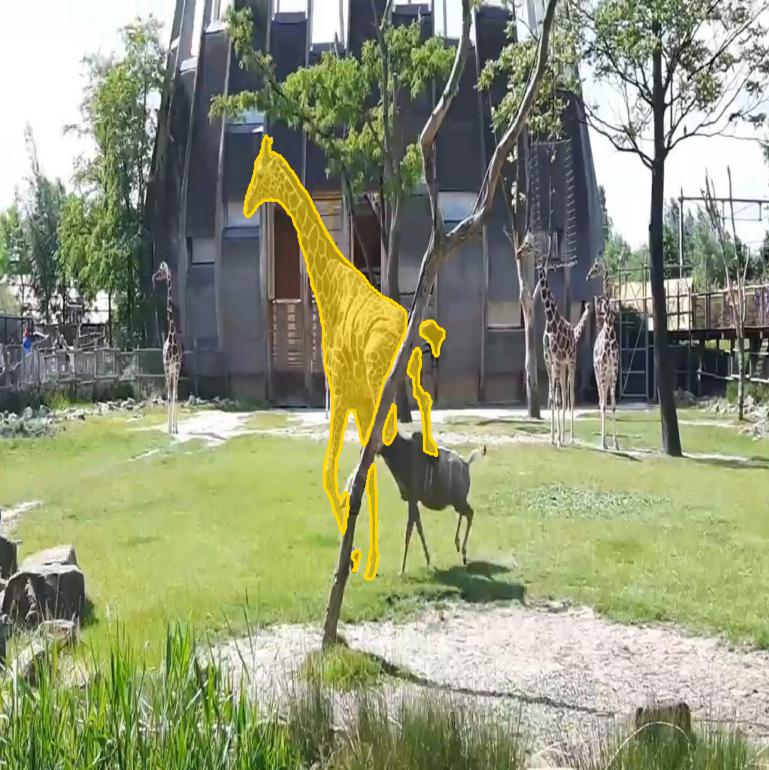} &
		\includegraphics[width=0.24\textwidth]{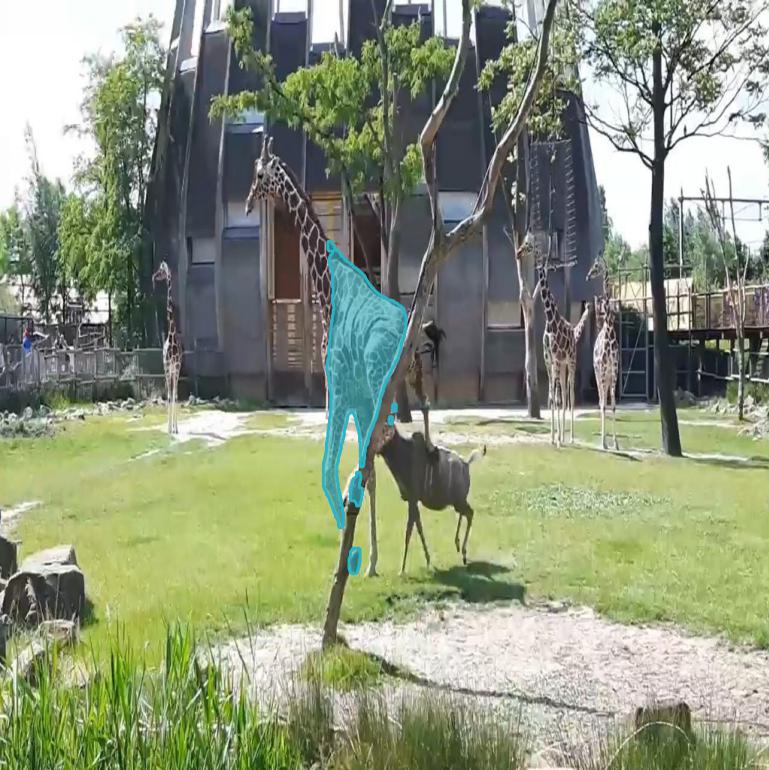} &
		\includegraphics[width=0.24\textwidth]{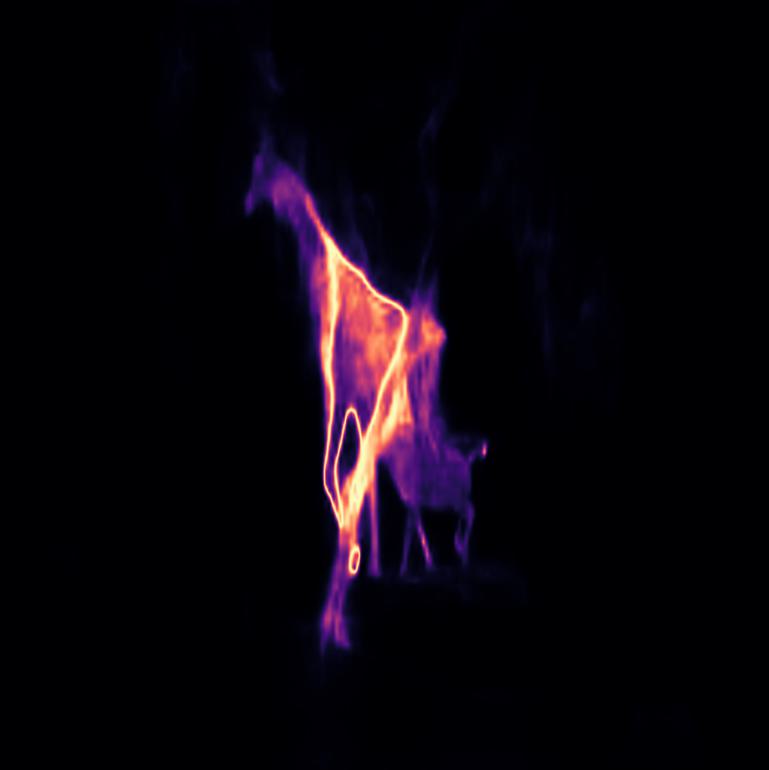} &
		\includegraphics[width=0.24\textwidth]{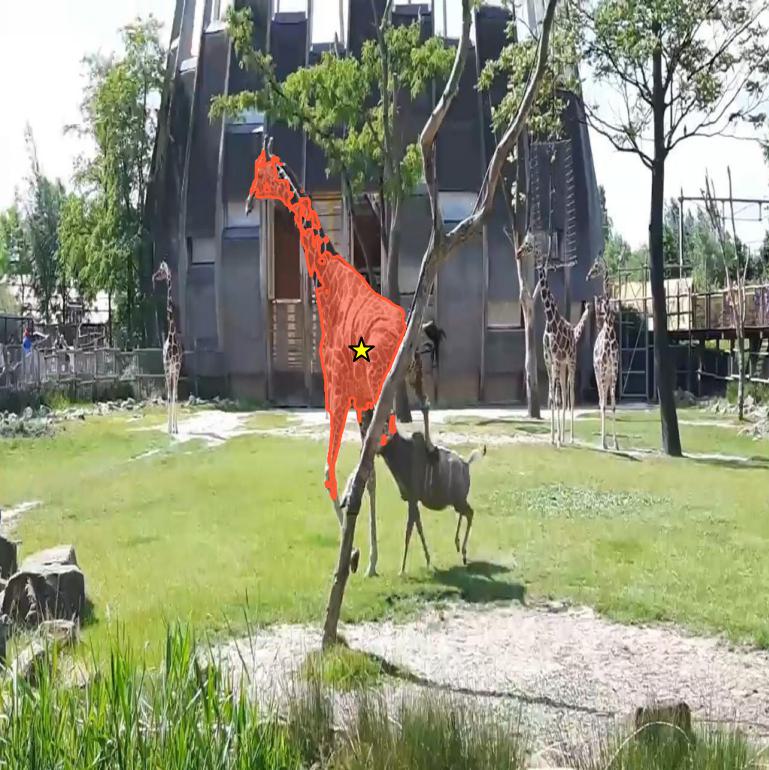} \\
		\includegraphics[width=0.24\textwidth]{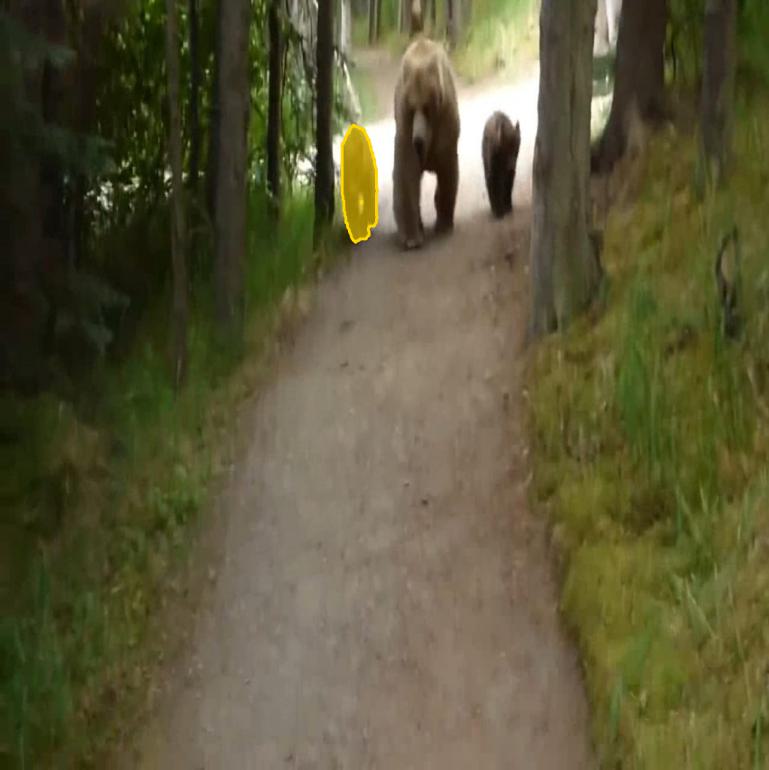} &
		\includegraphics[width=0.24\textwidth]{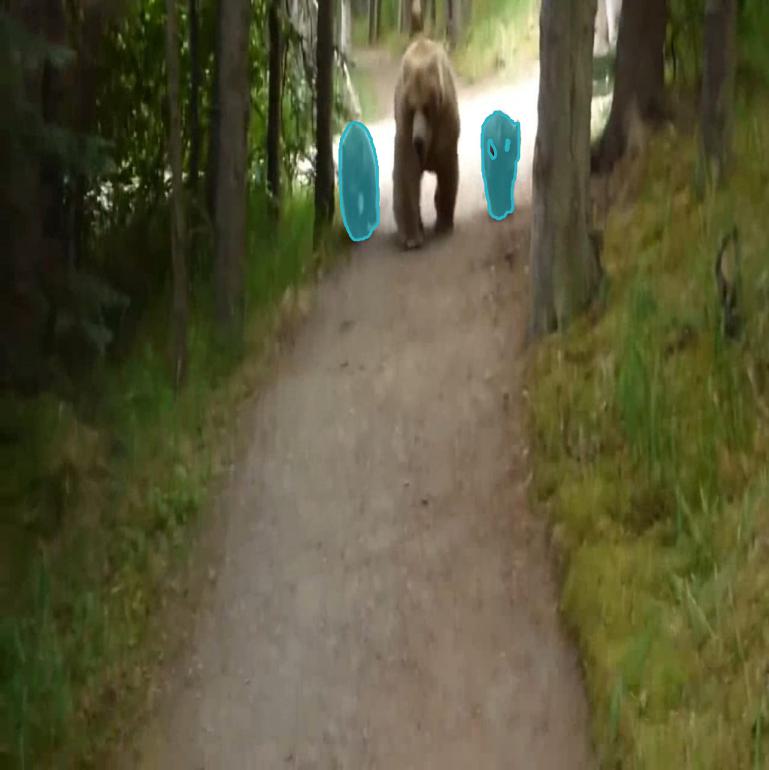} &
		\includegraphics[width=0.24\textwidth]{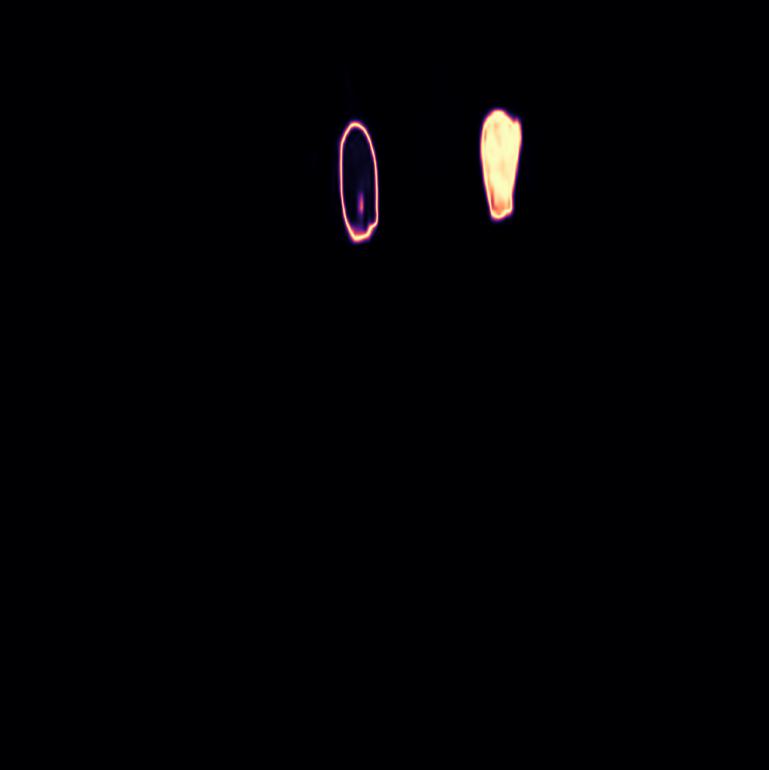} &
		\includegraphics[width=0.24\textwidth]{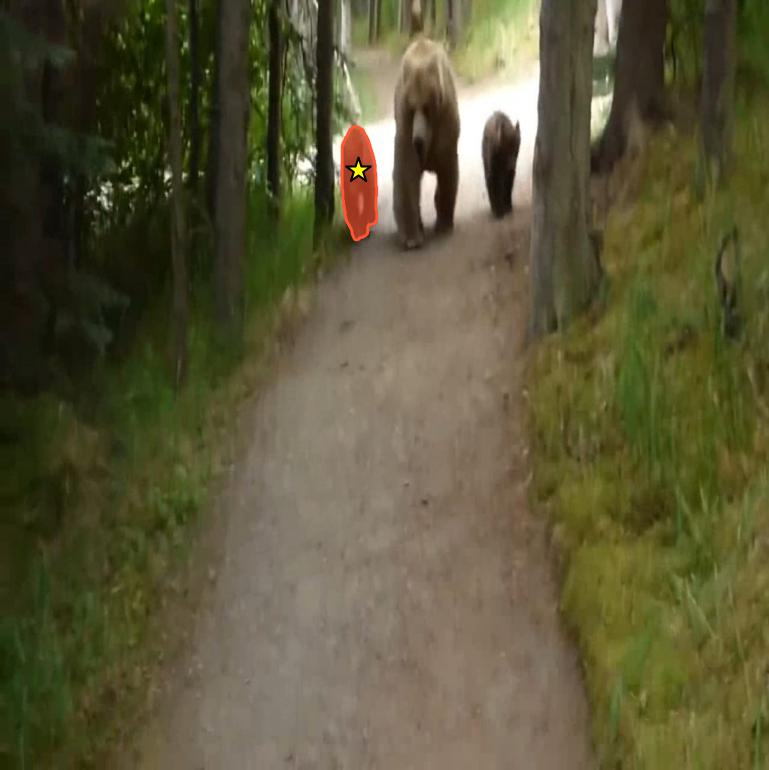} \\
		\includegraphics[width=0.24\textwidth]{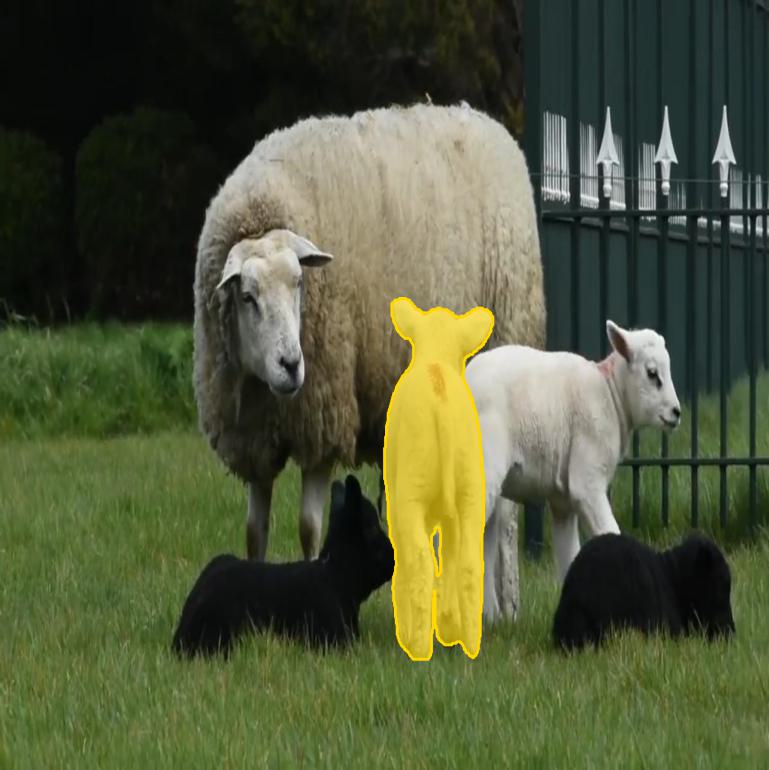} &
		\includegraphics[width=0.24\textwidth]{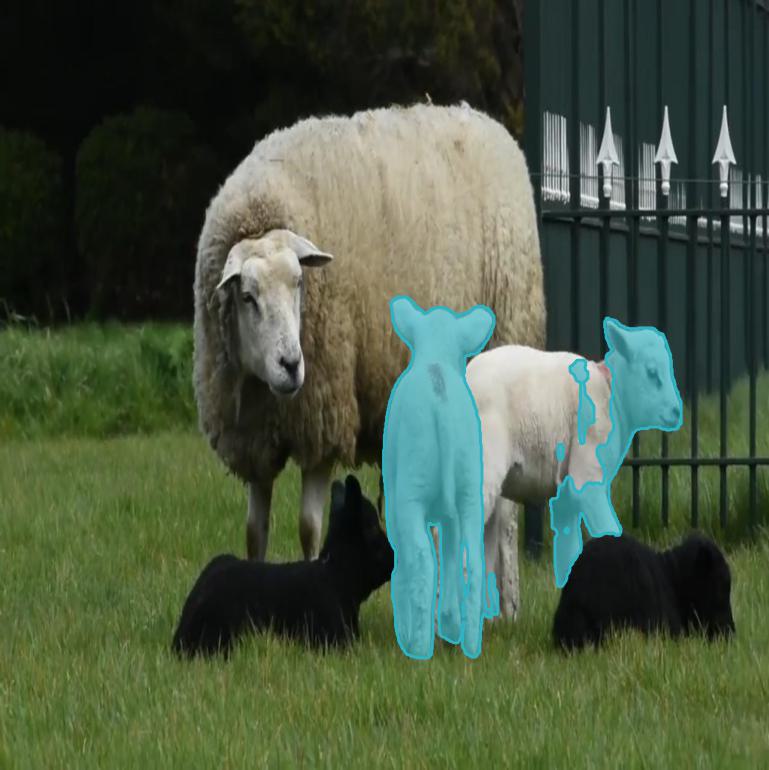} &
		\includegraphics[width=0.24\textwidth]{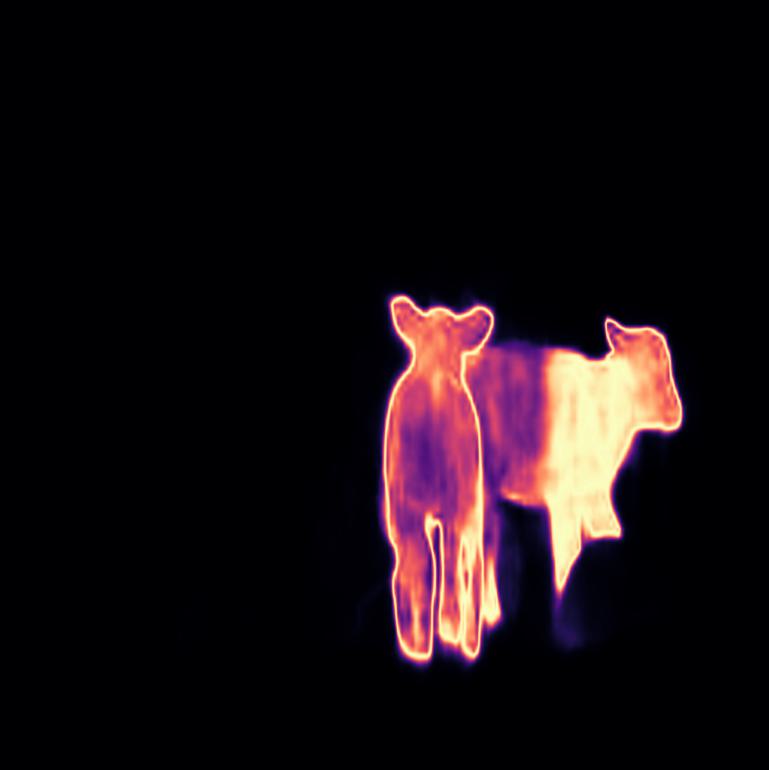} &
		\includegraphics[width=0.24\textwidth]{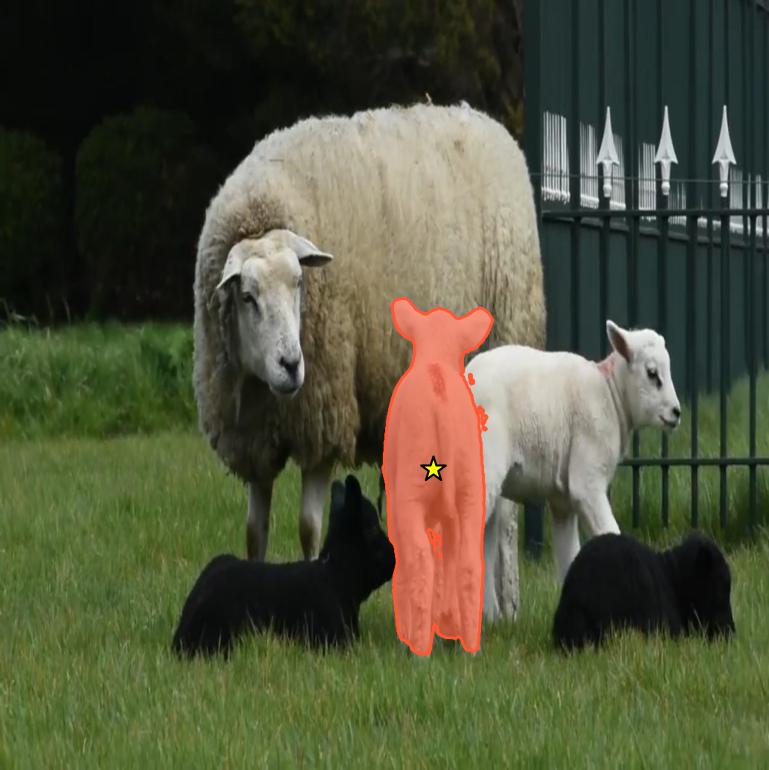} \\
		\includegraphics[width=0.24\textwidth]{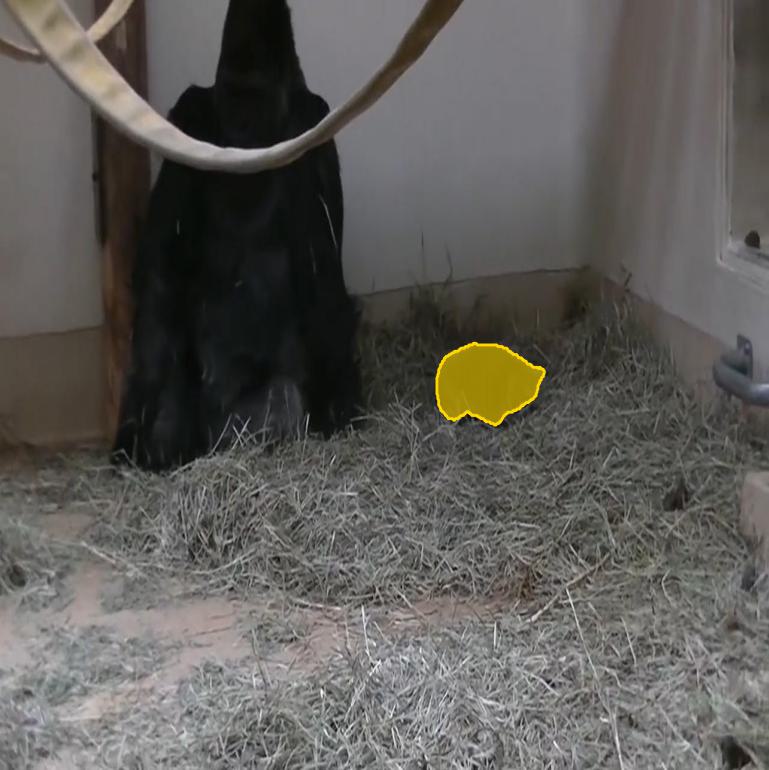} &
		\includegraphics[width=0.24\textwidth]{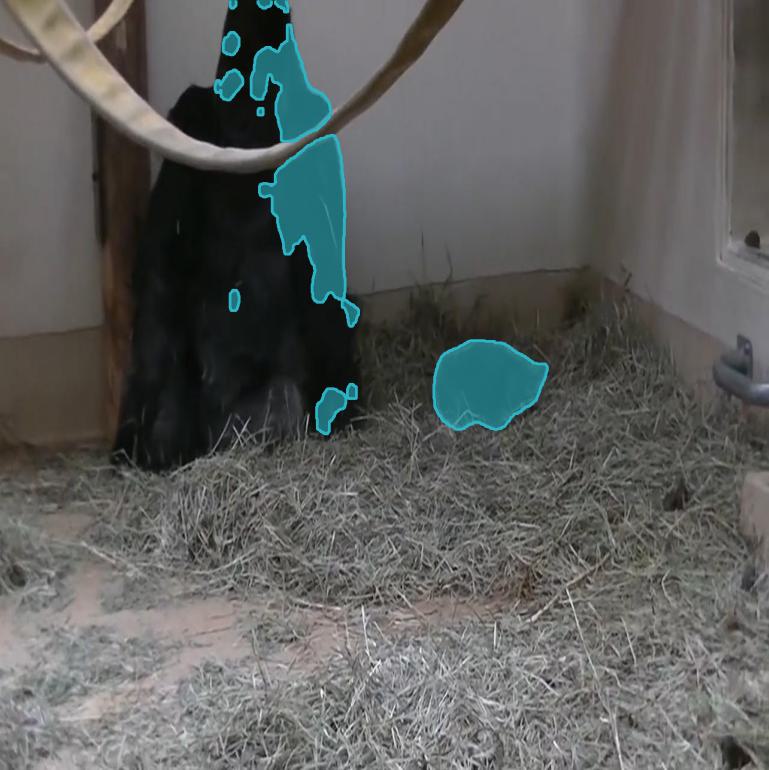} &
		\includegraphics[width=0.24\textwidth]{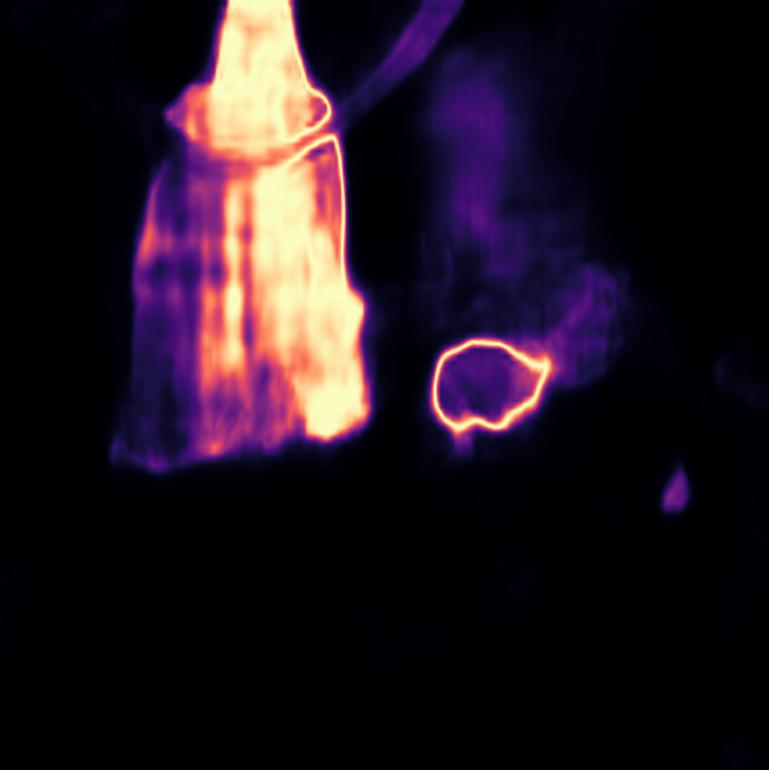} &
		\includegraphics[width=0.24\textwidth]{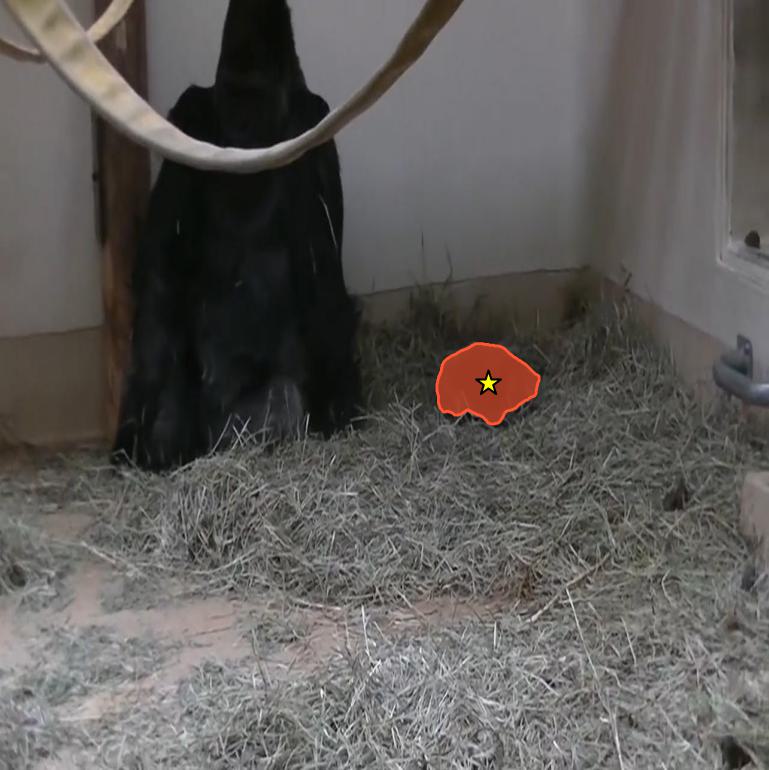} 
		\\
		\includegraphics[trim={3cm 5cm 5cm 3cm},clip, width=0.24\textwidth]{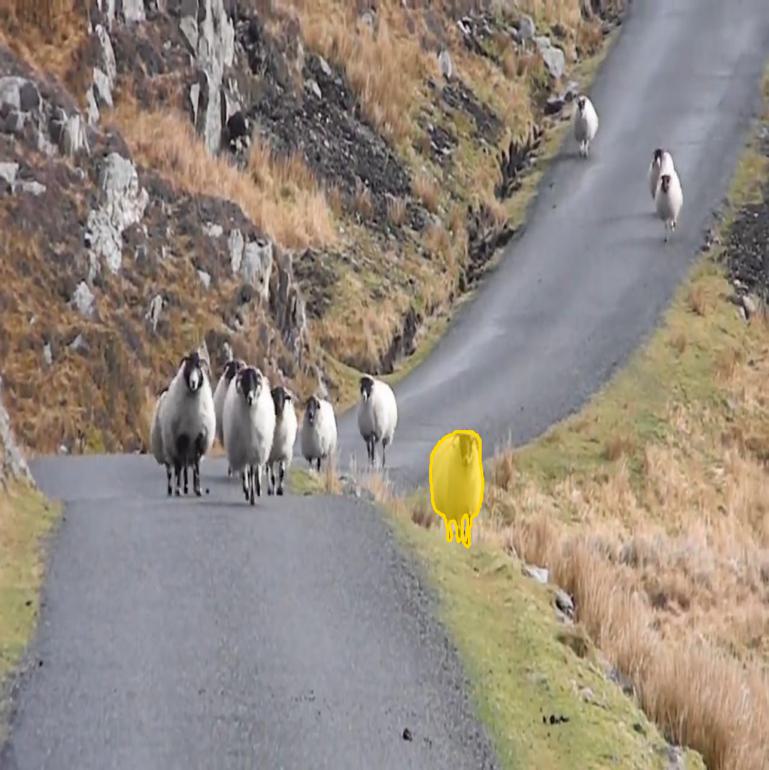} 
		\begin{tikzpicture}(0,0)
			\put(-3.05cm,1.55cm){\includegraphics[frame, cfbox=red, height=1.4cm]{D4AgqLQL/D4AgqLQL_130_1_GT.jpg}};
		\end{tikzpicture} &
		\includegraphics[trim={3cm 5cm 5cm 3cm},clip, width=0.24\textwidth]{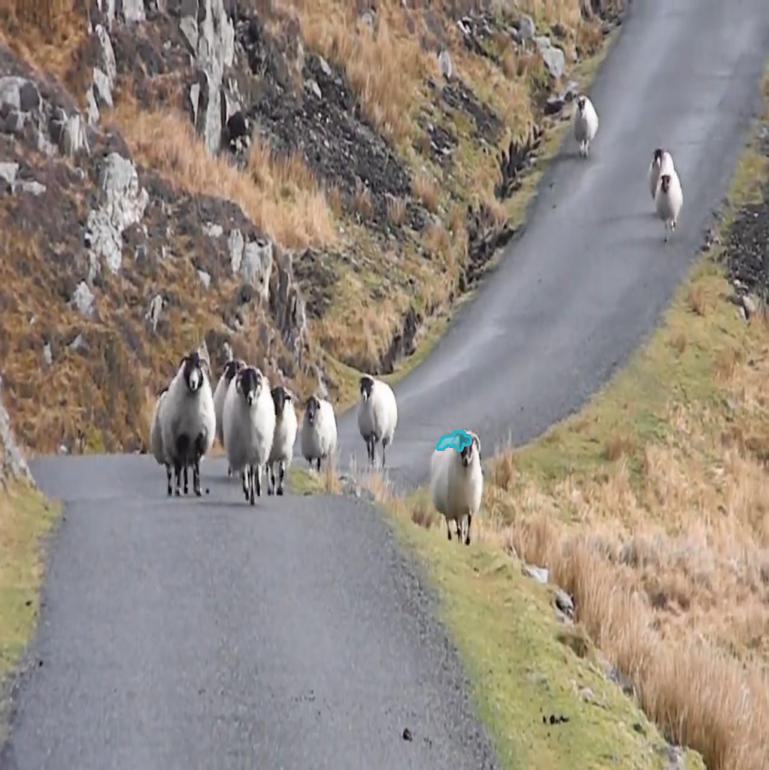} &
		\includegraphics[trim={3cm 5cm 5cm 3cm},clip, width=0.24\textwidth]{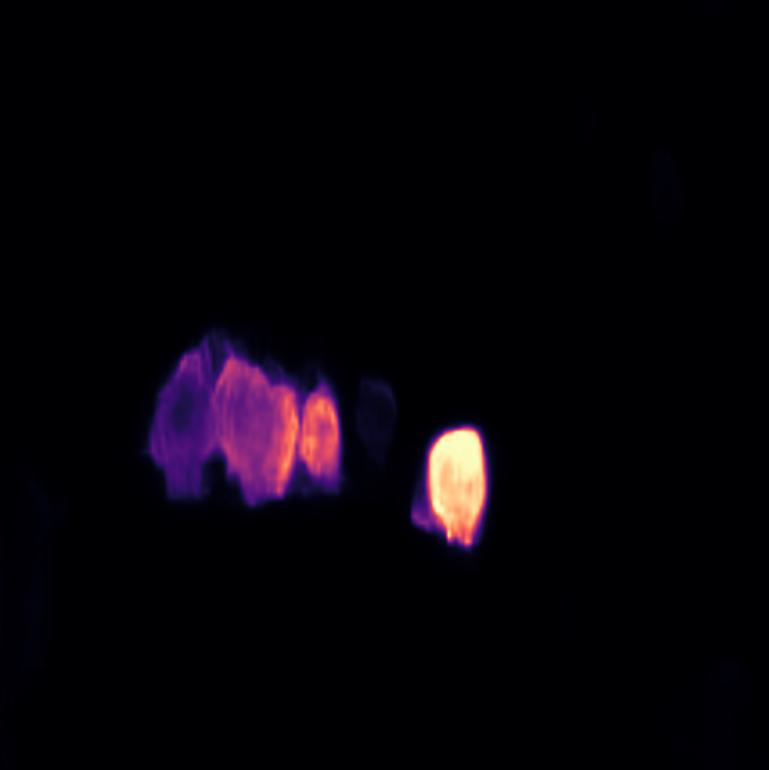} &
		\includegraphics[trim={3cm 5cm 5cm 3cm},clip, width=0.24\textwidth]{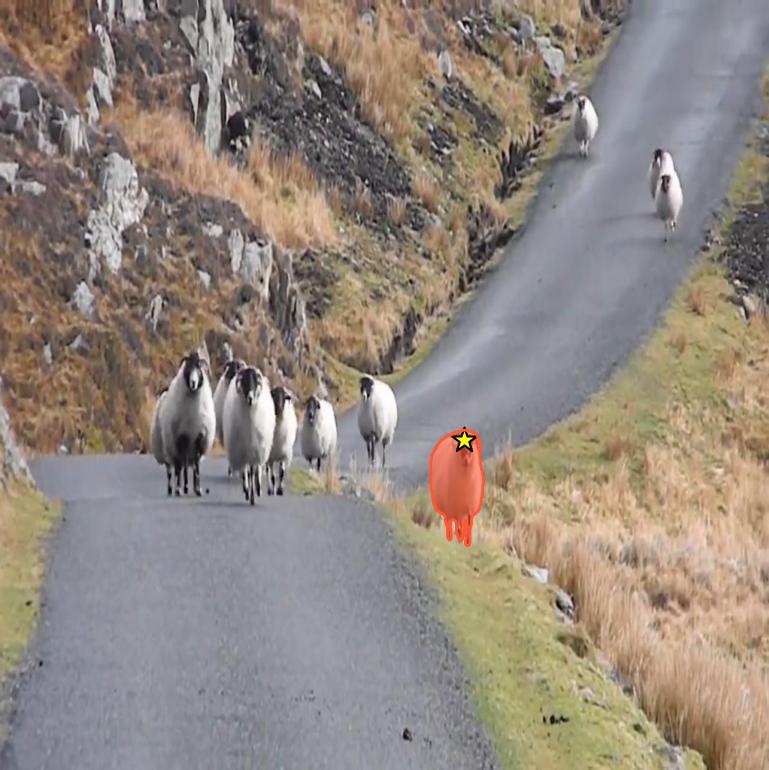}
		\\
		\includegraphics[width=0.24\textwidth]{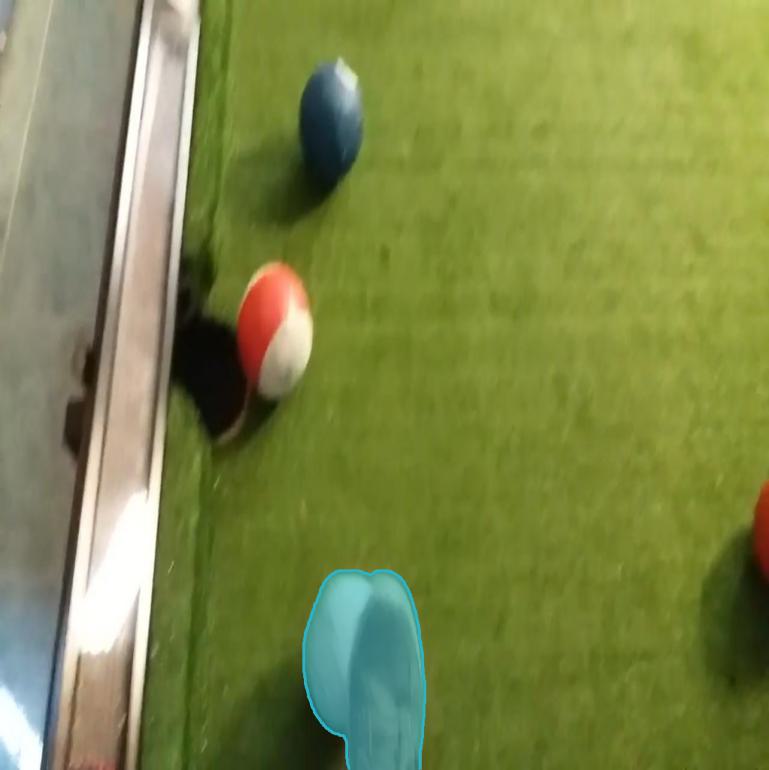} &
		\includegraphics[width=0.24\textwidth]{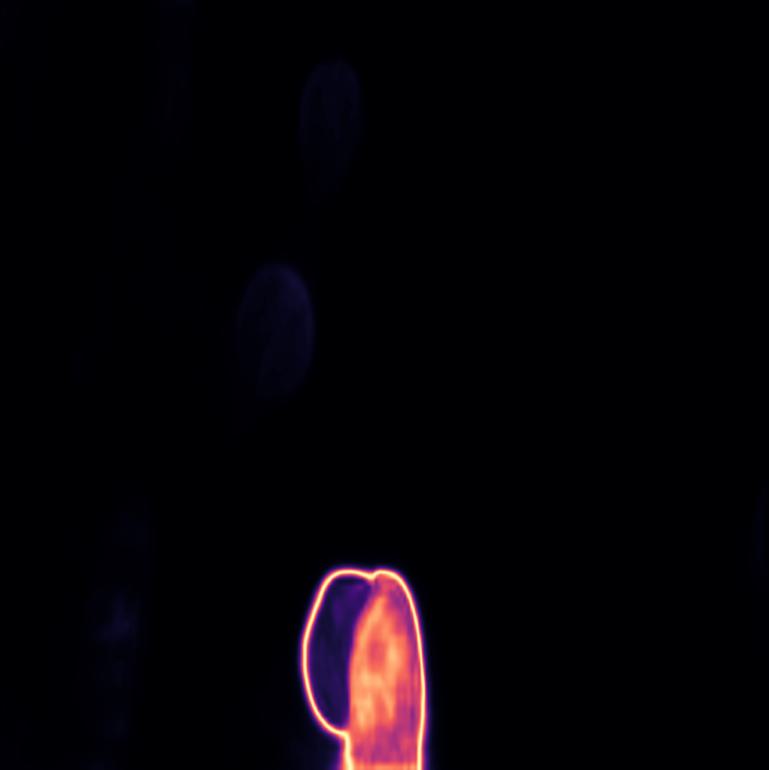} &
		\includegraphics[width=0.24\textwidth]{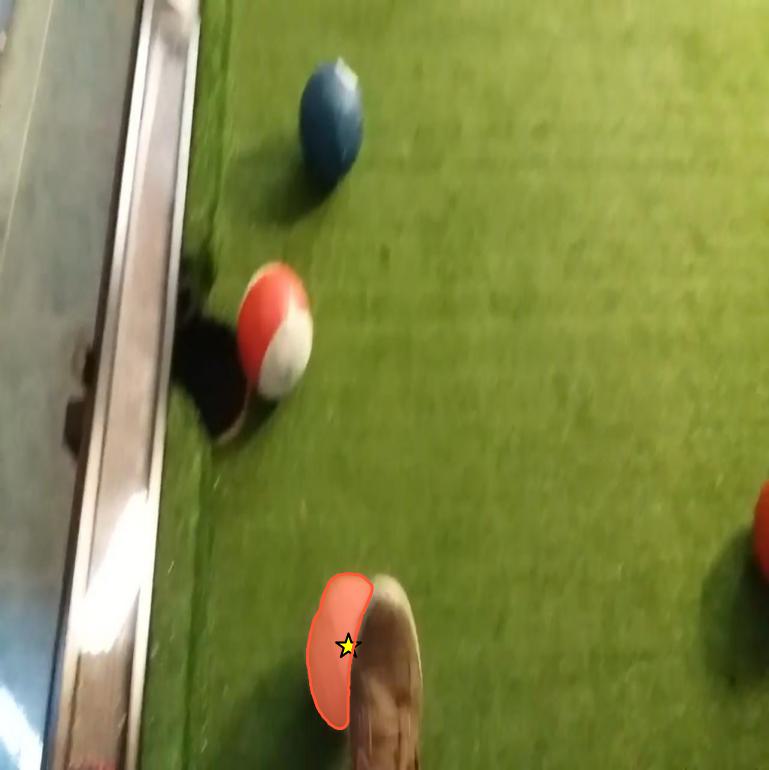}
	\end{tabular}
	\caption{Qualitative results on the validation set of LVOS~\cite{Hong_2023_ICCV} when refining the mask through pseudo-corrections (Success cases).}
	\label{tab:images_S}
\end{figure}

Through the pixel wise uncertainty estimation, we are able to identify confusing and confident regions, helpful for the generation of pseudo-corrections, allowing us to correct the segmentation whenever a distractors is present and anticipate when the method is likely to fail as shown in \cref{tab:images_S}.
We can observe that our proposed pseudo-correction generation strategy successfully recovers the original object of interest in the presence of distractors (e.g., rows two, three, and four). Additionally, objects that are about to be lost are also recovered (e.g., rows one, three, and five).

\begin{figure}
	\centering
	\begin{tabular}{cccc}
		Ground-truth & Original Mask & Entropy & Refined Mask \\
		\includegraphics[width=0.24\textwidth]{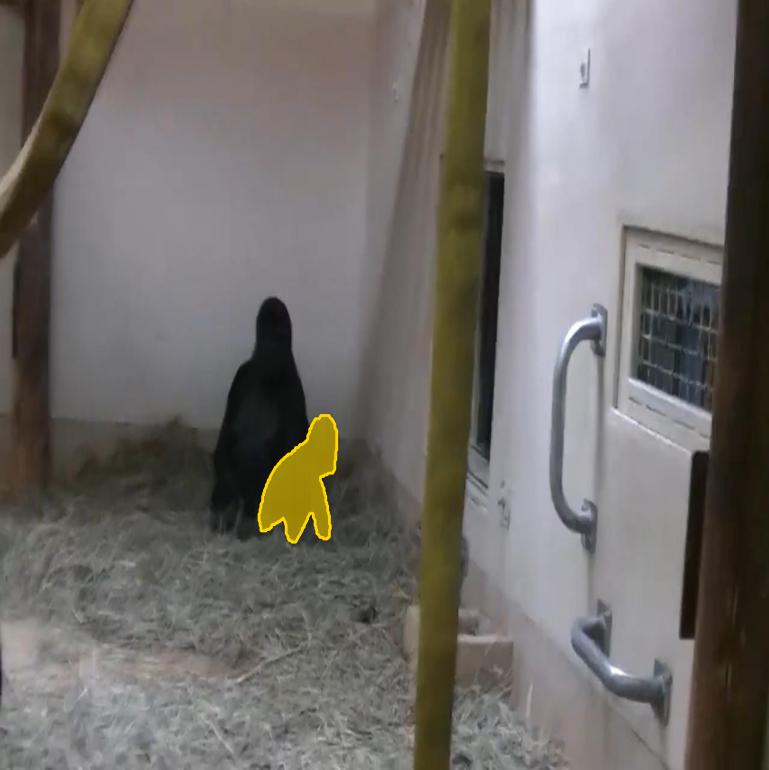} &
		\includegraphics[width=0.24\textwidth]{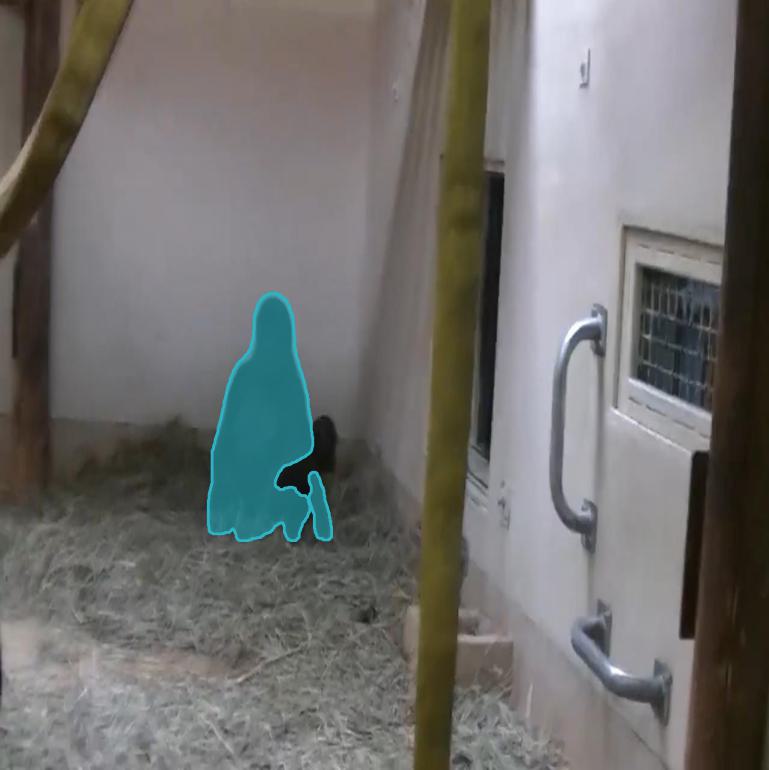} &
		\includegraphics[width=0.24\textwidth]{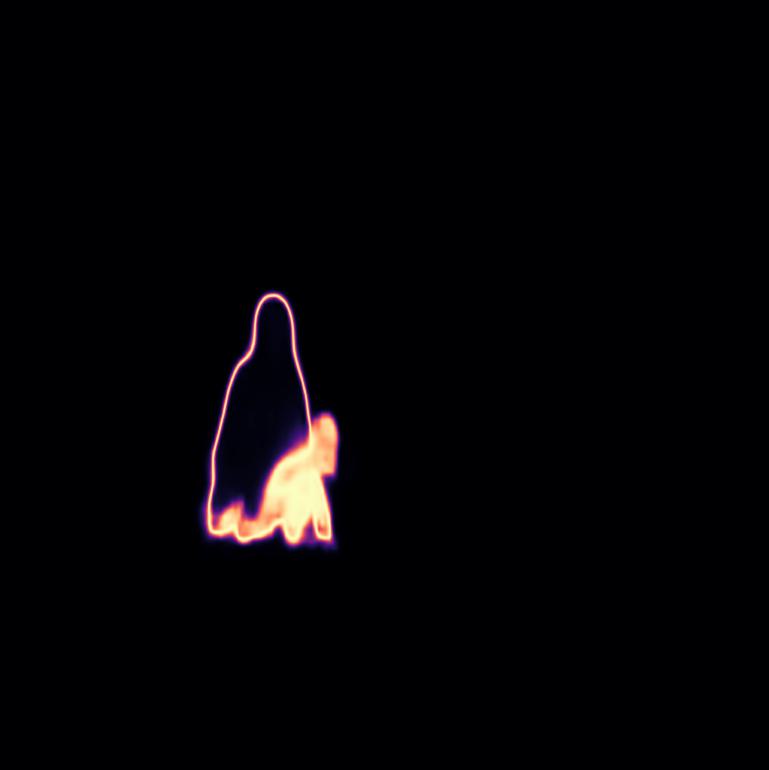} &
		\includegraphics[width=0.24\textwidth]{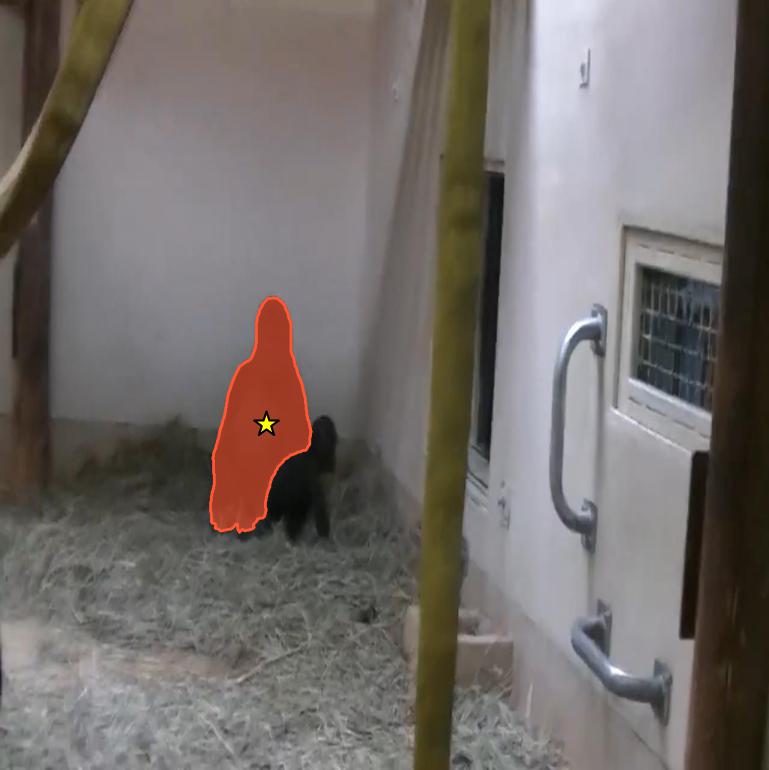}
		\\
		\includegraphics[trim={6cm 6cm 6cm 6cm},clip, width=0.24\textwidth]{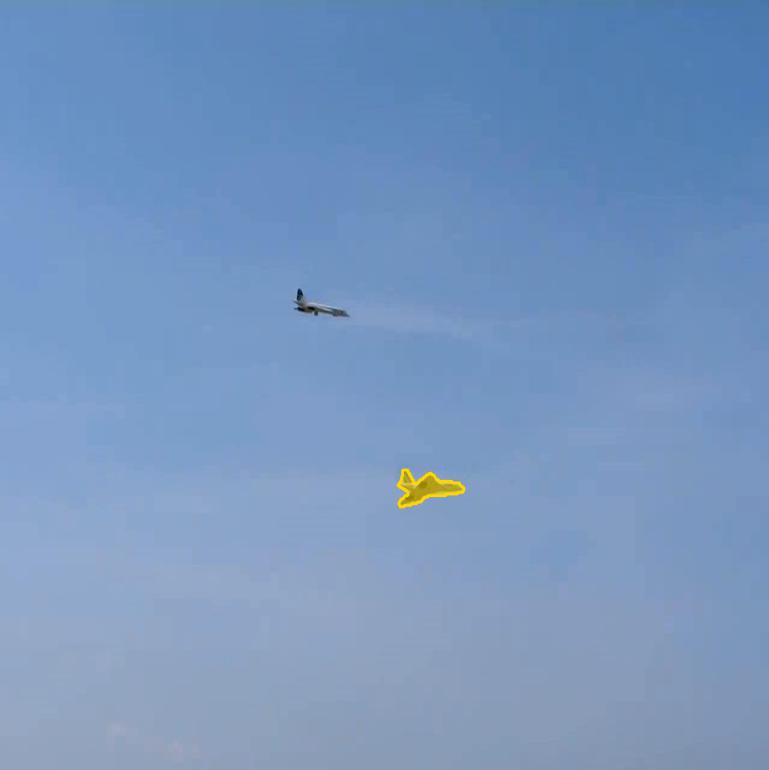} 
		\begin{tikzpicture}(0,0)
			\put(-3.05cm,0){\includegraphics[frame, cfbox=red, height=1.4cm]{3Zf4NFzn/3Zf4NFzn_359_1_GT.jpg}};
		\end{tikzpicture}
		&
		\includegraphics[trim={6cm 6cm 6cm 6cm},clip, width=0.24\textwidth]{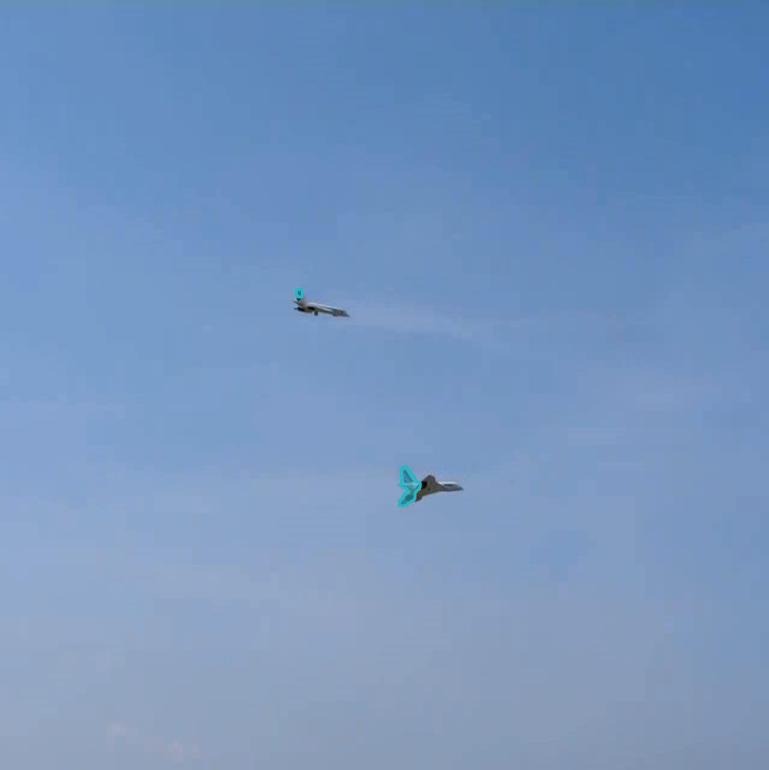}		&
		\includegraphics[trim={6cm 6cm 6cm 6cm},clip, width=0.24\textwidth]{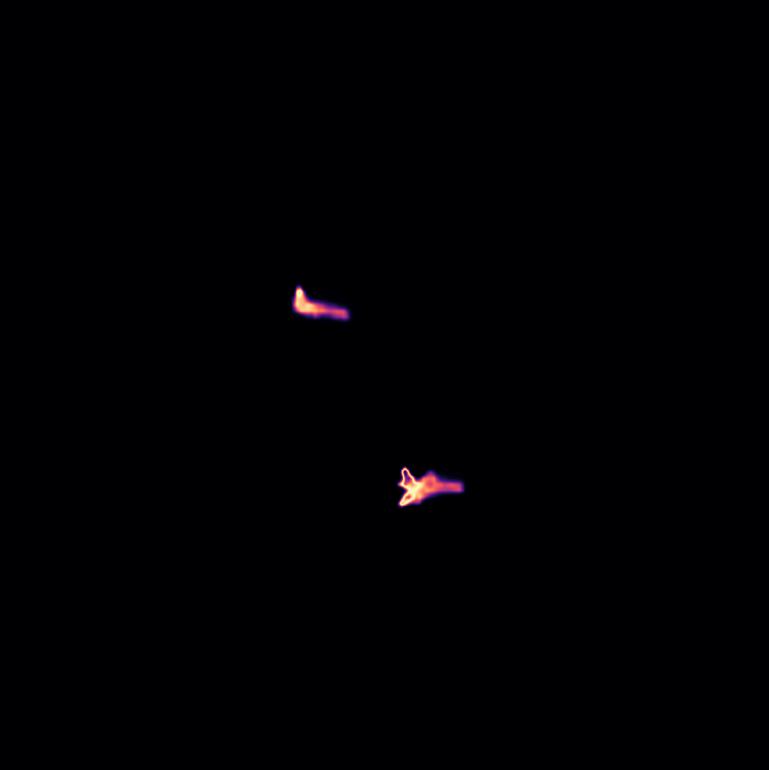}		&
		\includegraphics[trim={6cm 6cm 6cm 6cm},clip, width=0.24\textwidth]{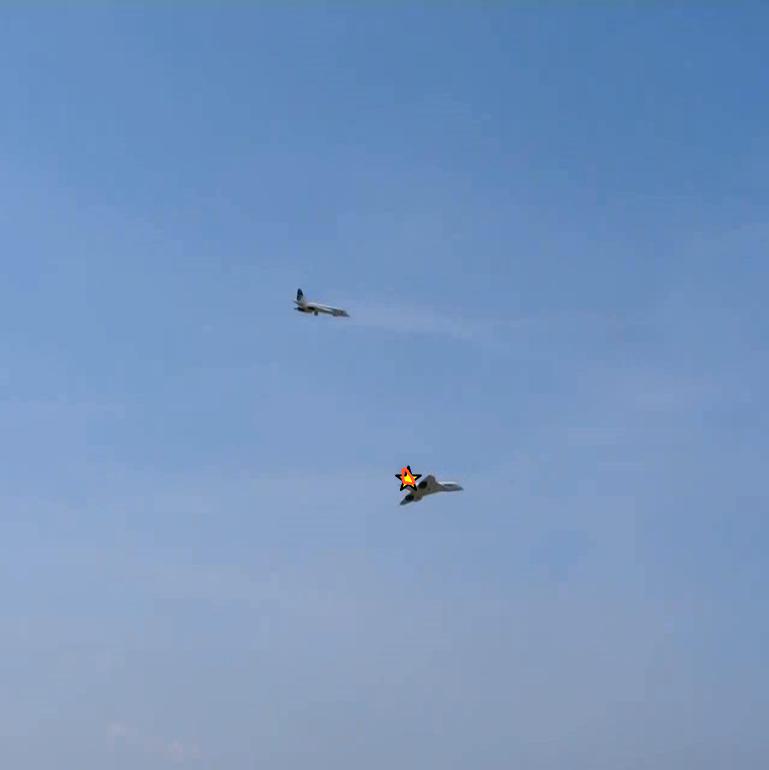}
		\\ 
		\includegraphics[width=0.24\textwidth]{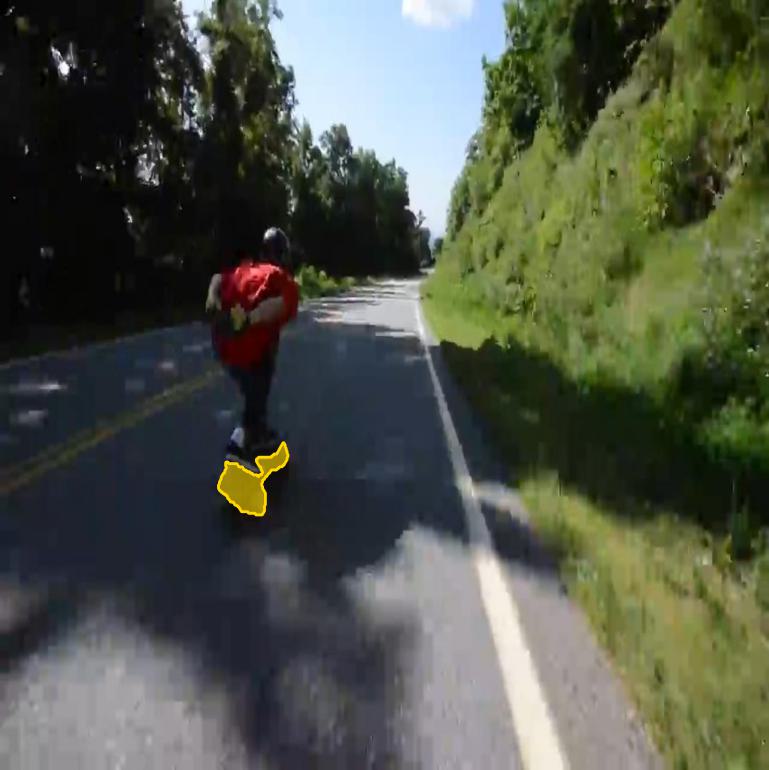} &
		\includegraphics[width=0.24\textwidth]{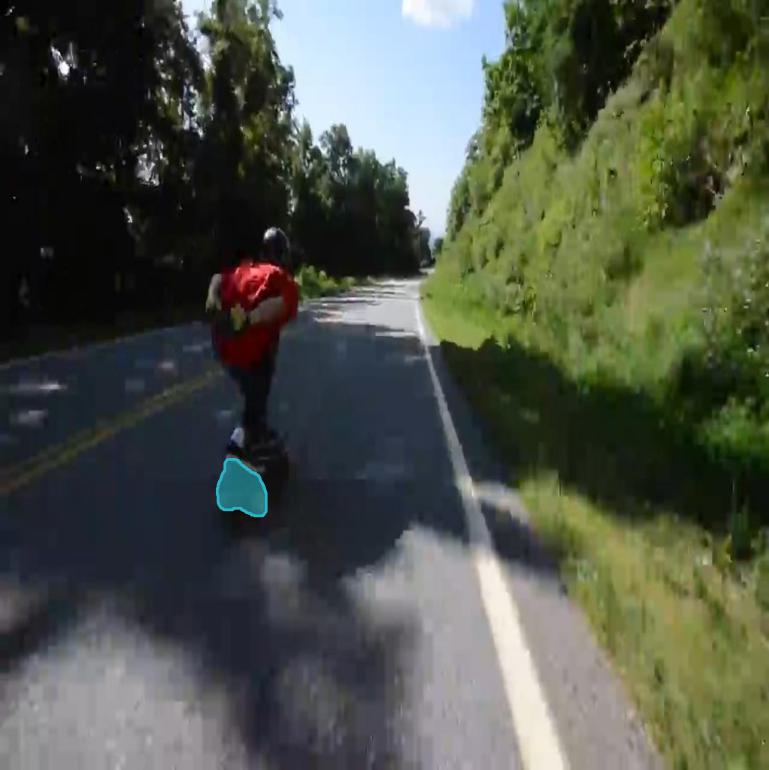} &
		\includegraphics[width=0.24\textwidth]{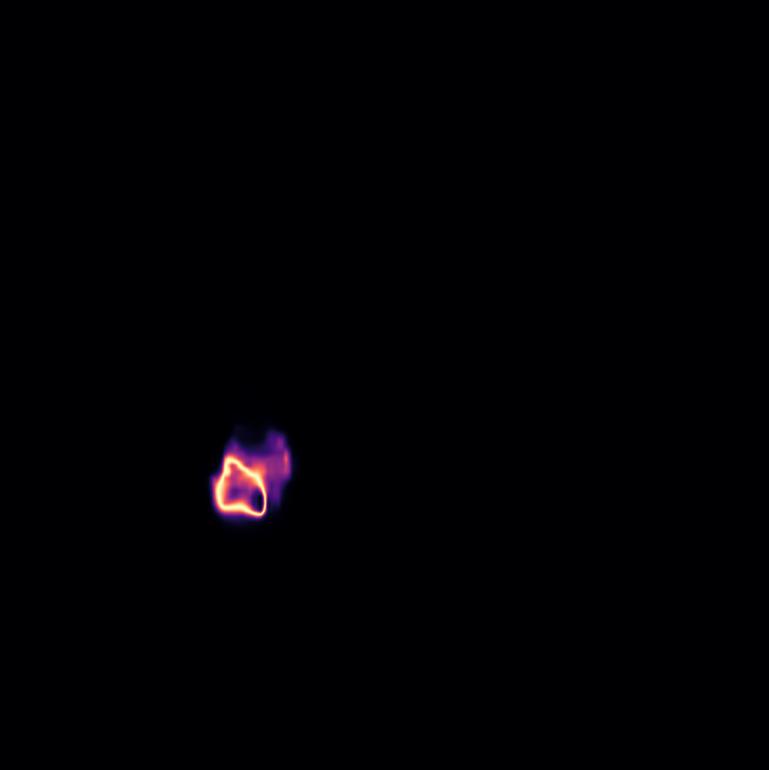} &
		\includegraphics[width=0.24\textwidth]{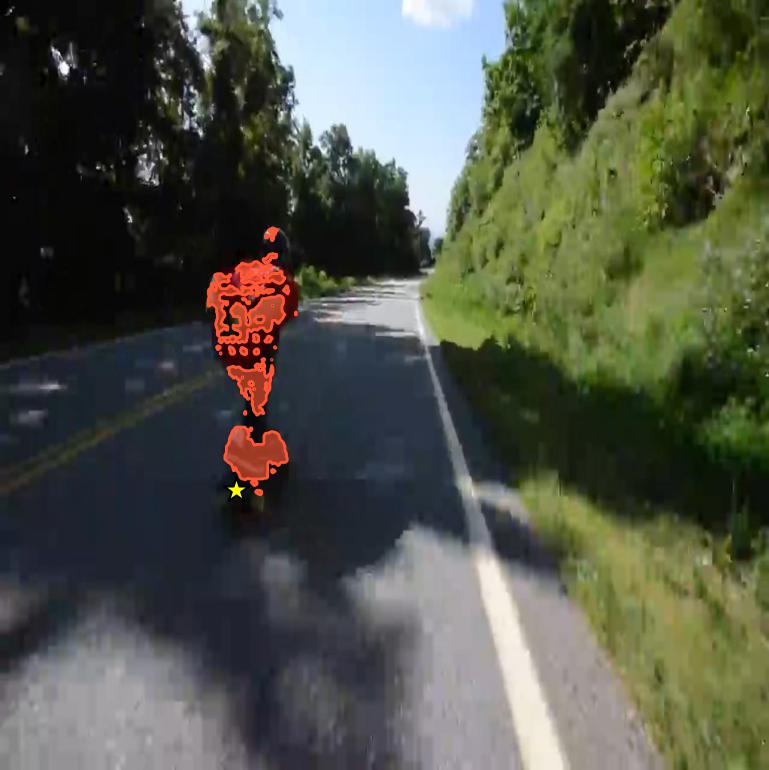}
		\\
		\includegraphics[width=0.24\textwidth]{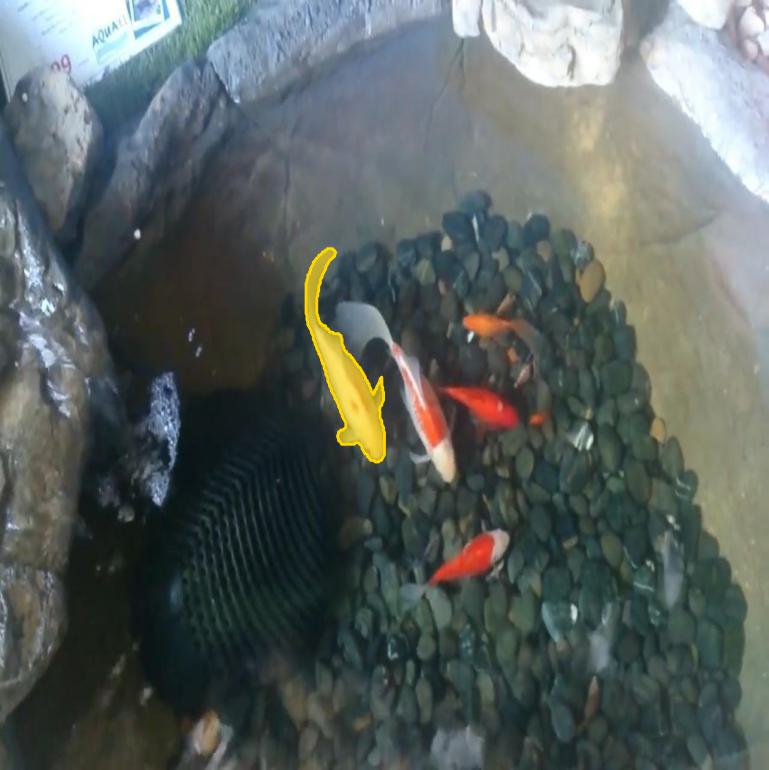} &
		\includegraphics[width=0.24\textwidth]{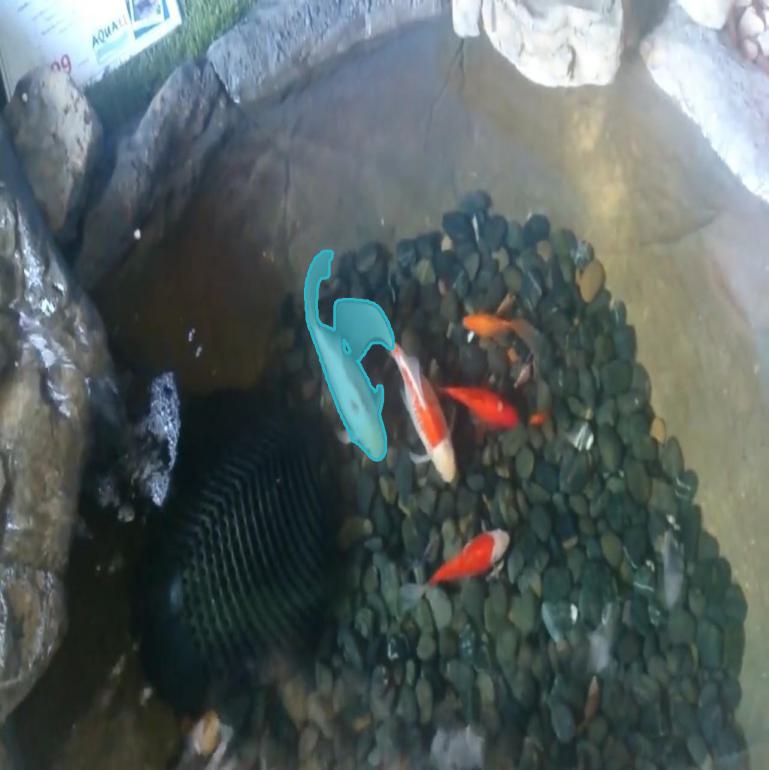} &
		\includegraphics[width=0.24\textwidth]{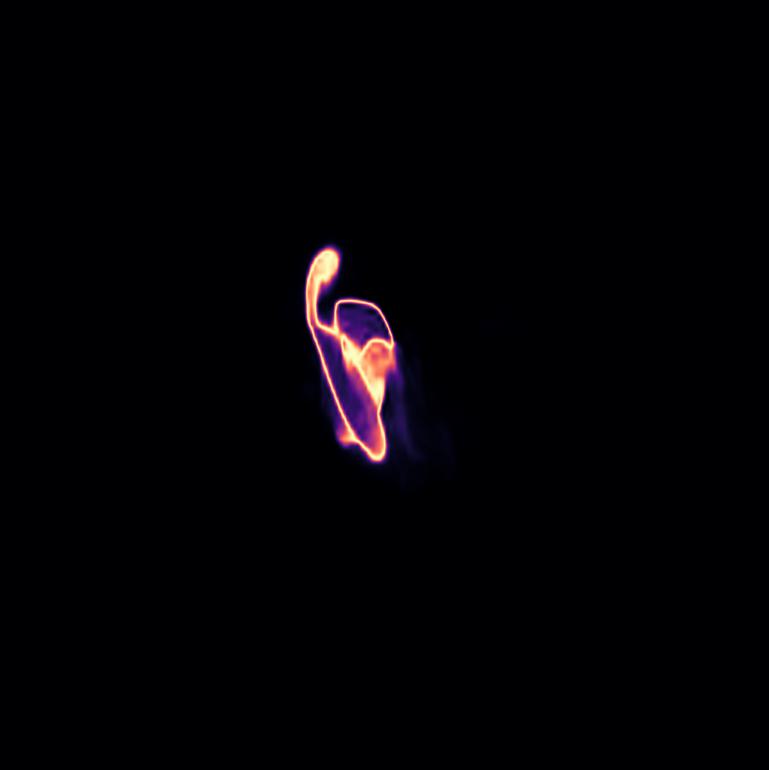} &
		\includegraphics[width=0.24\textwidth]{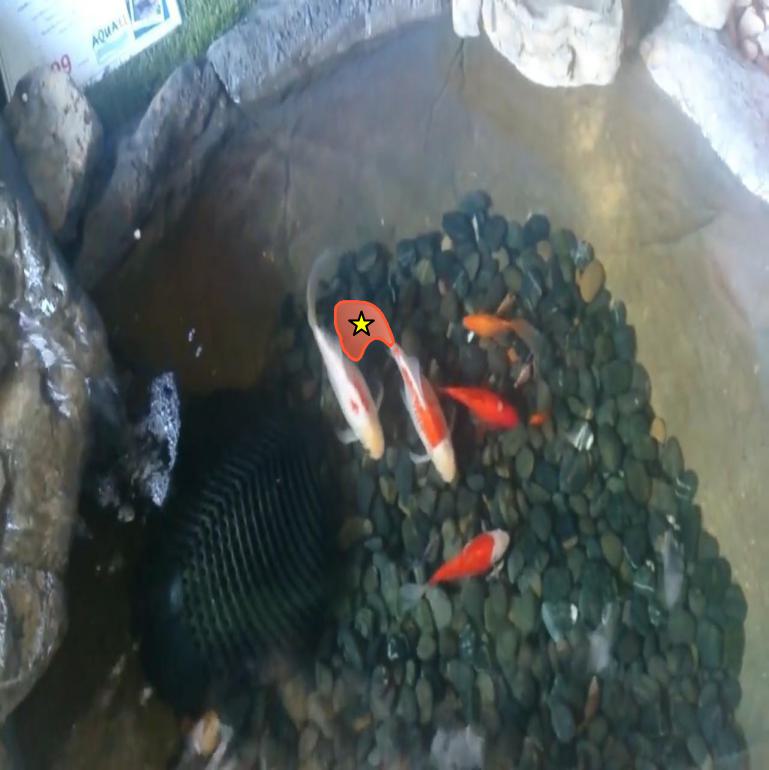}
		\\
		\includegraphics[trim={7cm 12cm 7cm 2cm},clip, width=0.24\textwidth]{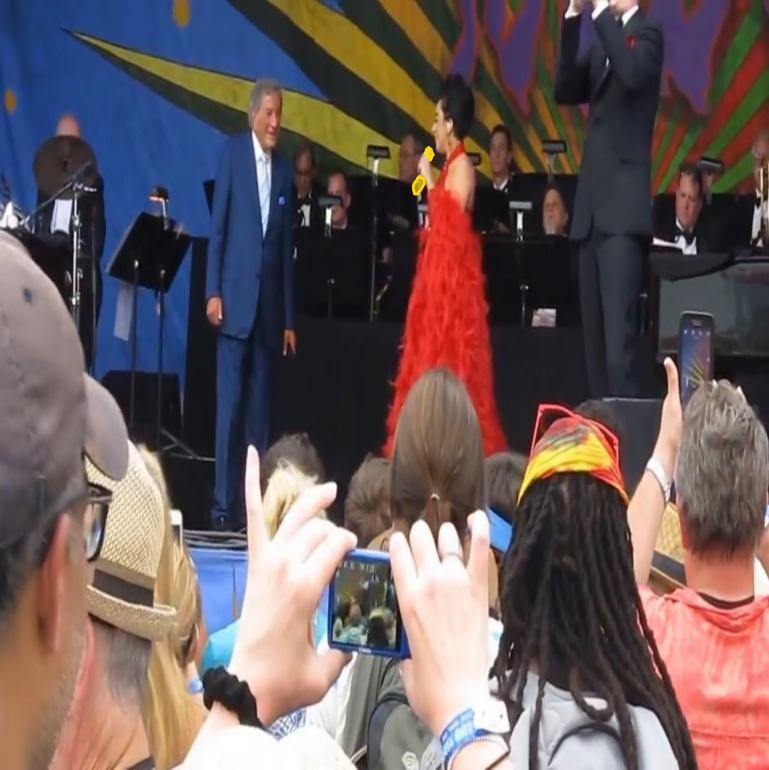} 
		\begin{tikzpicture}(0,0)
			\put(-3.05cm,0){\includegraphics[frame, cfbox=red, height=1.4cm]{K3OUeINk/K3OUeINk_40_1_GT.jpg}};
		\end{tikzpicture}
		&
		\includegraphics[trim={7cm 12cm 7cm 2cm},clip, width=0.24\textwidth]{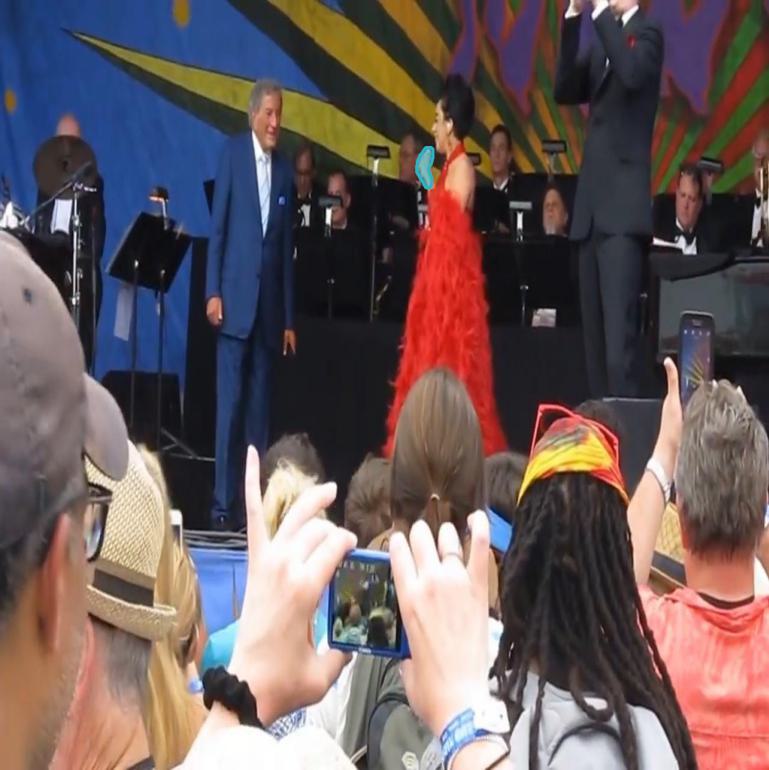} &
		\includegraphics[trim={7cm 12cm 7cm 2cm},clip, width=0.24\textwidth]{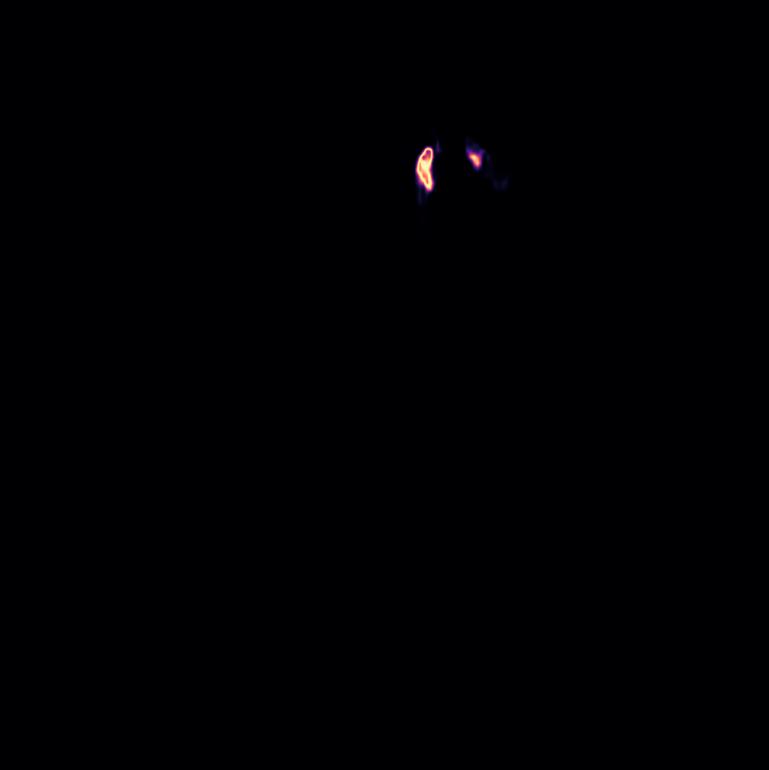} &
		\includegraphics[trim={7cm 12cm 7cm 2cm},clip, width=0.24\textwidth]{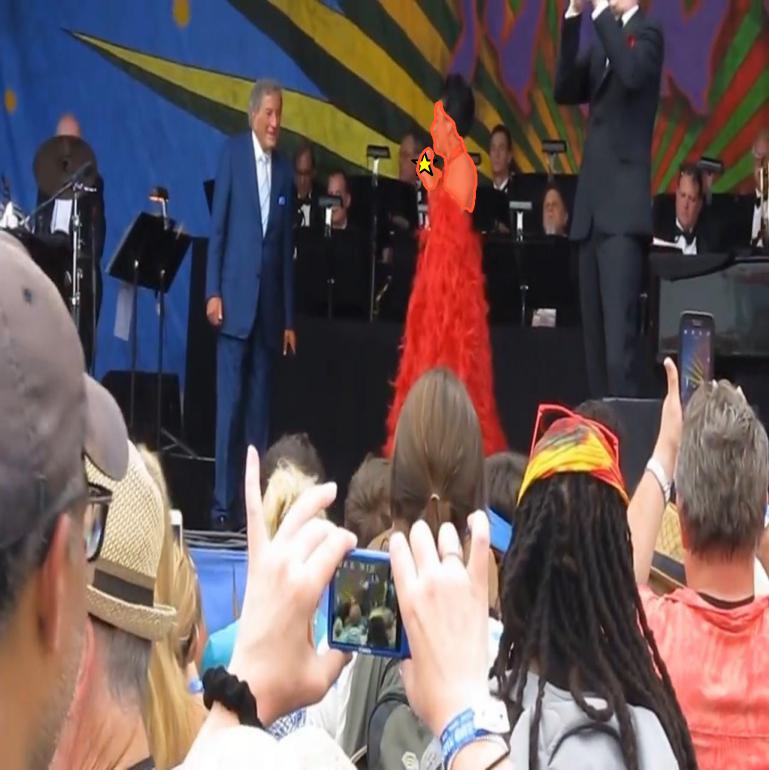}
		\\
		\includegraphics[width=0.24\textwidth]{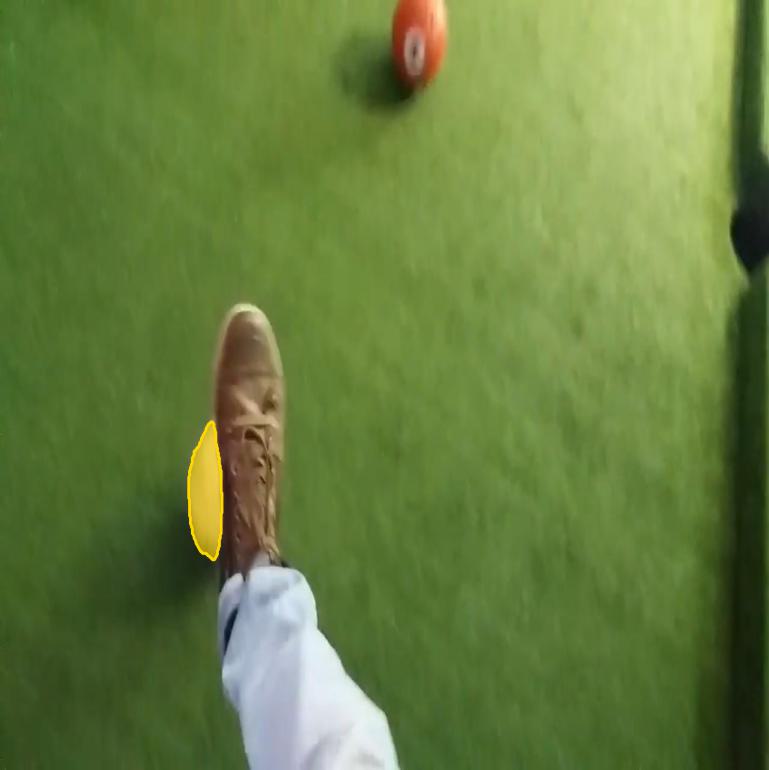} & 
		\includegraphics[width=0.24\textwidth]{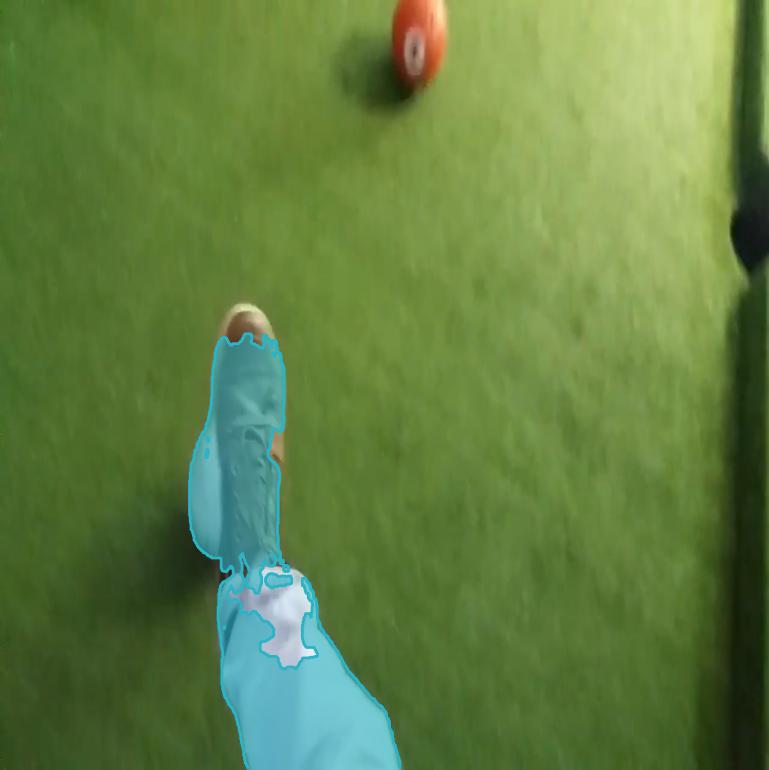} &
		\includegraphics[width=0.24\textwidth]{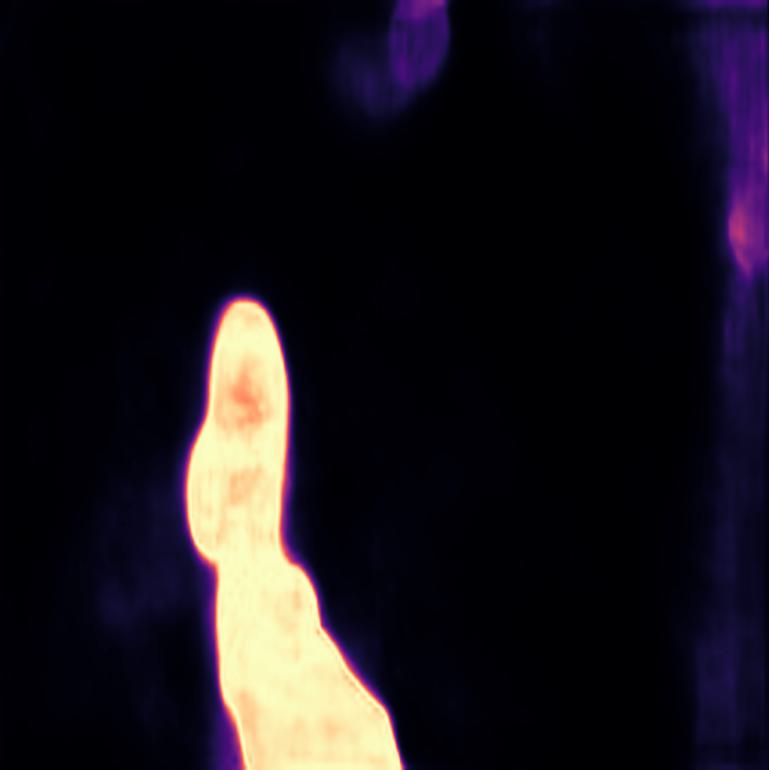} &
		\includegraphics[width=0.24\textwidth]{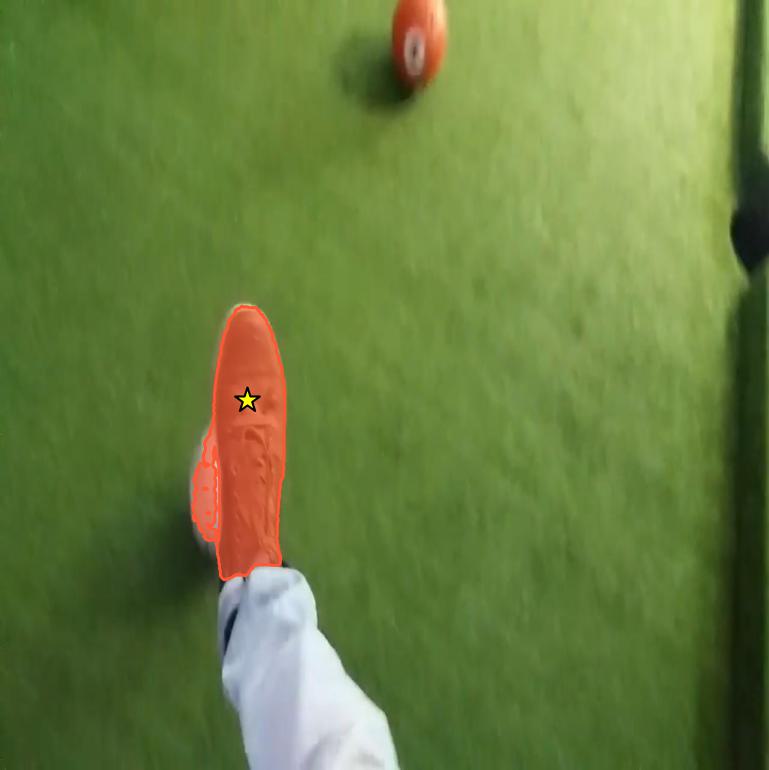}
		\\
	\end{tabular}
	\caption{Qualitative results on the validation set of LVOS~\cite{Hong_2023_ICCV} when refining the mask through pseudo-corrections (Failure cases).}
	\label{tab:images_F}
\end{figure}

Note that for small objects (refer to \cref{tab:images_F}), the mask refinement incorrectly generates masks, although the pseudo-correction location's lies on the target, as seen in rows two, three, and five.
In the first row, the small gorilla (target) is lost in favor to the adult gorilla, since the uncertainty is lower the method fails to issue correct pseudo-corrections or request a user-corrections. Ideally, the method should detect the transition from the small gorilla to the adult gorilla, while the pixel level uncertainty for both objects is still high, to indicate confusion. 
In row 6, we note that the pixel uncertainty for the foot region and the ball (target) are very similar, consequently the method is unable to find a correct location for the pseudo-correction generation as both object are as likely considered as the actual object to track by the \ac{svos} baseline, here the method failed to actually issue a user-correction.

\subsection{User-Corrections}

\begin{figure}
	\centering
	\begin{tabular}{cccc}
		Ground-truth & Original Mask & Entropy & Refined Mask \\
		\includegraphics[width=0.24\textwidth]{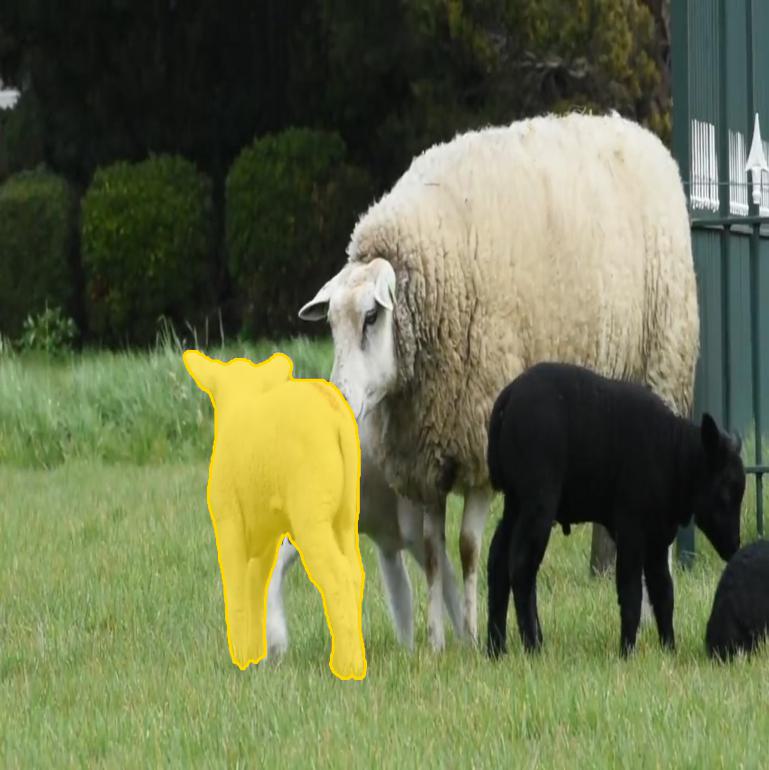} &
		\includegraphics[width=0.24\textwidth]{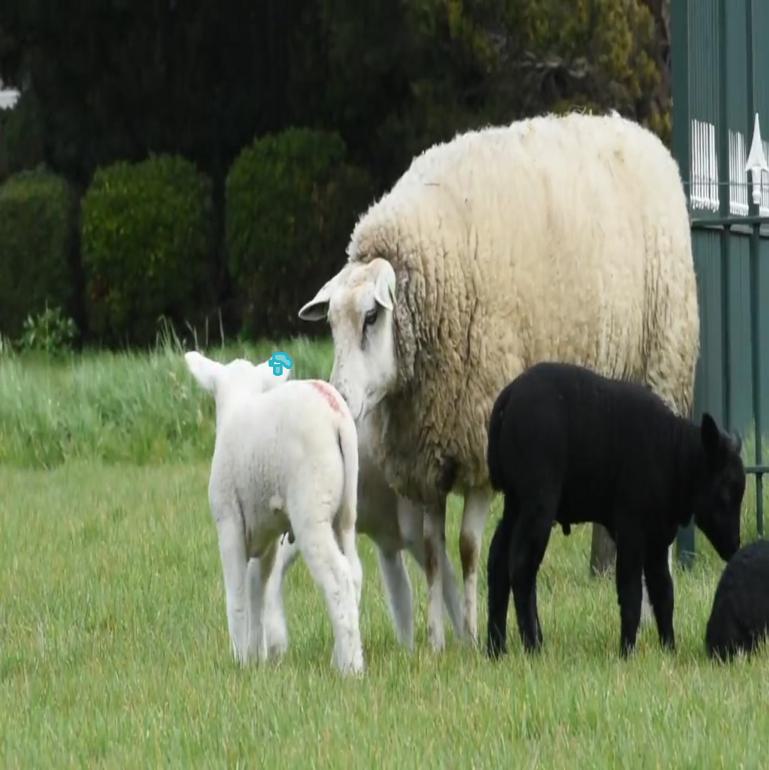} &
		\includegraphics[width=0.24\textwidth]{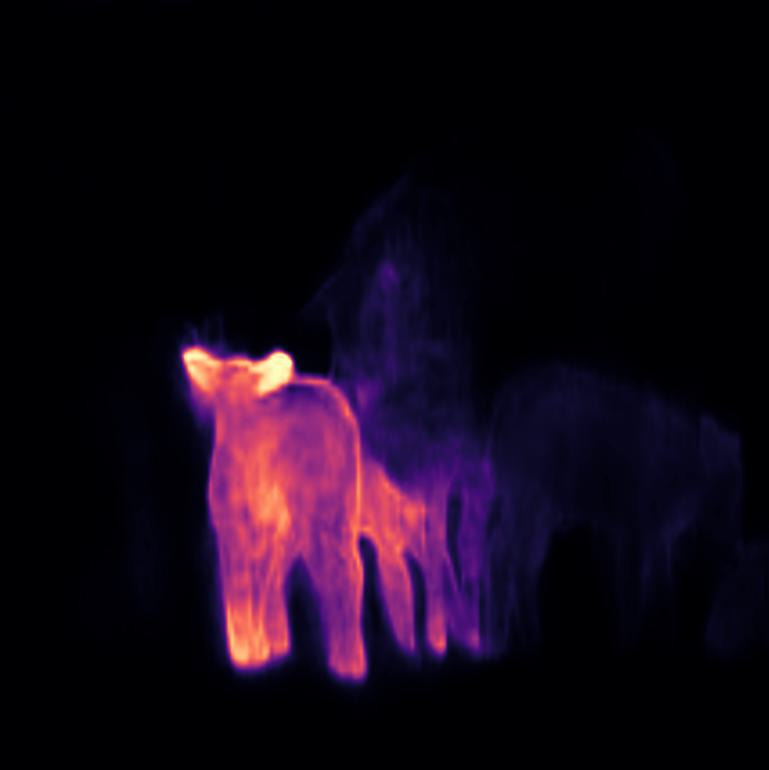} &
		\includegraphics[width=0.24\textwidth]{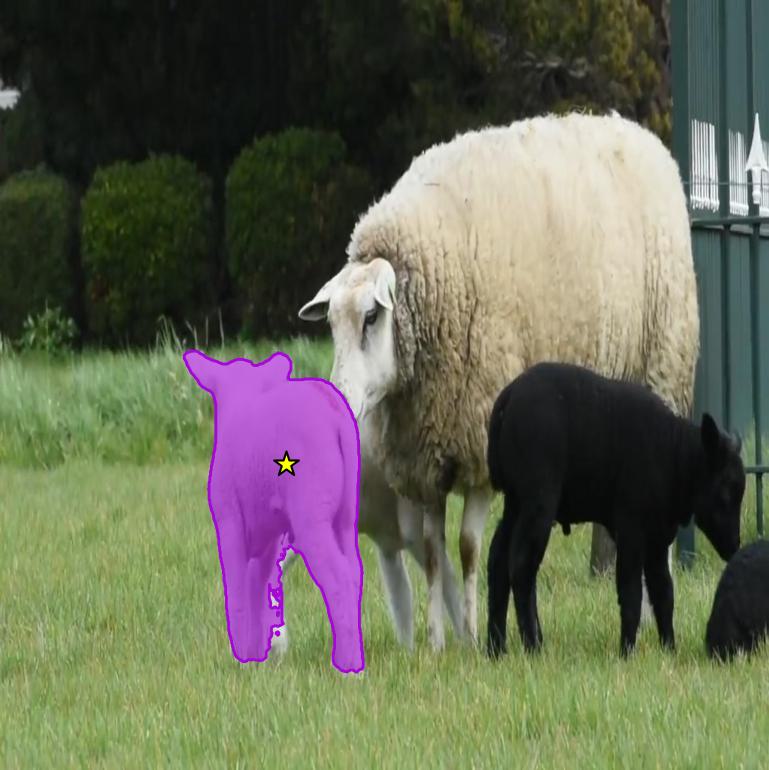} \\
		\includegraphics[width=0.24\textwidth]{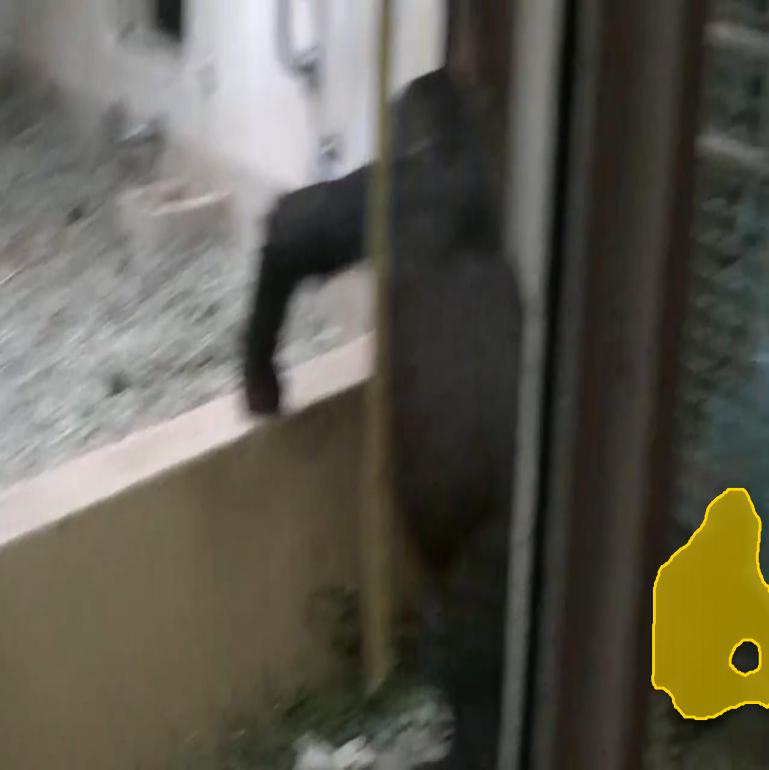} &
		\includegraphics[width=0.24\textwidth]{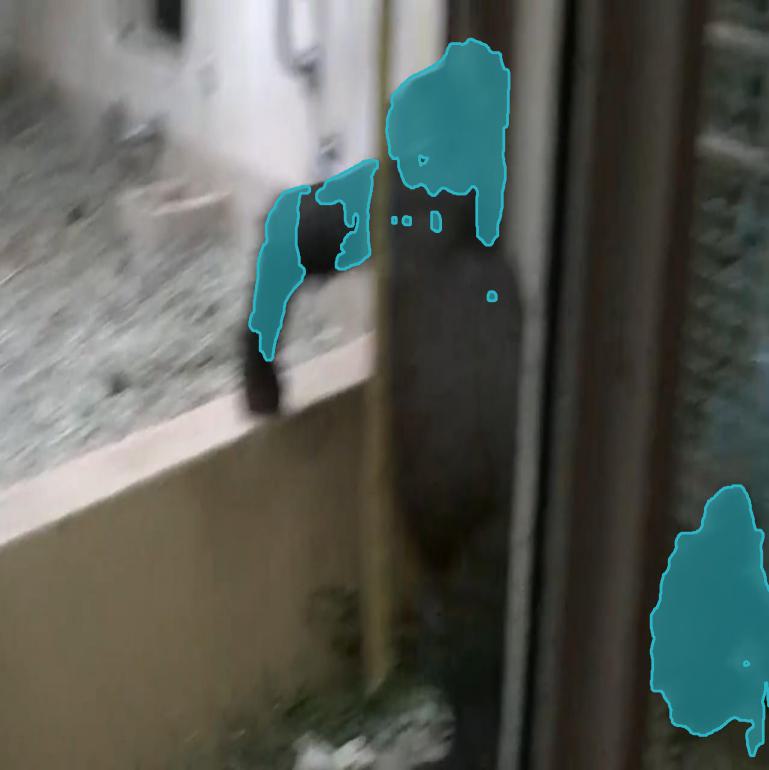} &
		\includegraphics[width=0.24\textwidth]{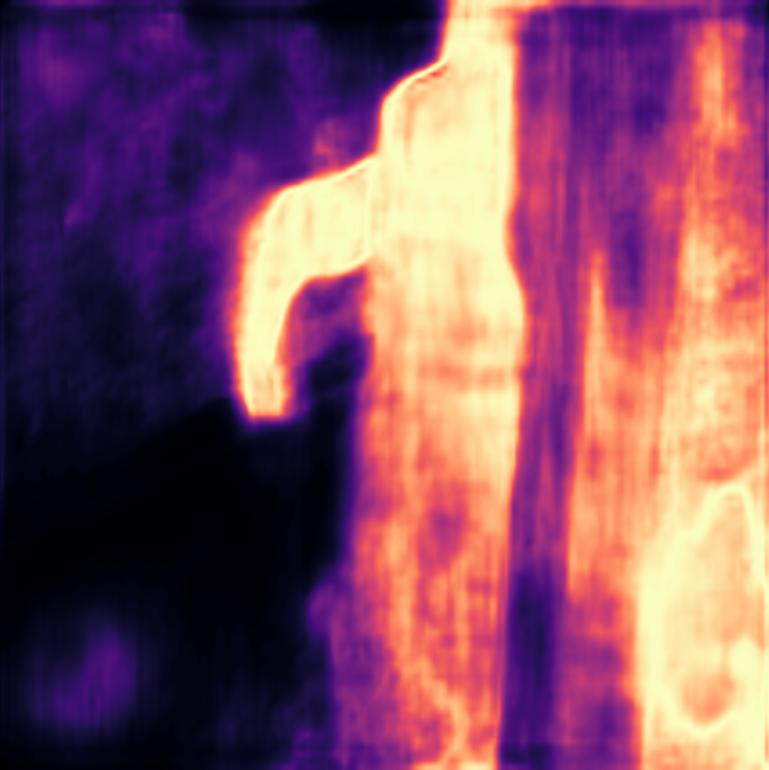} &
		\includegraphics[width=0.24\textwidth]{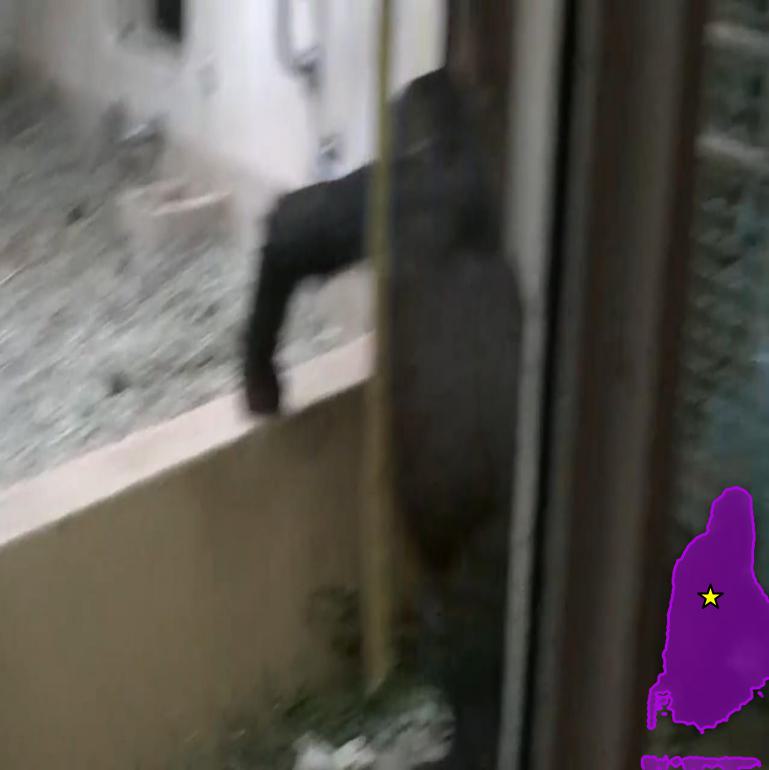} \\
		\includegraphics[trim={4cm 4cm 7cm 7cm},clip, width=0.24\textwidth]{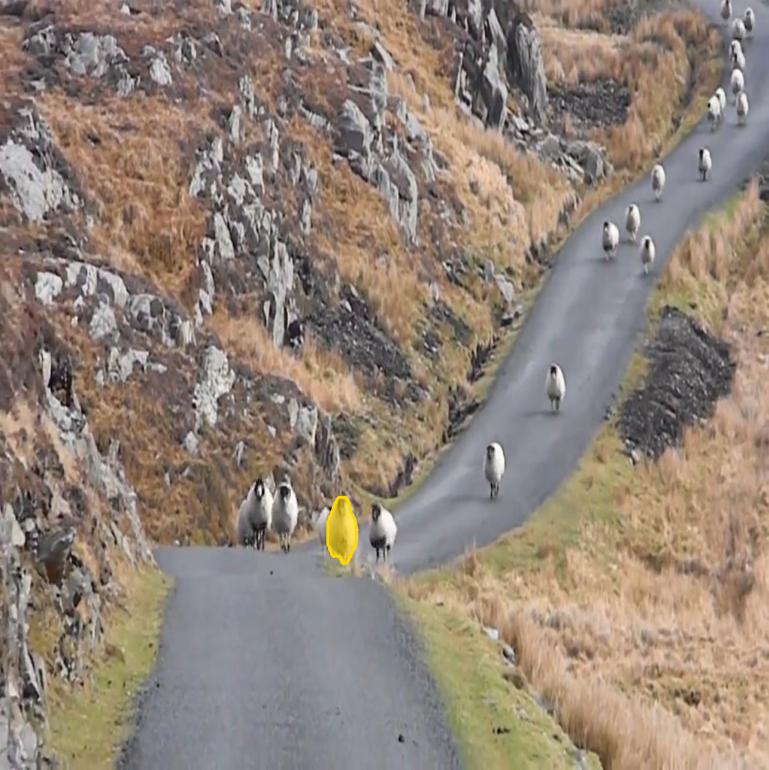}
		\begin{tikzpicture}(0,0)
			\put(-3.05cm,1.55cm){\includegraphics[frame, cfbox=red, height=1.4cm]{D4AgqLQL/D4AgqLQL_81_0_GT.jpg}};
		\end{tikzpicture}
		&
		\includegraphics[trim={4cm 4cm 7cm 7cm},clip, width=0.24\textwidth]{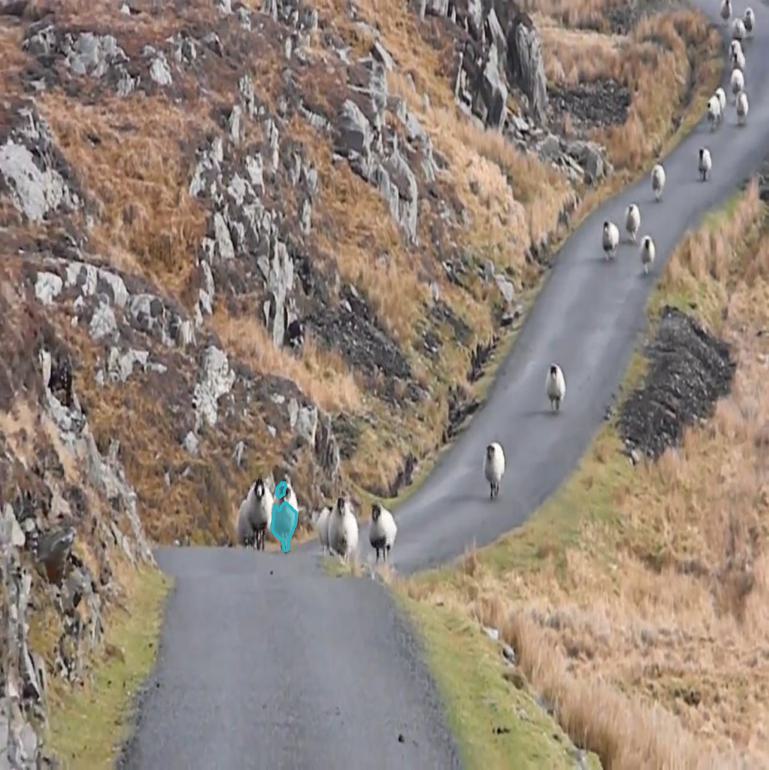} 		 &
		\includegraphics[trim={4cm 4cm 7cm 7cm},clip, width=0.24\textwidth]{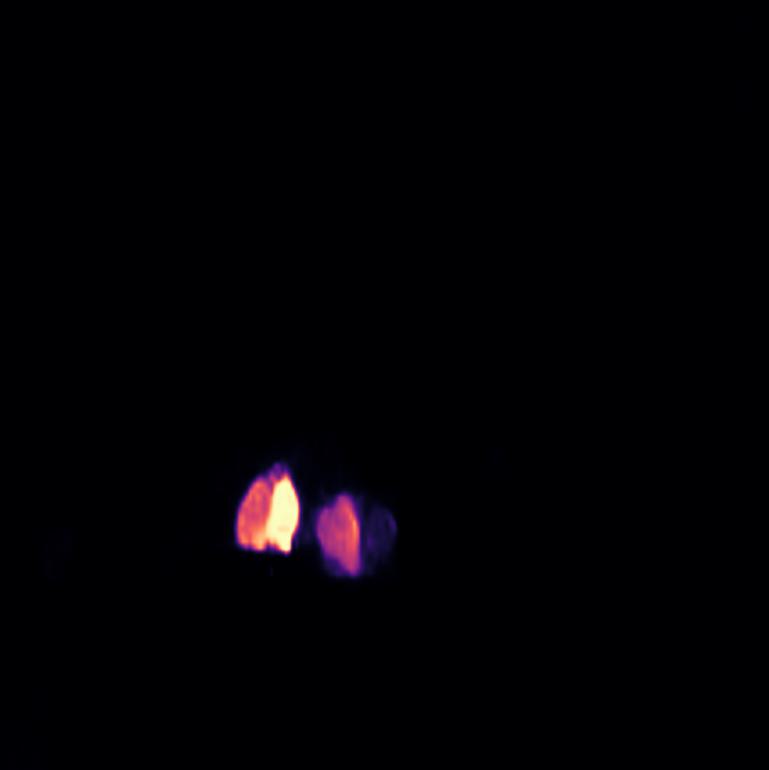} 	 &
		\includegraphics[trim={4cm 4cm 7cm 7cm},clip, width=0.24\textwidth]{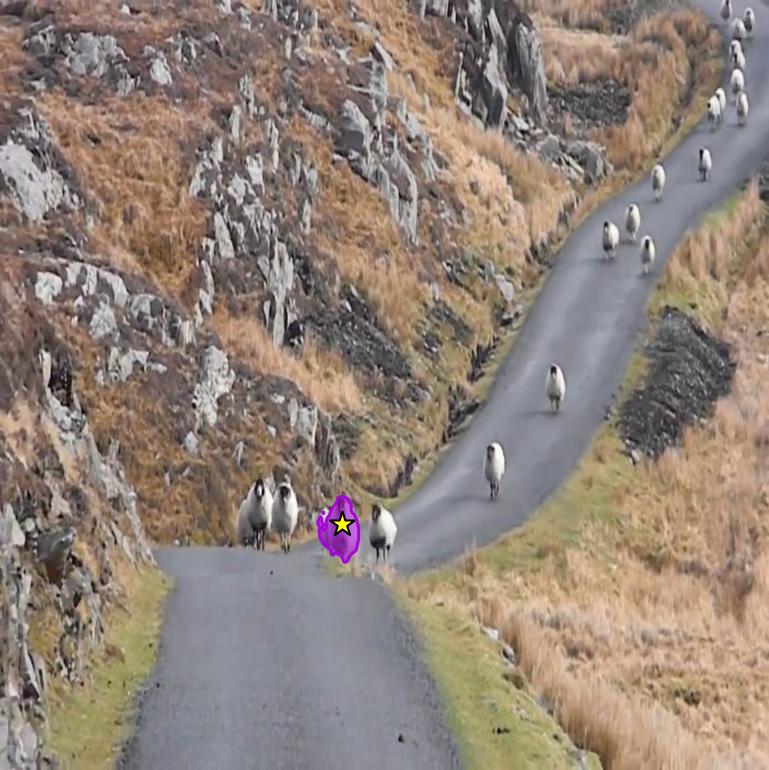} 
		\\
		\includegraphics[trim={0cm 0cm 0cm 0cm},clip, width=0.24\textwidth]{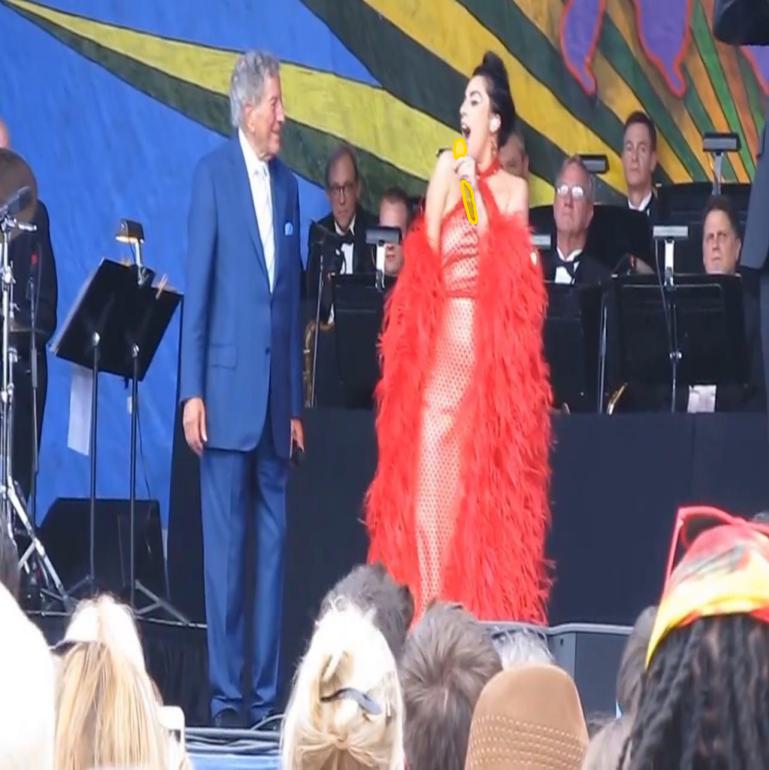}	 &
		\includegraphics[trim={0cm 0cm 0cm 0cm},clip, width=0.24\textwidth]{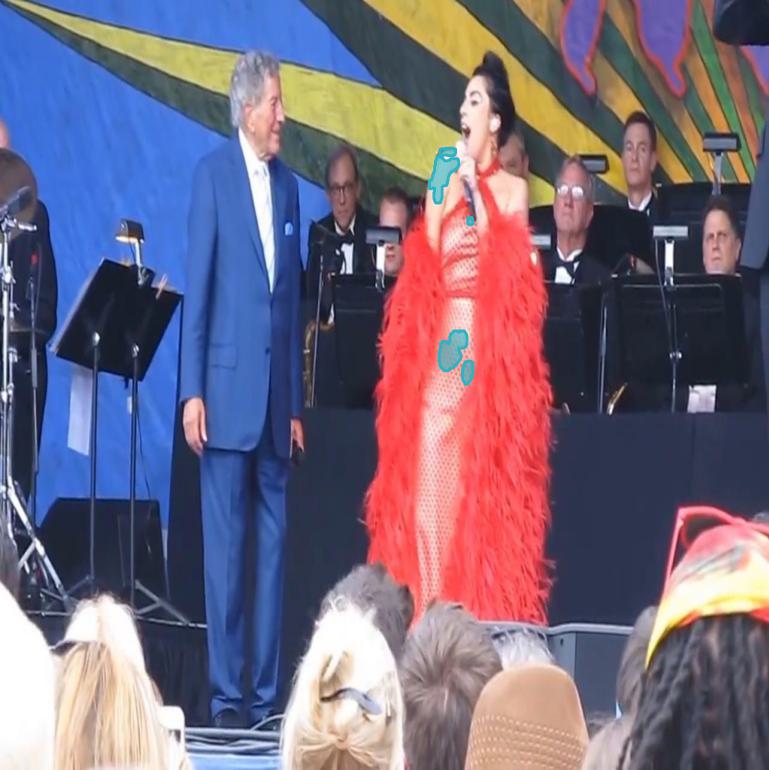} 	 &
		\includegraphics[trim={0cm 0cm 0cm 0cm},clip, width=0.24\textwidth]{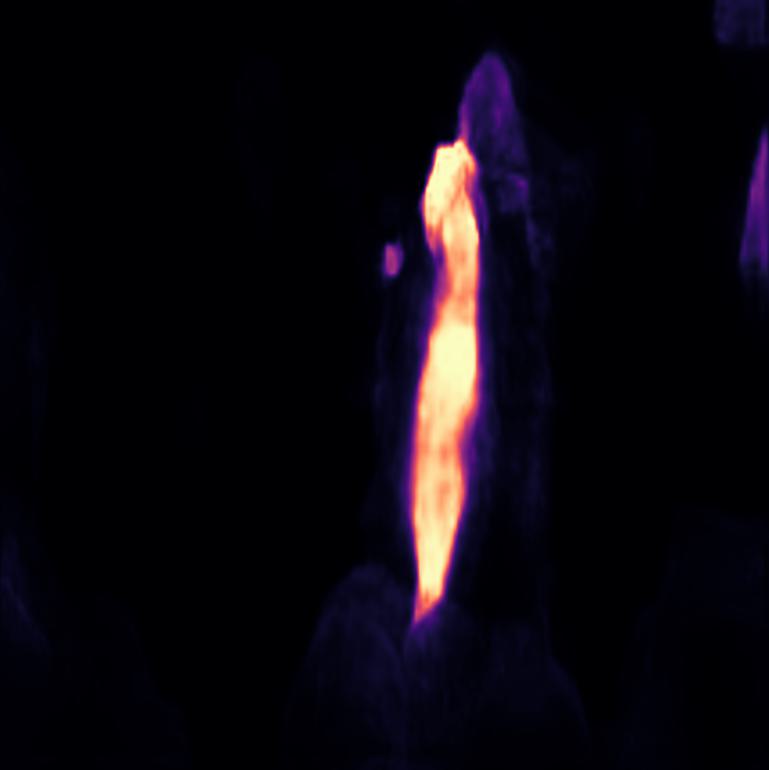} &
		\includegraphics[trim={0cm 0cm 0cm 0cm},clip, width=0.24\textwidth]{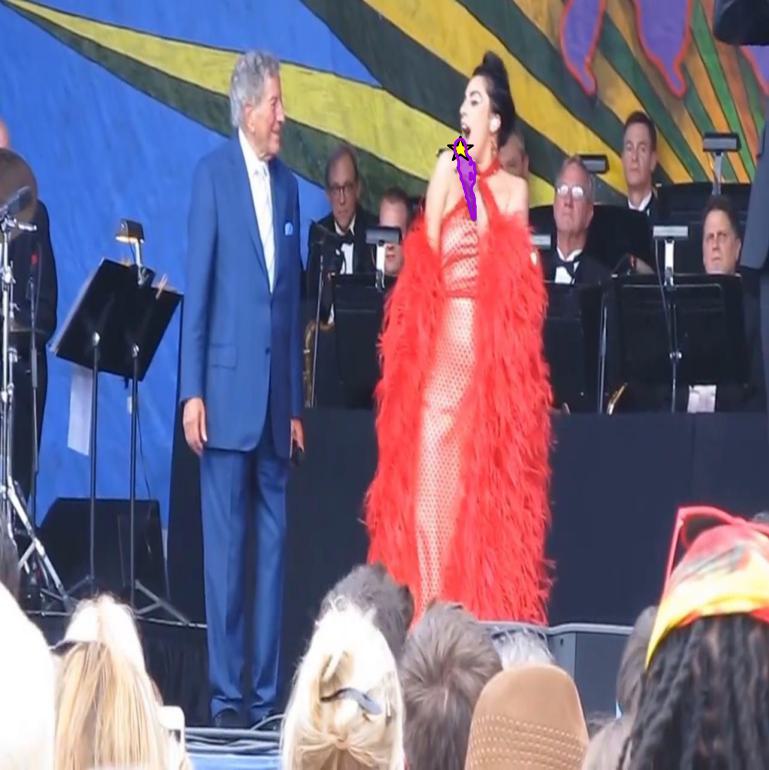}
		\\
		\includegraphics[trim={0cm 0cm 0cm 0cm},clip, width=0.24\textwidth]{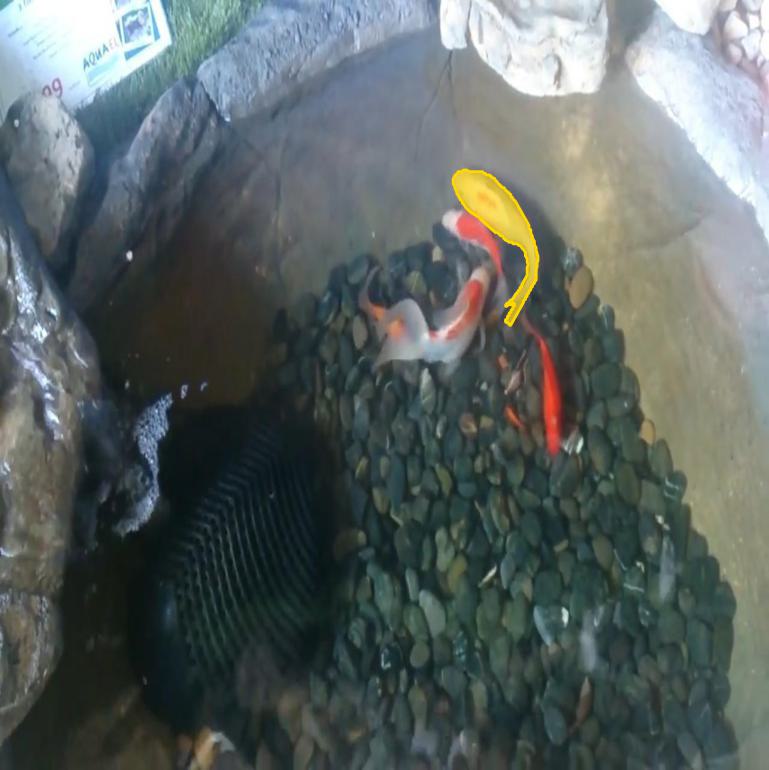}	 &
		\includegraphics[trim={0cm 0cm 0cm 0cm},clip, width=0.24\textwidth]{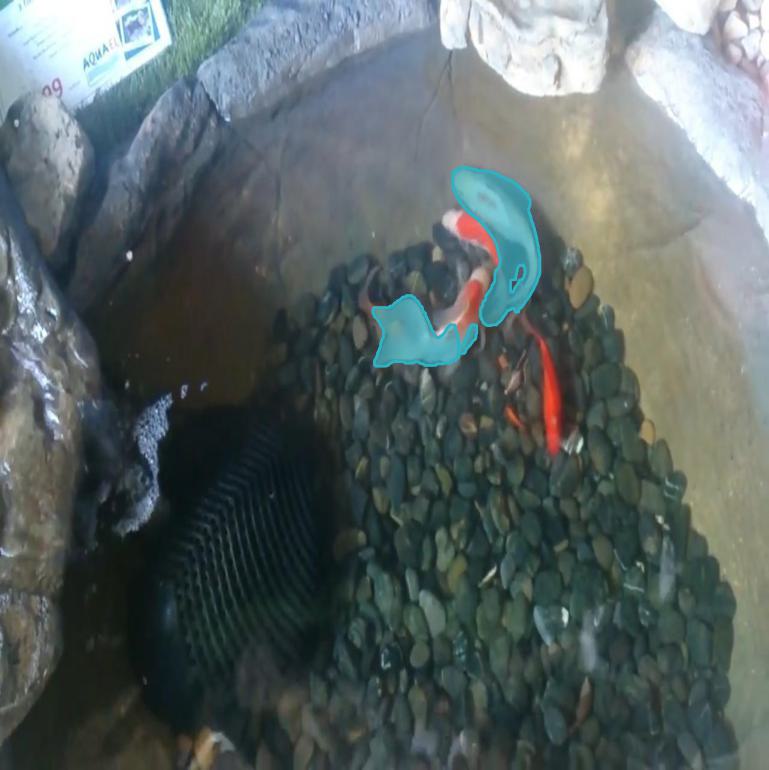} 	 &
		\includegraphics[trim={0cm 0cm 0cm 0cm},clip, width=0.24\textwidth]{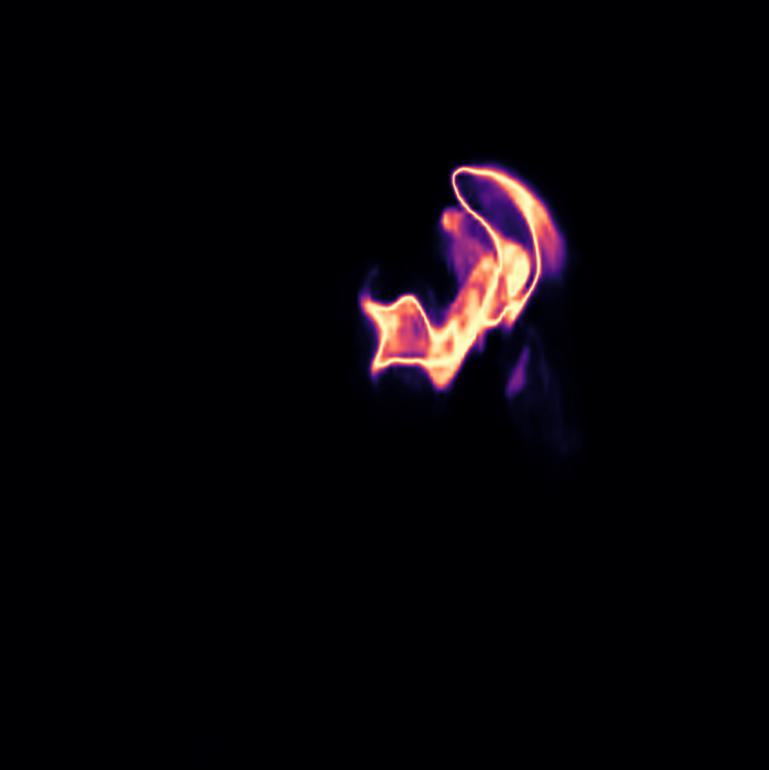} &
		\includegraphics[trim={0cm 0cm 0cm 0cm},clip, width=0.24\textwidth]{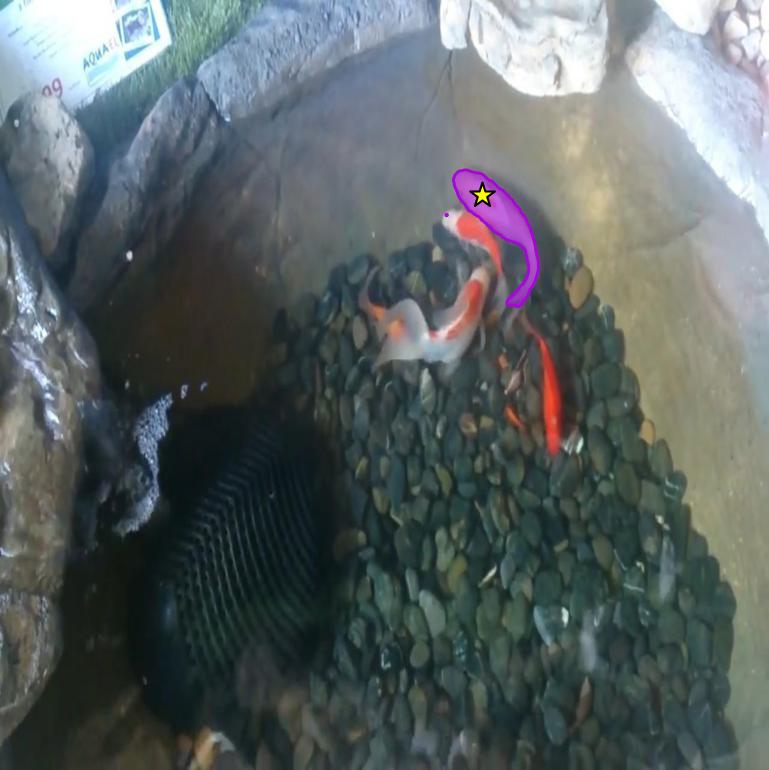}
		\\
		\includegraphics[trim={0cm 0cm 0cm 0cm},clip, width=0.24\textwidth]{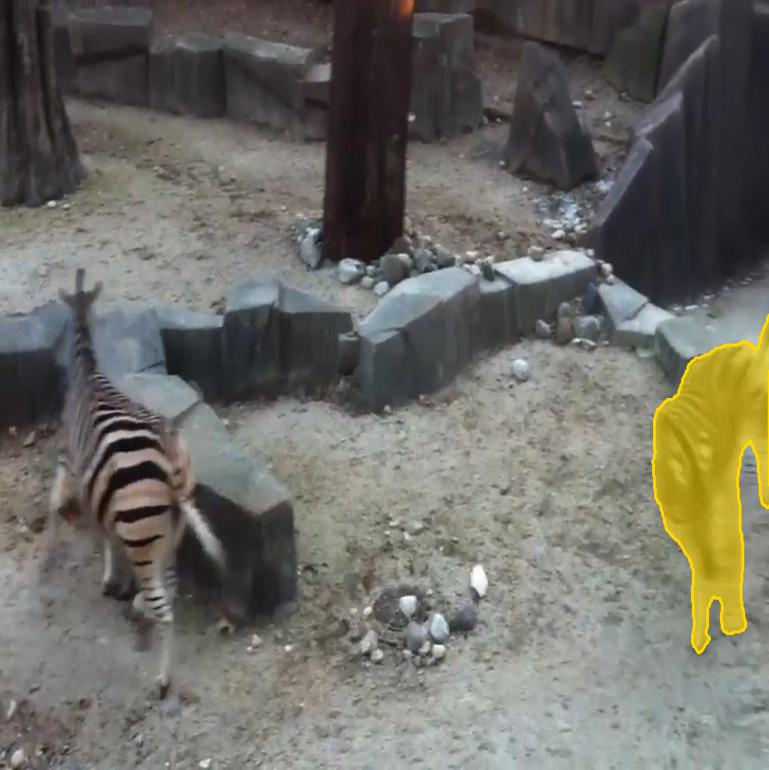}	 &
		\includegraphics[trim={0cm 0cm 0cm 0cm},clip, width=0.24\textwidth]{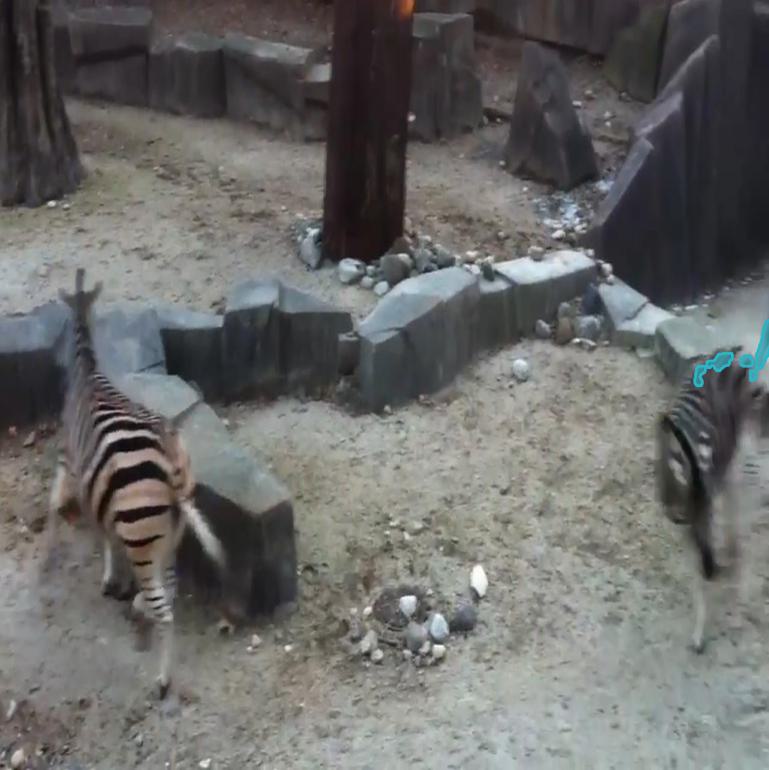} 	 &
		\includegraphics[trim={0cm 0cm 0cm 0cm},clip, width=0.24\textwidth]{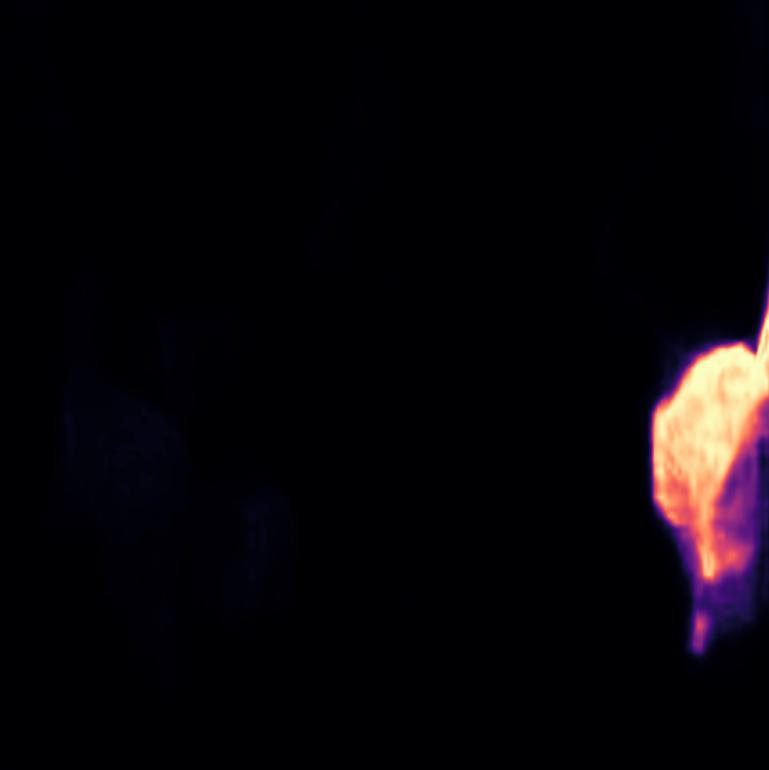} &
		\includegraphics[trim={0cm 0cm 0cm 0cm},clip, width=0.24\textwidth]{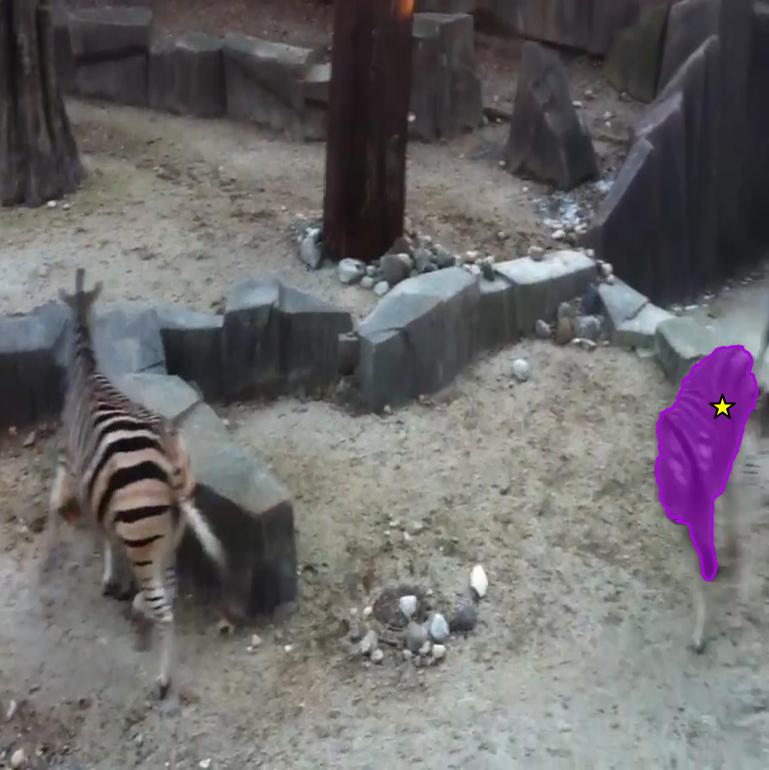} \\
	\end{tabular}
	\caption{Qualitative results on the validation set of LVOS~\cite{Hong_2023_ICCV} when refining the mask through user-corrections (Success cases).}
	\label{tab:images_U_good}
\end{figure}

In the first and last rows of~\cref{tab:images_U_good}, we note that the method correctly issues a user interactions, as only the ear of the sheep and the back of the zebra are still segmented, preventing the loss of the target.
Similarly, in the second and third rows, the method manages to issue and interaction to the user while losing the target in favor to a distractors.
Note that in the third row, the method correctly issues a user-correction instead of a pseudo-correction, as otherwise the pseudo-correction would be generated on the wrong sheep.

\begin{figure}
	\centering
	\begin{tabular}{cccc}
		Ground-truth & Original Mask & Entropy & Refined Mask \\
		\includegraphics[width=0.24\textwidth]{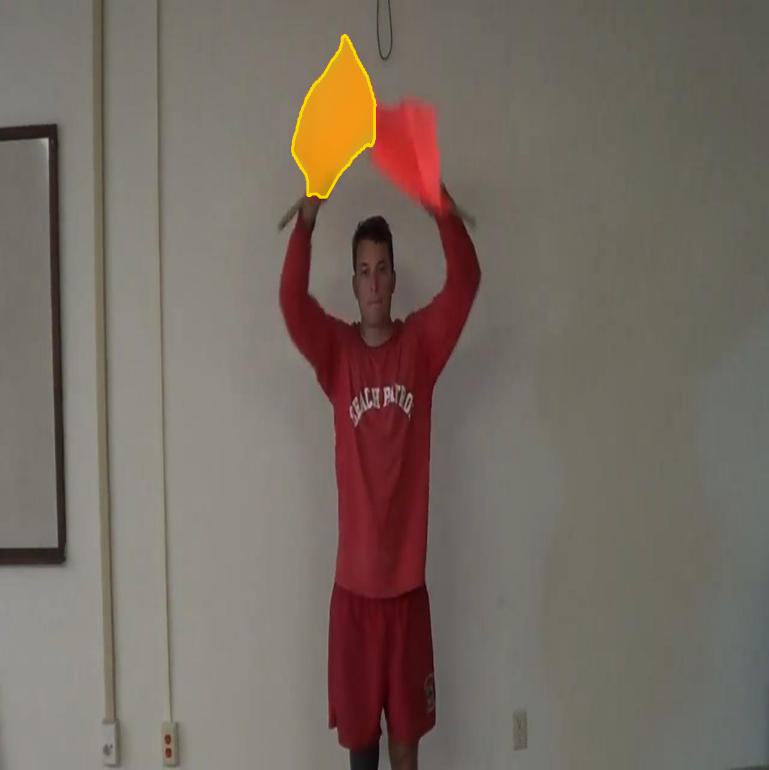} &
		\includegraphics[width=0.24\textwidth]{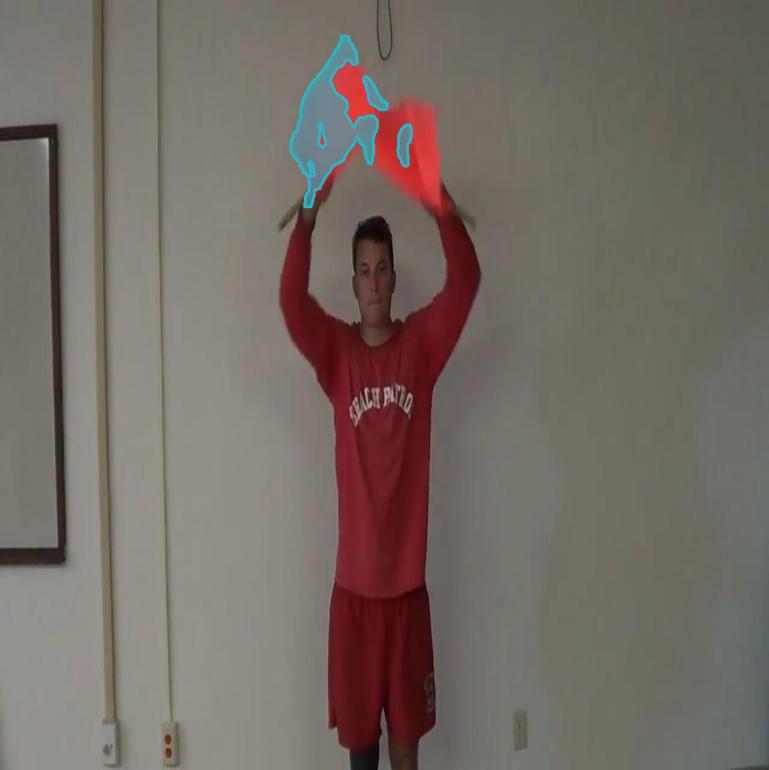} &
		\includegraphics[width=0.24\textwidth]{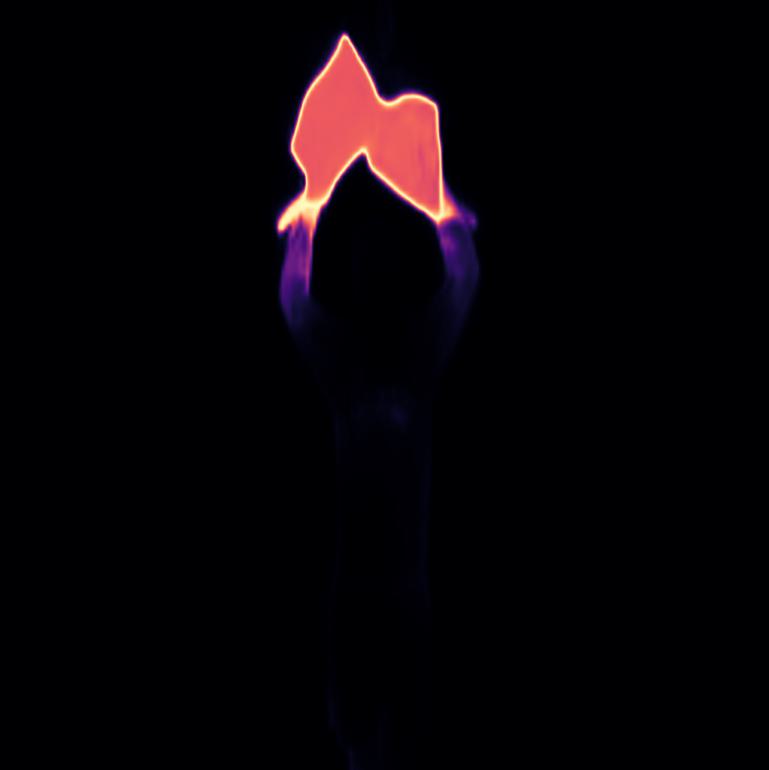} &
		\includegraphics[width=0.24\textwidth]{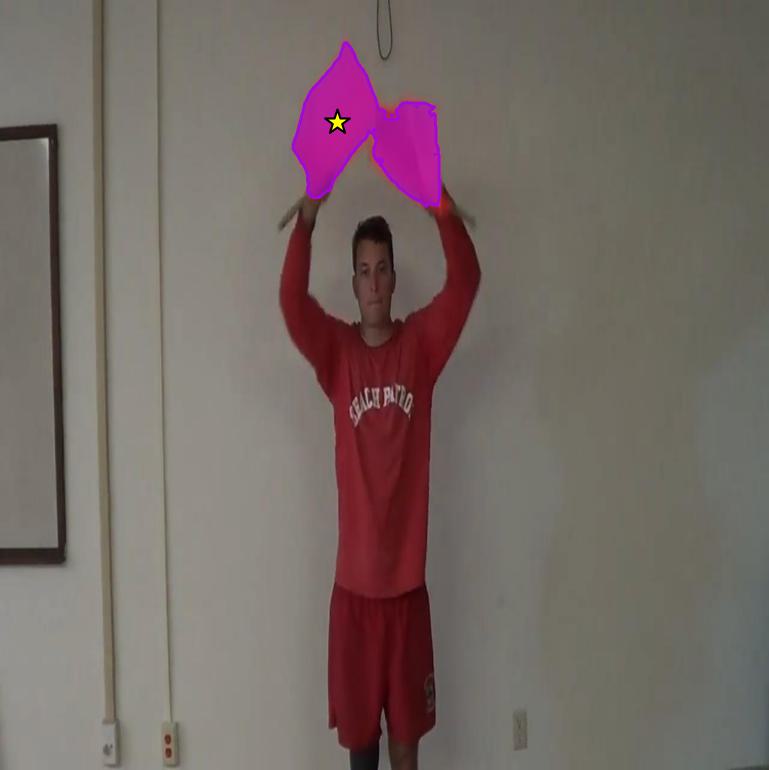} \\
		\includegraphics[width=0.24\textwidth]{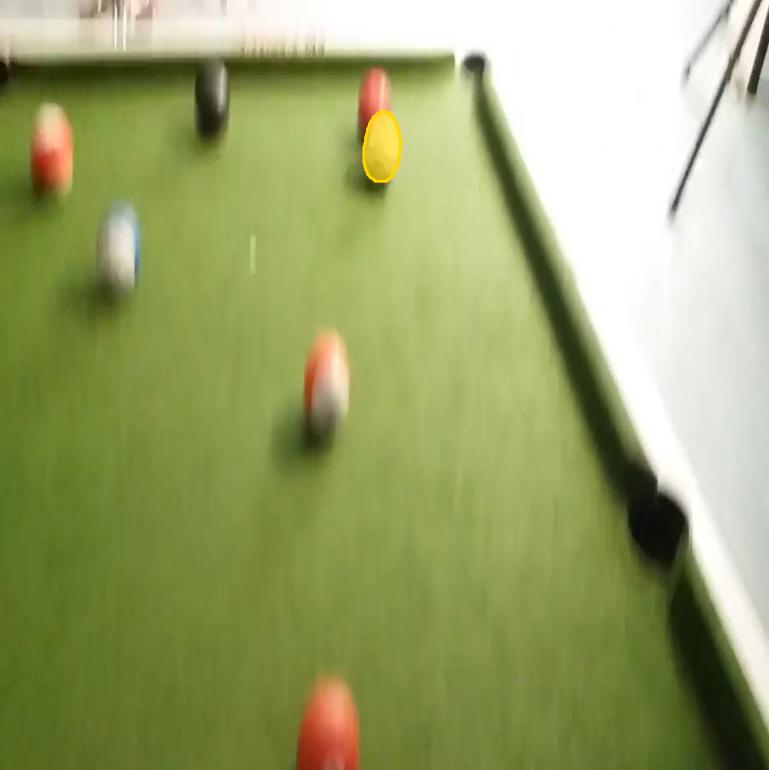} & 
		\includegraphics[width=0.24\textwidth]{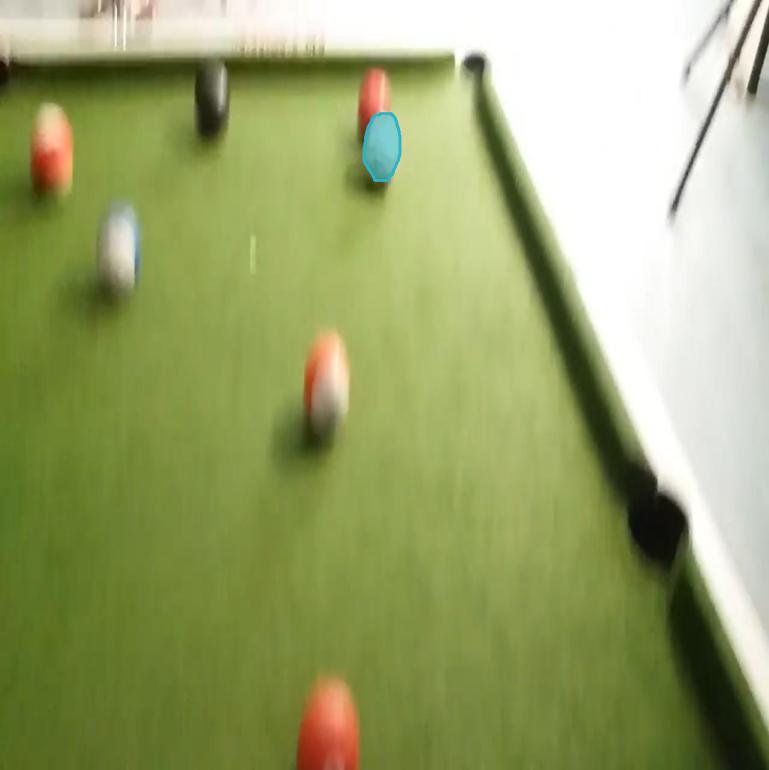} &
		\includegraphics[width=0.24\textwidth]{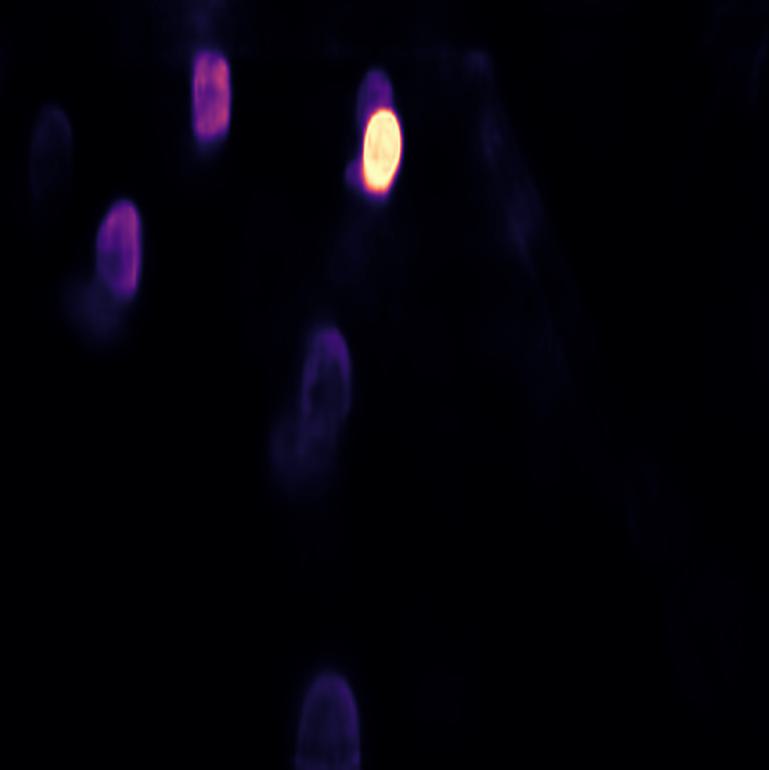} &
		\includegraphics[width=0.24\textwidth]{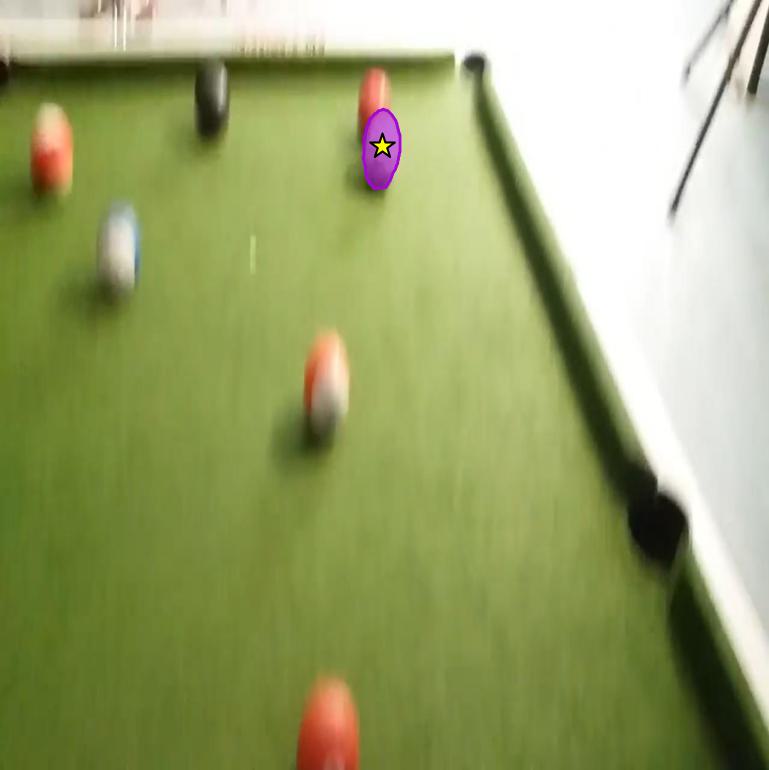}
		\\
		\includegraphics[trim={6cm 4cm 8cm 4cm},clip, width=0.24\textwidth, height=0.24\textwidth]{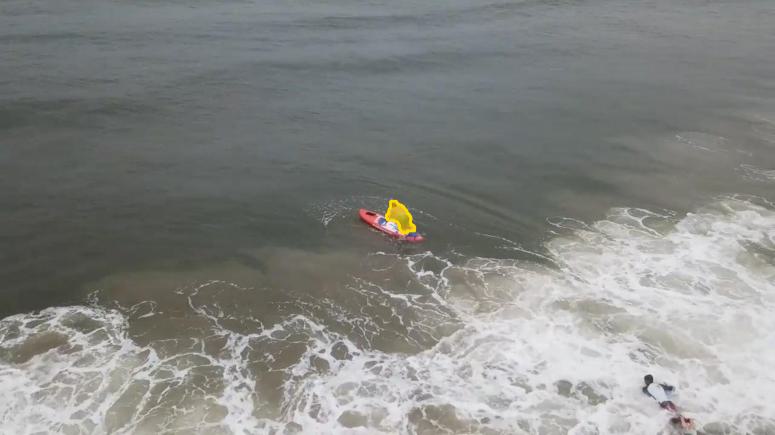} 
		\begin{tikzpicture}(0,0)
			\put(-3.05cm,0){\includegraphics[frame, cfbox=red, height=1.4cm, width=1.4cm]{VhwRKgVS/VhwRKgVS_27_2_under_0.1_GT.jpg}};
		\end{tikzpicture}
		& 
		\includegraphics[trim={6cm 4cm 8cm 4cm}, clip, width=0.24\textwidth, height=0.24\textwidth]{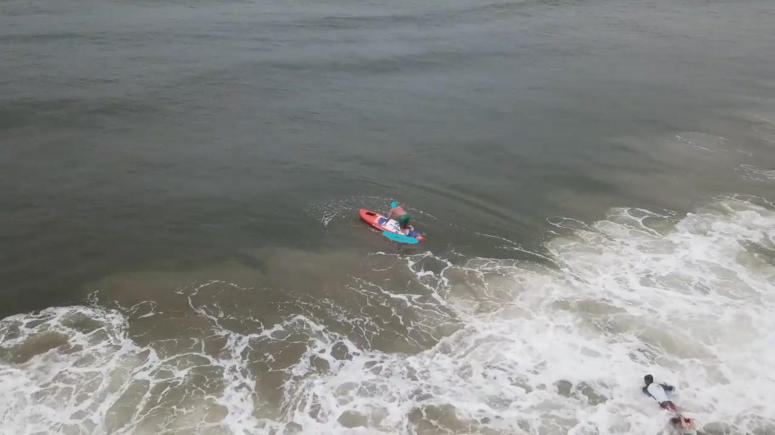} & 
		\includegraphics[trim={6cm 4cm 8cm 4cm}, clip, width=0.24\textwidth, height=0.24\textwidth]{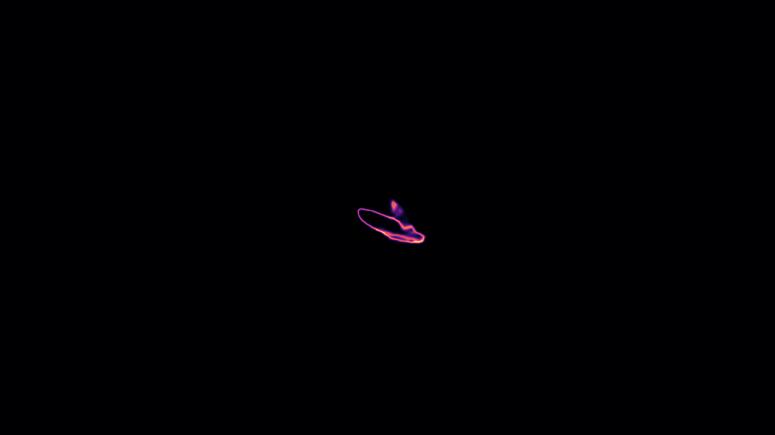}
		& 
		\\
		\includegraphics[width=0.24\textwidth, height=0.24\textwidth]{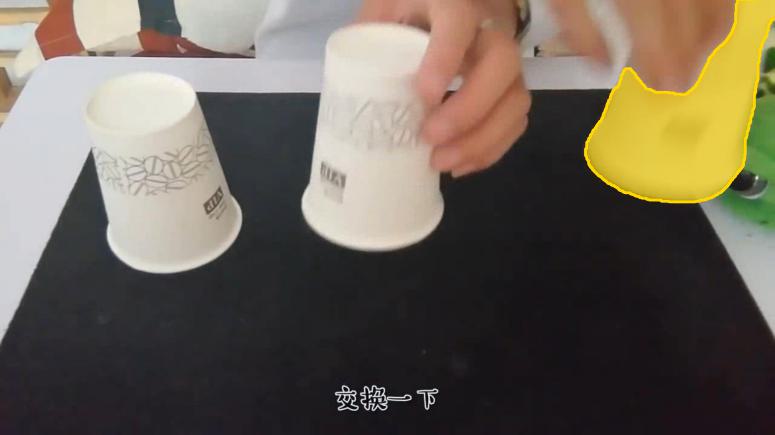} & 
		\includegraphics[width=0.24\textwidth,
		height=0.24\textwidth]{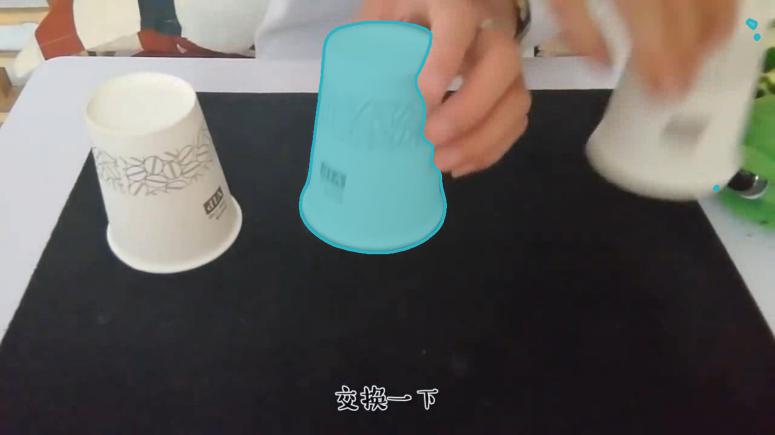} &
		\includegraphics[width=0.24\textwidth,
		height=0.24\textwidth]{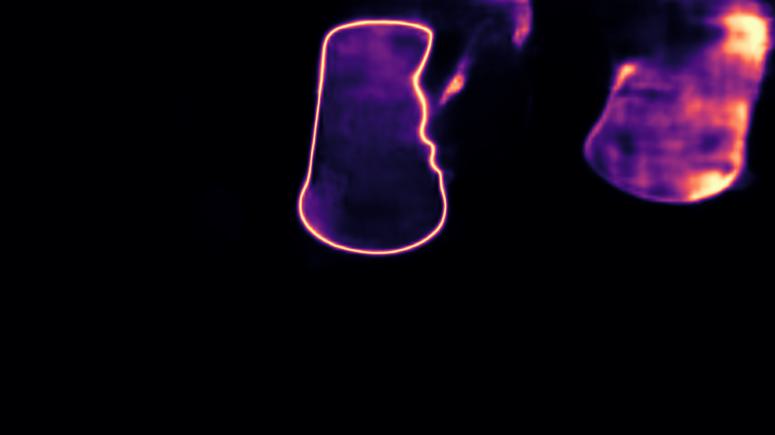} &
		\\
		\includegraphics[width=0.24\textwidth, height=0.24\textwidth]{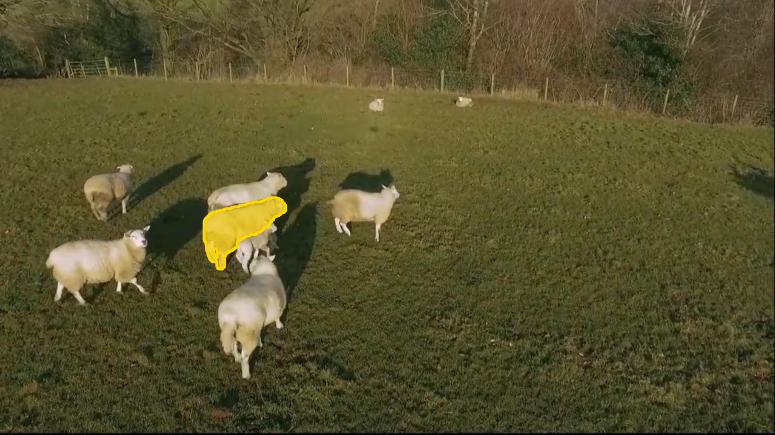} & 
		\includegraphics[width=0.24\textwidth,
		height=0.24\textwidth]{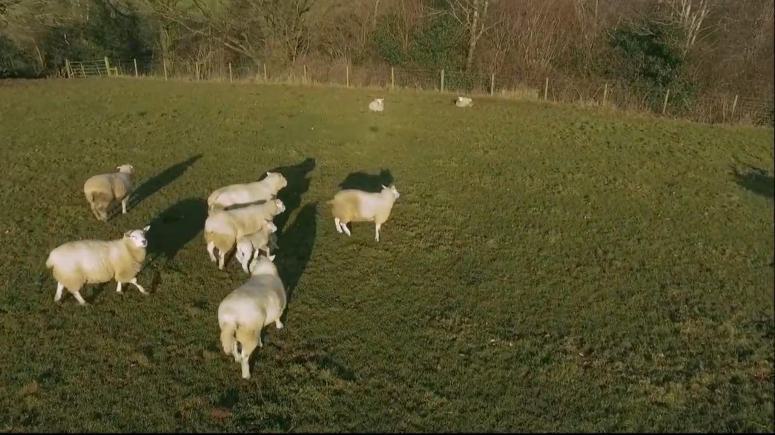} &
		\includegraphics[width=0.24\textwidth,
		height=0.24\textwidth]{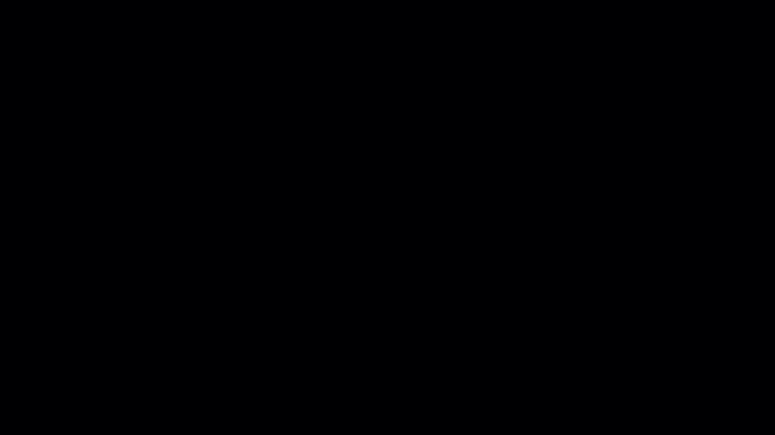} &
		
	\end{tabular}
	\caption{Qualitative results on the validation set of LVOS~\cite{Hong_2023_ICCV} when refining the mask through user-corrections (Failure and miss cases). Here we considered a missed opportunity to generate a pseudo- or user-correction whenever the \ac{iou} between the original prediction and the ground-truth annotation is below $0.1$.}
	\label{tab:images_U_bad}
\end{figure}

In~\cref{tab:images_U_bad}, we observe that the method sometimes unnecessarily calls for user interaction even when a good portion of the object is correctly predicted (\ie, first and second row), and where a pseudo-correction would be more appropriate (first row).

Additionally, there are instances where a user (or pseudo) correction is missed, as seen in rows three, four and five. In the fourth row, the tracker confidently segments a distractor after the disappearance of the object of interest, while indicating the actual object with some uncertainty. Lastly, when the SVOS backbone loses track of the object of interest, it is unable to recover it, as shown in the fifth row.

\section{Kernel Size for Dilating the Mask}
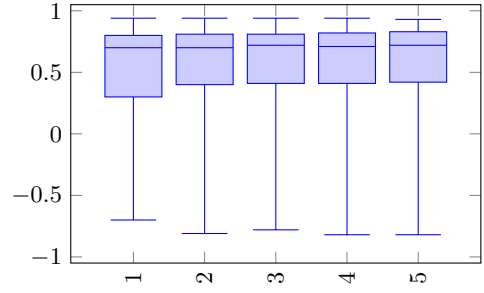
\begin{figure}
	\centering
	\begin{subfigure}{0.49\textwidth}
		\begin{tikzpicture}
			\begin{axis}
				[
				xtick={1, 2, 3, 4, 5},
				xticklabels={1, 2, 3, 4, 5},
				x tick label style={rotate=90},
			boxplot/draw direction=y,
			width=7cm,height=5cm,
			ymin=-1.05, ymax=1.05]
			\addplot[
			fill,
			fill opacity=0.2,
			color=susielblue,
			boxplot prepared={
				lower whisker=0.94,
				lower quartile=0.8,
				median=0.7,
				upper quartile=0.3,
				upper whisker=-0.7
			},
			] coordinates {};
			\addplot[
			fill,
			fill opacity=0.2,
			color=susielblue,
			boxplot prepared={
				lower whisker=0.94,
				lower quartile=0.81,
				median=0.7,
				upper quartile=0.4,
				upper whisker=-0.81
			},
			] coordinates {};
			\addplot[
			fill,
			fill opacity=0.2,
			color=susielblue,
			boxplot prepared={
				lower whisker=0.94,
				lower quartile=0.81,
				median=0.72,
				upper quartile=0.41,
				upper whisker=-0.78
			},
			] coordinates {};
			\addplot[
			fill,
			fill opacity=0.2,
			color=susielblue,
			boxplot prepared={
				lower whisker=0.94,
				lower quartile=0.82,
				median=0.71,
				upper quartile=0.41,
				upper whisker=-0.82
			},
			] coordinates {};
			\addplot[
			fill,
			fill opacity=0.2,
			color=susielblue,
			boxplot prepared={
				lower whisker=0.93,
				lower quartile=0.83,
				median=0.72,
				upper quartile=0.42,
				upper whisker=-0.82
			},
			] coordinates {};
		\end{axis}
	\end{tikzpicture}
	\caption{Spearman correlation distribution for different kernel sizes when computing the masked entropy~$S_{\mathcal{R}_c}$ on the DAVIS 2017 validation set~\cite{Pont-Tuset_arXiv_2017}.}
\end{subfigure}
\caption{Varying the dilation of the masked entropy}
\label{dila_rate}
\end{figure}

In \cref{dila_rate}, we present the distribution of the Spearman correlation coefficient~\cite{Spearman1904} on the DAVIS dataset~\cite{Pont-Tuset_arXiv_2017}. Our experiments use a kernel size of 2. However, as shown in \cref{dila_rate}, this choice is rather permissive, as larger kernel sizes (and none for the first case) yield similar outcomes.

\section{Limitations and Future Directions}
Currently, Lazy-XMem generates only click-based pseudo-corrections, which are fed to the mask-refiner without including the predicted mask.
This approach limits the impact of the initial mask proposed by the \ac{svos} pipeline. 

This bottleneck is inherent to SAM-based models, as they do not consider masks as prompts in practice.
An alternative approach, explored by Delatolas \etal~\cite{delatolas2024learning}, involves iteratively prompting the mask-refiner with pseudo-prompts generated from the initial mask until a certain level of alignment is achieved between the SAM-predicted mask and the original \ac{svos} initial mask.
However, this method assumes that the initial mask (from the \ac{svos} pipeline) is accurate enough to serve as a reliable base for further prompting the mask-refiner with uncertainty-based prompts.

An additional direction to follow in future work is to incorporate other types of prompts, like bounding-boxes or scribble-type, which might add more context to the prompt.
Additionally, while we mostly rely on positive pseudo-clicks, including negative interactions could further enhance the method's capabilities.


\end{document}